\documentclass{article}
\pdfoutput=1

\usepackage[preprint]{neurips_2024}

\usepackage{amsmath}
\usepackage{multirow}
\usepackage{graphicx} 
\usepackage{subcaption}

\usepackage{algorithm}
\usepackage{algorithmic}
\usepackage{overpic}
\usepackage{multirow}
\usepackage{rotating}
\usepackage[table]{xcolor}
\usepackage{enumitem}
\usepackage[utf8]{inputenc} % allow utf-8 input
\usepackage[T1]{fontenc}    % use 8-bit T1 fonts
\usepackage{hyperref}       % hyperlinks
\usepackage{url}            % simple URL typesetting
\usepackage{booktabs}       % professional-quality tables
\usepackage{amsfonts}       % blackboard math symbols
\usepackage{nicefrac}       % compact symbols for 1/2, etc.
\usepackage{microtype}      % microtypography
\usepackage{xcolor}         % colors
\usepackage{comment}
\usepackage{wrapfig}
\usepackage{makecell}

\newtheorem{technique}{Technique}

\newcommand{\wz}[1]{\textcolor{red}{\textsf{[wz: #1]}}}

\usepackage{xspace}

\makeatletter
\DeclareRobustCommand\onedot{\futurelet\@let@token\@onedot}
\def\@onedot{\ifx\@let@token.\else.\null\fi\xspace}

\def\eg{\emph{e.g}\onedot} 
\def\ie{\emph{i.e}\onedot} 
 
\def\etc{\emph{etc}\onedot}

\makeatother

\newcommand{\name}{LatentDEM}

\newcommand{\signal}{\boldsymbol{x}}
\newcommand{\observation}{\boldsymbol{y}}
\newcommand{\forwardFunc}{\mathcal{A}}

\newcommand{\forwardParam}{{\phi}}

\newcommand{\probDistribution}[1]{p(#1)}
\newcommand{\scoreFunc}[2]{\nabla_{#1} \log p_t({#2})}

\newcommand{\finiteNote}[1]{\mathrm{d}{#1}}
\newcommand{\noisetime}{t}
\newcommand{\wiener}{\boldsymbol{w}}

\newcommand{\latent}{\boldsymbol{z}}

\newcommand{\decodeNote}{\mathcal{D}}

\newcommand{\pretraiedDecodeFunc}[1]{\decodeNote^*\left( #1 \right)}

\newcommand{\sdModel}{\boldsymbol{s}_{\theta}}
\newcommand{\pretrainSdModel}{\sdModel^*}

\newcommand{\image}{\boldsymbol{x}}

\newcommand{\skipEnd}{S_T}
\newcommand{\skipNum}{K}
\title{Blind Inversion using Latent Diffusion Priors}

\author{%
  Weimin Bai\thanks{Equal contribution.}\qquad Siyi Chen$^*$\qquad Wenzheng Chen$^\dag$ \qquad He Sun\thanks{Correspondence to hesun@pku.edu.cn and wenzhengchen@pku.edu.cn} \\
  \\
  Peking University\\
  \\
  \href{https://ai4imaging.github.io/latentdem/}{\large{\color{blue}{\textbf{Project Page}}}} 
}

\begin{document}

\maketitle

\begin{abstract}
  % We introduce the first universal inverse problem solver LatentEM: no need for domain-specific priors nor the knowledge of the forward operators.

Diffusion models have emerged as powerful tools for solving inverse problems due to their exceptional ability to model complex prior distributions. However, existing methods predominantly assume known forward operators (\ie, non-blind), limiting their applicability in practical settings where acquiring such operators is costly. Additionally, many current approaches rely on pixel-space diffusion models, leaving the potential of more powerful latent diffusion models (LDMs) underexplored.
In this paper, we introduce {\name}, an innovative technique that addresses more challenging blind inverse problems using latent diffusion priors. At the core of our method is solving blind inverse problems within an iterative Expectation-Maximization (EM) framework:  (1) the E-step recovers clean images from corrupted observations using LDM priors and a known forward model, and (2) the M-step estimates the forward operator based on the recovered images. Additionally, we propose two novel optimization techniques tailored for LDM priors and EM frameworks, yielding more accurate and efficient blind inversion results.
As a general framework, {\name} supports both linear and non-linear inverse problems. Beyond common 2D image restoration tasks, it enables new capabilities in non-linear 3D inverse rendering problems. We validate {\name}'s performance on representative 2D blind deblurring and 3D sparse-view reconstruction tasks, demonstrating its superior efficacy over prior arts. % and adaptability to real-world scenarios. \wz{if mentioning real world, can we show real examples? Otherwise change it to ...over prior arts.}

\end{abstract}
\vspace*{-0.2cm}
\section{Introduction}

% inverse problem is important
% inverse problem rely on priors, ldm is better than dm
% most methods focus on 
% 

%Inverse problems aim to recover underlying signals $\signal$ from partial or corrupted observations $\observation$ generated by a forward operator $\forwardFunc_\forwardParams(\cdot)$. 
%Such problems are prevalent in computational imaging, ranging from 2D image restoration like denoising, deblurring, inpainting to 3D imaging tasks such as CT, NLOS and inverse rendering~\cite{bibid}. 
%Typically, inverse problem solvers assume knowing the forward model and its physical parameters, $\forwardParams$ (\ie, \textit{non-blind})~\cite{bibid}. 
%However, acquiring accurate forward models is often challenging or impractical in real-world settings, necessitating a more complex task of solving \textit{blind} inverse problems, where both the hidden signals $\signal$ and the forward model parameters $\forwardParam$ must be jointly estimated.

Inverse problems aim to recover underlying signals $\signal$ from partial or corrupted observations $\observation$ generated by a forward operator $\forwardFunc_\forwardParam(\cdot)$. Such problems are prevalent in computer vision and graphics, encompassing a variety of tasks ranging from 2D image restoration(denoising, deblurring, and inpainting~\cite{bertero2021introduction, bertalmio2000image}) to 3D reconstruction(CT, NLOS, inverse rendering~\cite{marschner1998inverse, mait2018computational, faccio2020non}), \etc.
Typically, inverse problem solvers assume the forward model $\forwardFunc$ and its physical parameters $\forwardParam$ are known (\ie, \textit{non-blind})~\cite{schuler2013machine}. However, acquiring accurate forward models is often challenging or impractical in real-world settings. This necessitates solving \textit{blind} inverse problems, where both the hidden signals $\signal$ and the forward model parameters $\forwardParam$ must be jointly estimated. % For example, this involves jointly estimating clean images and blur kernels from blurry images, or recovering 3D objects and their camera parameters from unposed images.

Being heavily ill-posed, inverse problems largely rely on data priors in their computation. 
Traditional supervised learning approaches train an end-to-end neural network to map observations directly to hidden images ($\observation \to \signal$)~\cite{li2020nett, jin2017deep, mccann2017convolutional}. 
Recently, diffusion models (DMs)~\cite{ho2020denoising, song2020score, sohl2015deep} have emerged as powerful inverse problem solvers due to their exceptional ability to model the complex data distribution $\probDistribution{\signal}$ of underlying signals ${\signal}$. 
DMs approximate $\probDistribution{\signal}$ by learning the distribution’s score function $\scoreFunc{\signal_t}{\signal_t}$~\cite{song2019generative}, %\wz{better to use $\scoreFunc{\signal}{\signal}$?}
allowing data-driven priors to be integrated into Bayesian inverse problem solvers (\eg, diffusion posterior sampling~\cite{chung2022diffusion}). %as replacements for conventional handcrafted priors (e.g., total variation, sparsity), facilitating the reconstruction of hidden images in an unsupervised learning manner~\cite{bibid}. 
Later, latent diffusion models (LDMs) have evolved as a new foundational model standard~\cite{rombach2022high} by projecting signals into a lower-dimensional latent space $\latent$ and performing diffusion there. This strategy mitigates the curse of dimensionality typical in pixel-space DMs and demonstrates superior capability, flexibility, and efficiency in modeling complex, high-dimensional distributions, such as those of videos, audio, and 3D objects~\cite{rombach2022high, wang2023audit, stan2023ldm3d, blattmann2023align}.

Although both DM-based and LDM-based solvers have demonstrated impressive posterior sampling performance in diverse computational imaging inverse problems, existing methods predominantly focus on non-blind settings (\ie, optimizing images $\signal$ with known forward model parameters $\forwardParam$)~\cite{chung2022diffusion, rout2024solving, song2023solving}. 
Blind inversion poses more challenges since jointly solving $\signal$ and $\forwardParam$ involves non-convex optimization, often leading to instabilities.
While recent advances have explored the feasibility of solving blind inverse problems using pixel-based DMs~\cite{laroche2024fast, chung2022parallel}, these methods suffer from computational inefficiencies and limited capability in modeling complex image priors, rendering them unsuitable for more challenging, high-dimensional blind inversion tasks like 3D inverse rendering. % \wz{need a better name to refer 3D tasks}

In this paper, we introduce {\name}, a novel approach that solves blind inverse problems using powerful LDM priors.
The core concept of {\name} involves a variational EM framework that alternates between reconstructing underlying images $\signal$ through latent diffusion posterior sampling (E-step) and estimating forward model parameters $\forwardParam$ using the reconstructed images (M-step).  %\wz{why EM? add one sentence}
% expectation-maximization (EM) algorithms with
We further design an annealing optimization strategy  % tailored for the latent space optimization
to enhance the stability of the vulnerable latent space optimization, 
as well as a skip-gradient method to accelerate the training process. % particularly in the early stages when the model parameters may significantly deviate from their true values.
% By incorporating latent priors,
% The proposes techniques significantly reduces the computational burden 
%associated with LDM’s encoder-decoder processes 
% and achieves much more stable and efficient inversion results. 
Consequently, {\name} allows us to leverage the capabilities of pre-trained foundational diffusion models to effectively solve a wide range of blind 2D and 3D inverse problems. 
% By decoupling signal reconstruction and foorward model parameter estimation into separate EM steps, our method achieves much more stable and efficient inversion results. 

%As a general framework, {\name} supports both linear and non-linear inverse problems.  
To the best of our knowledge, {\name} is the first method that incorporates powerful LDM priors~\cite{rombach2022high} in the blind inverse problems. % which allows  us to leverage the capabilities of pre-trained foundational diffusion models to effectively solve a wide range of blind 2D and 3D inverse problems.
% By decoupling image reconstruction and model parameter update into separate EM steps, our methods achieves much more stable and efficient inversion results. 
We first validate our method with Stable Diffusion~\cite{rombach2022high} priors and perform the representative 2D blind motion deblurring task, where we showcase superior imaging quality and efficiency over prior arts. 
{\name} further demonstrates new capabilities in more challenging non-linear 3D inverse rendering problems. Given a set of unposed sparse-view input images, we apply Zero123 priors~\cite{liu2023zero1to3} %in our framework and jointly optimize novel view images plus camera parameters, 
to synthesize the corresponding novel view images, % from unposed sparse-view images,
supporting {pose-free, sparse-view} 3D reconstruction. Our results exhibit more 3D view consistency and achieve new state-of-the-art novel view synthesis performance.

\begin{comment}

\begin{figure}
\centering
\includegraphics[width=13.5cm]{Styles/figure/blindinverse-teaser.pdf}
\centering
\vspace*{-0.3cm}
\caption{\textbf{Representative results of the proposed method.}  
	% We conduct two blind inverse problems, motion deblurring and multi-view consistent 3D modeling without poses, on blurry observations and 2D views, respectively. Notably, our method estimates the kernels and poses accurately. \wz{pick up new images}
	We validate our method on two representative blind inverse problems, 2D blind deblurring (top) and pose-free, spare-view 3D novel view synthesis (bottom).  Notably, our method achieves accurate image and  kernel estimation, as well as consistent novel view images. \wz{pick up new images}
	}
\label{fig:exp-aba-annealing}
\vspace*{-0.5cm}
\end{figure}
\end{comment}

\begin{figure*}
	% \vspace*{-0.5cm}
	\centering
	\setlength{\tabcolsep}{1pt}
	\setlength{\fboxrule}{1pt}
	\resizebox{1.0\textwidth}{!}{
	%\vspace*{1.5cm}
	\begin{tabular}{c}
		\begin{tabular}{cccccccccc}
		& & \small{Observation}  & \small{~\cite{chung2022parallel}} & \small{ Ours } &\small{GT}&  \small{Observation}  & \small{~\cite{chung2022parallel}} & \small{ Ours } &\small{GT}
			\\ 
			\begin{turn}{90} \,\,\,\small{2D Blind} \end{turn} & 	\begin{turn}{90} \,\small{Deblurring} \end{turn} &
			\multicolumn{1}{c}{
				\begin{overpic}[width=0.11\linewidth]{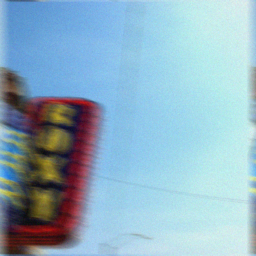}
				\end{overpic}
				}  &
				\multicolumn{1}{c}{
					\begin{overpic}[width=0.11\linewidth]{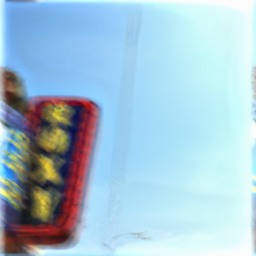}
						\put(65,0){{\includegraphics[width=0.0385\linewidth]{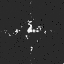}}}
					\end{overpic}
				}  &
				\multicolumn{1}{c}{
					\begin{overpic}[width=0.11\linewidth]{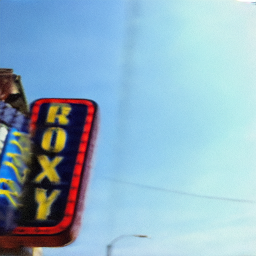}
						\put(65,0){{\includegraphics[width=0.0385\linewidth]{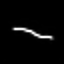}}}
					\end{overpic}
				}  &
				\multicolumn{1}{c}{
					\begin{overpic}[width=0.11\linewidth]{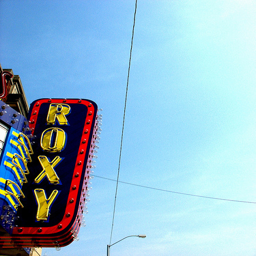}
						\put(65,0){{\includegraphics[width=0.0385\linewidth]{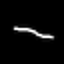}}}
					\end{overpic}
				}  &
				\multicolumn{1}{c}{
					\begin{overpic}[width=0.11\linewidth]{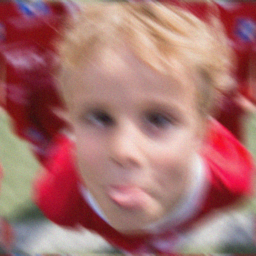}
					\end{overpic}
				}  &
				\multicolumn{1}{c}{
					\begin{overpic}[width=0.11\linewidth]{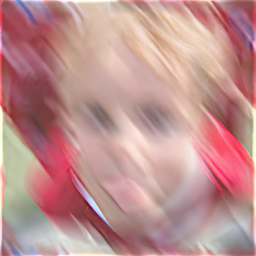}
						\put(65,0){{\includegraphics[width=0.0385\linewidth]{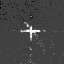}}}
					\end{overpic}
				}  &
				\multicolumn{1}{c}{
					\begin{overpic}[width=0.11\linewidth]{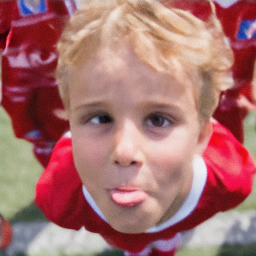}
						\put(65,0){{\includegraphics[width=0.0385\linewidth]{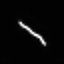}}}
					\end{overpic}
				}  &
				\multicolumn{1}{c}{
					\begin{overpic}[width=0.11\linewidth]{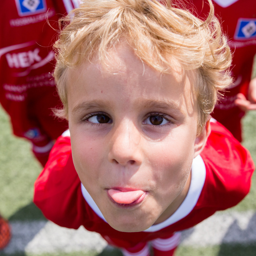}
						\put(65,0){{\includegraphics[width=0.0385\linewidth]{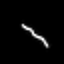}}}
					\end{overpic}
				}  
			 \\
			 \hline
			 \vspace*{-0.3cm}
			 \\
			 & & 
			\small{ Input Views } &  \multicolumn{3}{c}{\small{Synthesized Novel Views}} & \small{ Input Views } &  \multicolumn{3}{c}{\small{Synthesized Novel Views}}  \\
		\begin{turn}{90}{\!\!\!\!\!\!\!\!\!\!\!\!\!\!\small{Sparse-view}} \end{turn}   &
			\begin{turn}{90}\!\!\!\!\!\!\!\!\!\!\!\!\!\!\!\!\!\!\!\!\!\small{3D Reconstruction}\end{turn}  
		&
			 \multicolumn{1}{c}{
			 	\begin{overpic}[width=0.11\linewidth]{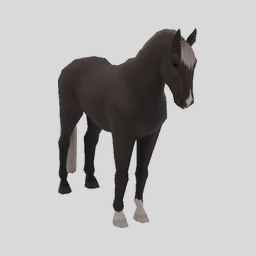}
			 	\end{overpic}
			 }  &
			 \multicolumn{1}{c}{
			 	\begin{overpic}[width=0.11\linewidth]{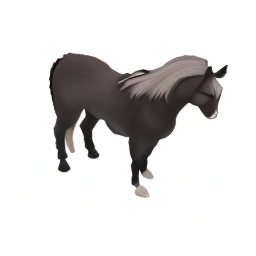}
			 	\end{overpic}
			 }  &
			 \multicolumn{1}{c}{
			 	\begin{overpic}[width=0.11\linewidth]{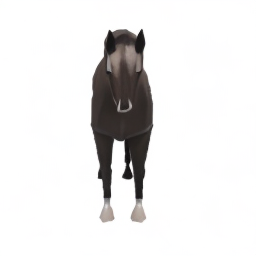}
			 	\end{overpic}
			 }  &
			 \multicolumn{1}{c}{
			 	\begin{overpic}[width=0.11\linewidth]{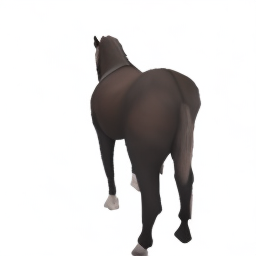}
			 	\end{overpic}
			 }  &
			 \multicolumn{1}{c}{
			 	\begin{overpic}[width=0.11\linewidth]{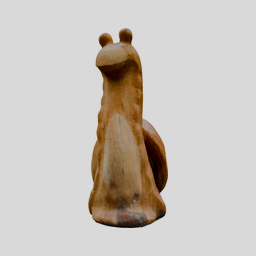}
			 	\end{overpic}
			 }  &
			 \multicolumn{1}{c}{
			 	\begin{overpic}[width=0.11\linewidth]{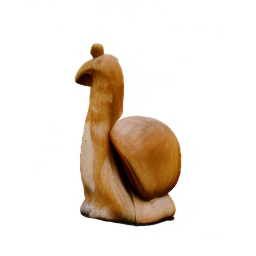}
			 	\end{overpic}
			 }  &
			 \multicolumn{1}{c}{
			 	\begin{overpic}[width=0.11\linewidth]{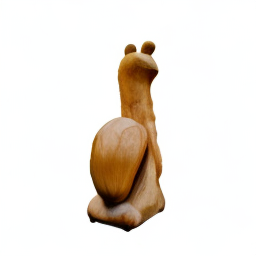}
			 	\end{overpic}
			 }  &
			 \multicolumn{1}{c}{
			 	\begin{overpic}[width=0.11\linewidth]{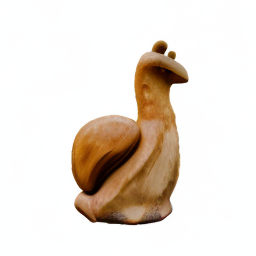}
			 	\end{overpic}
			 }  
			 \\
			  & &
			 \multicolumn{1}{c}{
			 	\begin{overpic}[width=0.11\linewidth]{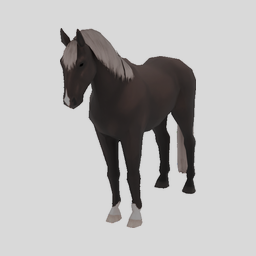}
			 	\end{overpic}
			 }  &
			 \multicolumn{1}{c}{
			 	\begin{overpic}[width=0.11\linewidth]{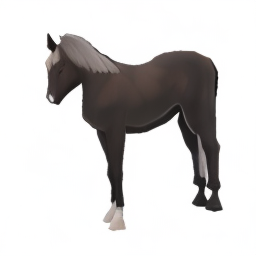}
			 	\end{overpic}
			 }  &
			 \multicolumn{1}{c}{
			 	\begin{overpic}[width=0.11\linewidth]{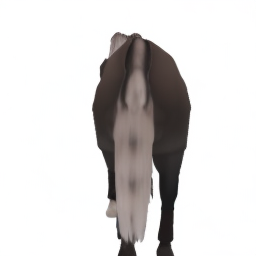}
			 	\end{overpic}
			 }  &
			 \multicolumn{1}{c}{
			 	\begin{overpic}[width=0.11\linewidth]{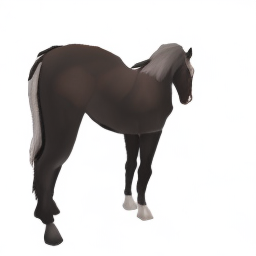}
			 	\end{overpic}
			 }  &
			 \multicolumn{1}{c}{
			 	\begin{overpic}[width=0.11\linewidth]{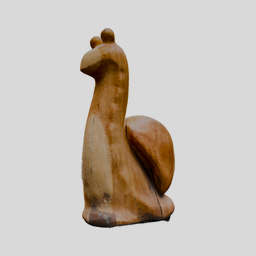}
			 	\end{overpic}
			 }  &
			 \multicolumn{1}{c}{
			 	\begin{overpic}[width=0.11\linewidth]{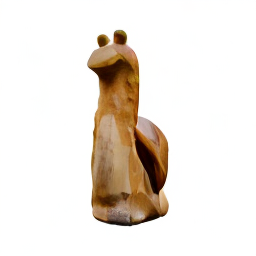}
			 	\end{overpic}
			 }  &
			 \multicolumn{1}{c}{
			 	\begin{overpic}[width=0.11\linewidth]{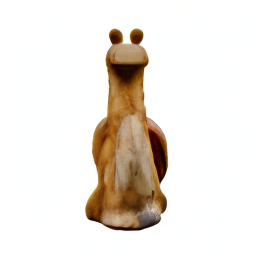}
			 	\end{overpic}
			 }  &
			 \multicolumn{1}{c}{
			 	\begin{overpic}[width=0.11\linewidth]{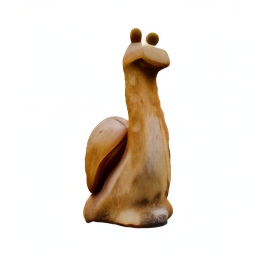}
			 	\end{overpic}
			 }  
		\end{tabular}
	\end{tabular}
}
 \vspace{-0.2cm}
	\caption{
 %\textbf{Representative Blind Inversion Tasks}  
	% We conduct two blind inverse problems, motion deblurring and multi-view consistent 3D modeling without poses, on blurry observations and 2D views, respectively. Notably, our method estimates the kernels and poses accurately. \wz{pick up new images}
	We apply our method on two representative blind inverse problems: \textbf{2D Blind Deblurring} and \textbf{Pose-free Spare-view 3D Reconsturction}.  Notably, in 2D task, our method achieves more accurate image recovery and kernel estimation over BlindDPS~\cite{chung2022parallel}, while in 3D task we successfully reconstruct consistent novel view images from unposed input views.}
\label{fig:exp-aba-annealing}
 \vspace{-0.2cm}
\end{figure*}

\vspace*{-0.2cm}
\section{Related Work}
% \vspace*{-0.2cm}

\textbf{Inverse Problems.} \ \ \ 
The goal of general inverse problems is to recover signals $\signal \in \mathbb{R}^D$ from partial observations $\observation \in \mathbb{R}^M$:
\begin{equation}
	\observation = \forwardFunc_\forwardParam \left( \signal \right) + \boldsymbol{n},
	\label{equ: inverse}
\end{equation}
where $\forwardFunc$,  $\forwardParam$ and $\boldsymbol{n}\sim \mathcal{N}(\mathbf{0}, \sigma^2\mathbf{I})$ represent the forward operator, its parameters, and the observation noise, respectively. 
%
%Given that $\boldsymbol{y}$ often lacks information about the underlying $\boldsymbol{x}$ ($M \le D$) and includes noise $\boldsymbol{n}$, solving inverse problems is inherently ill-posed. 
%
% Inverse problems can be classified into two categories based on the knowledge of $\phi$: \textit{non-blind} and \textit{blind} inverse problems.
%Depending on whether $\phi$ is known, inverse problems can be divided into two categories: \textit{non-blind} and \textit{blind} inverse problems.  
%
%The latter is more challenging but aligns better with real-world scenarios, as acquiring $\forwardParam$ can be expensive. %, \eg, blur kernels or camera parameters.
%\wz{seem repetitive}
%
The signal $\signal$ can be either solved by supervised learning approaches~\cite{li2020nett, jin2017deep, mccann2017convolutional}, or recovered within the Bayesian framework to maximize the posterior: $p(\boldsymbol{x}|\boldsymbol{y}) \propto p(\boldsymbol{x})p(\boldsymbol{y}|\boldsymbol{x})$, where data priors $\probDistribution{\signal}$ are of vital importance.
Traditional methods use handcrafted priors such as sparsity or total variation (TV)~\cite{kuramochi2018superresolution, bouman1993generalized}.
However, these priors cannot capture the complex % nature of 
natural image distributions, limiting the solvers' ability to produce high-quality reconstructions~\cite{danielyan2011bm3d,ulyanov2018deep, candes2007sparsity}.
Recent advances in diffusion models (DMs), particularly latent diffusion models (LDMs), have made them attractive for inverse problems due to their powerful data prior modeling capabilities~\cite{chung2022diffusion, rout2024solving, song2023solving}.
In this paper, we focus on solving \emph{blind} inverse problems using % state-of-the-art 
latent diffusion models~\cite{liu2023zero, rombach2022high}. 

\textbf{Diffusion Models for 2D Inverse Problems.} \ \ \ 
DMs have been applied to a wide range of 2D inverse problems, including natural image deblurring, denoising, and super-resolution tasks~\cite{wang2022zero,chung2022diffusion,chung2022improving,chung2022come,feng2023efficient}, as well as medical and astronomy image enhancement~\cite{adam2022posterior,song2021solving,chung2022score,wang2022zero}. Diffusion Posterior Sampling (DPS) pioneered the use of DMs as strong data priors to solve non-blind 2D inverse problems in a maximum-a-posteriori (MAP) manner~\cite{chung2022diffusion,chung2022improving,chung2022come}. Later works~\cite{rout2024solving, song2023solving} evolved DPS with Latent Diffusion Model (LDM) priors, demonstrating improved performance due to better priors. While these methods~\cite{chung2022diffusion,rout2024solving,song2023solving} all address non-blind problems, BlindDPS~\cite{chung2022parallel} extends DPS to the blind setting by modeling diffusion priors of both data and forward model parameters. Similar to our approach, FastEM~\cite{laroche2024fast} proposes to address blind inversion within an Expectation-Maximization (EM) framework. However,  ~\cite{chung2022parallel,laroche2024fast} remain limited to pixel-based DMs, as the instability of LDMs makes the optimization even harder. In this paper, we investigate how to integrate more powerful LDM priors with EM frameworks in blind inversion tasks and demonstrate new state-of-the-art results.

\textbf{Diffusion Models for 3D Inverse Problems.} \ \ \ 
3D reconstruction from 2D images, also known as inverse graphics, has long been a significant goal in the fields of vision and graphics~\cite{loper2014opendr,chen2019_dibr,mildenhall2020nerf}. 
Recently, diffusion models are also largely involved in tackling this problem~\cite{poole2022dreamfusion,lin2023magic3d,muller2023diffrf,shi2023MVDream,liu2023zero1to3}. 
In this context, the underlying signals $\signal$ and the observation  $\observation$ represent 3D data and 2D images, while $\forwardFunc$ denotes the forward rendering process and $\forwardParam$ are the camera parameters. 
Although the most straightforward way 
%to recover 3D information with DMs 
is to directly model 3D distributions~\cite{muller2023diffrf,zeng2022lion}, this way is not feasible due to the scarcity of 3D data~\cite{shapenet2015,deitke2023objaverse}. Alternatively, recent works focus on utilizing 2D diffusion priors to recover 3D scenes with SDS loss~\cite{poole2022dreamfusion} but suffer from view inconsistency issues~\cite{lin2023magic3d,tang2023make,wang2024prolificdreamer,chen2023fantasia3d}.

To mitigate this problem, a branch of work fine-tunes Latent Diffusion Models (LDMs) with multi-view images, transforming LDMs into conditional renderers~\cite{liu2023zero1to3,shi2023MVDream,tewari2023forwarddiffusion}. Given an image and its camera parameter, they predict the corresponding novel views of the same 3D object. % 
In other words, these models can also be utilized to provide 3D data priors.
However, existing methods typically operate in a feed-forward fashion, still leading accumulated inconsistency during novel view synthesis and requiring further correction designs~\cite{shi2023zero123plus,liu2023one2345,liu2023one2345++}. In contrast, {\name} treats the sparse-view 3D reconstruction~\cite{jiang2022LEAP} task as a blind inverse problem. Given sparse-view input images without knowing their poses, we apply %such a conditional renderer~\cite{liu2023zero1to3}
Zero123~\cite{liu2023zero1to3} priors to jointly optimize their relative camera parameters and synthesize new views. Our method utilizes information of all input views~\cite{song2023solving} and produces significantly better view-consistent objects compared to feed-forward baselines~\cite{liu2023zero1to3,jiang2022LEAP}. % and enables delicate 3D reconstruction results

% allows us to aggregate  significantly reducing the view inconsistency error.

%
%This approach further allows our method to support both single-view and sparse-view input, enabling challenging pose-free sparse-view 3D reconstruction~\cite{jiang2022LEAP}. We demonstrate that our method successfully gathers multi-view information and produces significantly better view-consistent images compared to previous works~\cite{liu2023zero1to3,jiang2022LEAP}, which further enables delicate 3D reconstruction results \wz{not sharp. Depending on final results. Do we support single view?}

% However, Recently, diffusion models becomes a new tools 
\vspace*{-0.2cm}
\section{Preliminary}
% \vspace*{-0.2cm}
% TODO 压缩这部分篇幅 把2和3合起来
\textbf{Diffusion Models and Latent Diffusion Models.} \ \ \ 
% We first briefly review the fundamentals of powerful generative DMs, namely score-based models~\cite{ho2020denoising, song2021scorebased, sohl2015deep}. 
DMs~\cite{ho2020denoising, song2021scorebased, sohl2015deep} model data distribution by learning the time-dependent score function $\scoreFunc{\signal_t}{\signal_t} $ with a parameterized neural networks $\sdModel$.
In the forward step, it progressively injects noise into data through a forward-time SDE; while in the inverse step, it generates data from noise through a reverse-time SDE~\cite{song2021scorebased}:
%Once the weights are learned, $\boldsymbol{s}_\theta$ can be used to generate data from noise through a reverse-time ~\cite{song2021scorebased} - the solution to a forward-time SDE that progressively injects noise into data: \wz{fix sentence}
%
% Specifically, DMs start by progressively injecting Gaussian noise into data, which can be described as a forward-time stochastic differential equation (SDE)~\cite{song2021scorebased}:
% \vspace*{-0.1cm}
\begin{equation}
	\resizebox{0.7\hsize}{!}{$
		\begin{split}
			\text{Forward-time SDE:}  \quad &\finiteNote{\signal}=-\frac{\beta_{\noisetime}}{2} \signal  \finiteNote{\noisetime}+\sqrt{\beta_{\noisetime}} \finiteNote{\wiener}, \\
			\text{Reverse-time SDE:}  \quad &\finiteNote{\signal} =\left[-\frac{\beta_{\noisetime}}{2}\signal-\beta_{\noisetime} \scoreFunc{\signal_\noisetime}{\signal_\noisetime} \right] \finiteNote{\noisetime}+\sqrt{\beta_{\noisetime}} \finiteNote{\overline{\wiener}},
		\end{split}
		$}
	\label{equ:diffusion}
\end{equation}
% \vspace*{-0.1cm}
where $\beta_{\noisetime} \in (0,1)$ is the noise schedule, $\noisetime \in [0,T]$, $\wiener$ and $\overline{\boldsymbol{w}}$ are the standard Wiener process running forward and backward in time, respectively.
This equation is also called variance-preserving SDE (VP-SDE) that equals DDPM~\cite{ho2020denoising}.
Through this paper, %$\tilde{\sigma}$ denotes the reverse diffusion variance as learned in~\cite{dhariwal2021diffusion}, 
we  define $\alpha_{t} := 1-\beta_{t}, \bar{\alpha}_{t} := \prod_{i=1}^{t} \alpha_{i}$ following~\cite{ho2020denoising}, as adopted in Algorithm~\ref{algo:1},~\ref{algo:2}.

%\wz{next para:why LDM is better is not correct?}
% A proper noise schedule perturbs data distribution $\probDistribution{\signal}$ into standard Gaussian distribution $\gaussianFunc{\zeroParam}{\identityParam}$.
% Then, the generative reverse SDE is formulated as:
% \vspace*{-0.1cm}
% \begin{equation}
	%     \finiteNote{\signal} =\left[-\frac{\noiseScheduleFunc{\noisetime}}{2}\signal-\noiseScheduleFunc{\noisetime} \scoreFunc{\signal_\noisetime}{\signal_\noisetime} \right] \finiteNote{\noisetime}+\sqrt{\noiseScheduleFunc{\noisetime}} \finiteNote{\overline{\wiener}},
	% \label{equ:reversesde}
	% \end{equation}
% where the score function $\scoreFunc{\signal_\noisetime}{\signal_\noisetime}$ is approximated by parameterized deep neural networks $\boldsymbol{s}_\theta(\signal_\noisetime, \noisetime)$, and $\overline{\boldsymbol{w}}$ is the standard Wiener process running backward in time. 

% \paragraph{Latent diffusion models.}
A significant drawback of pixel-based DMs is that they require substantial computational resources and a large volume of training data.
To reduce the computation overhead, a generalized family of Latent Diffusion Models (LDMs) are proposed~\cite{rombach2022high, blattmann2023align}.
LDMs embed data into a compressed latent space through $\boldsymbol{z}=\mathcal{E}(\boldsymbol{x})$ for efficiency and flexibility, and decode the latent code $\boldsymbol{z}$ back to the pixel space through $\boldsymbol{x}=\mathcal{D}(\boldsymbol{z})$,  % when necessary,
% \vspace*{-0.1cm}
% \begin{equation}
	%     \boldsymbol{z}=\mathcal{E}(\boldsymbol{x}), \boldsymbol{x}=\mathcal{D}(\boldsymbol{z}),
	% \end{equation}
where $\mathcal{E}: \mathbb{R}^D\to\mathbb{R}^N$ and $\mathcal{D}: \mathbb{R}^N\to\mathbb{R}^D$ are the encoder and decoder, respectively. % and $N\ll D$.
LDMs fuel state-of-the-art foundation models such as Stable Diffusion~\cite{rombach2022high}, which can serve as a powerful cross-domain prior.
% However, LDMs are notorious for their instability due to the vulnerability of latent space.
The versatility of LDMs makes them promising solvers for inverse problems.
However, such an efficient paradigm is a double-edged sword, as LDMs are notorious for their instability due to the vulnerability of latent space~\cite{rout2024solving, chung2023prompt}.

% gluing, data consistency删了
% hat在x和z上，不是在0上
% evolution
\begin{figure*}[tbp]
	\centering
	\includegraphics[width=13.5cm]{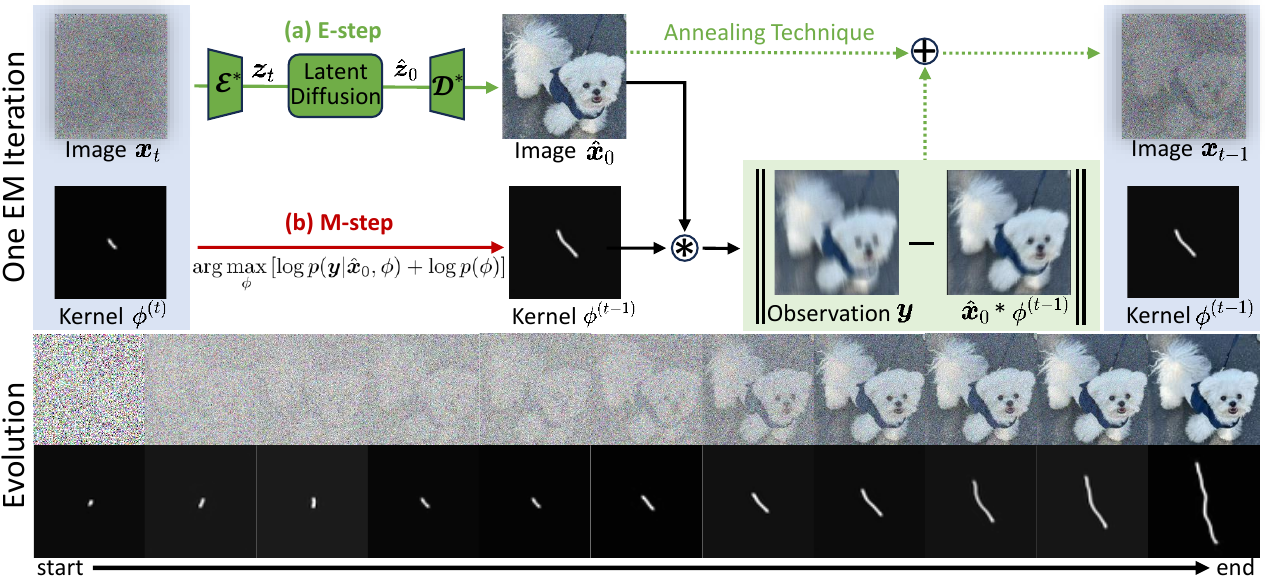}
	\centering
	\caption{\textbf{Overview of {\name}.} \textbf{Top}: 
		%Our expectation-maximum framework consists of two core components: (b1) \textit{E-step}, which draw samples from latent diffusion model and (b2) \textit{M-step} maximum a posterior for forward operator estimation. After random initialization of latent code $\boldsymbol{z}_T$ and forward operator in (a), EM optimization is performed by enforcing \textit{data consistency} and \textit{manifold consistency} in (c) iteratively. Two techniques in (d) are proposed for improvements in effectiveness and efficiency. \textbf{Bottom}: 
		One EM iteration. Given currently estimated data and kernel, in the E-step, we draw new samples with LDM priors with the proposed \emph{annealing technique}. % to stabilize the optimization process.  
		In the M-step we apply maximum-a-posterior (MAP) to update forward parameters.
		\textbf{Middle}: 
		Evolution of the optimized signals and forward parameters.
		% \textbf{Bottom}: 
		% We split E step into \textcolor[HTML]{8DB600}{latent-space}  and \textcolor[HTML]{008000}{pixel-space}  operators plus the \textcolor[HTML]{FF033E}{M-step},  and propose an \emph{accelerating schedule} that runs in \textcolor[HTML]{8DB600}{latent-space}  multiple times then executes the full EM iteration which largely accelerates the training process. %without hurting the performance.
	}
	\label{fig:overview}
	\vspace*{-0.2cm}
\end{figure*}
% TODO tweedie formula去掉
% E-step公式去掉
% M-step公式k换成phi
% 能不能画成EM-score的样子，
% 体现skip gradient，EM scheduling
% 体现出来迭代的过程，加箭头
% 参考cryodrgan

% one em step
% full iterations

\textbf{Diffusion Models for Inverse Problems.} \ \ \
A common approach to apply DM priors in non-blind inverse problems is to replace unconditional score function $\scoreFunc{\signal_\noisetime}{\signal_\noisetime}$ with conditional score function $\scoreFunc{\signal_\noisetime}{\signal_\noisetime | \observation}$ and apply posterior sampling~\cite{song2021solving,chung2022diffusion}.  With the Bayesian rules, we have
\begin{align}
	%\begin{aligned}
		\scoreFunc{\signal_\noisetime}{\signal_\noisetime | \observation} &= \scoreFunc{\signal_\noisetime}{\signal_\noisetime } + \scoreFunc{\signal_\noisetime}{  \observation | \signal_\noisetime} 
  \label{eq:dps1}
  \\
		& \approx \boldsymbol{s}_\theta(\signal_\noisetime, \noisetime)+ \log p(  \observation | \hat{\signal}_0( \signal_\noisetime))
	%\end{aligned}
	\label{eq:dps2}
\end{align}
While $\scoreFunc{\signal_\noisetime}{\signal_\noisetime }$(Eq.~\ref{eq:dps1} middle) can be approximated by diffusion models $\sdModel(\signal_\noisetime, \noisetime)$(Eq.~\ref{eq:dps2} left). However, $\scoreFunc{\signal_\noisetime}{  \observation | \signal_\noisetime} $(Eq.~\ref{eq:dps1} right)   is not tractable as the likelihood  $p_t( \observation | \signal_\noisetime )$ is not known when $\noisetime \neq 0$.  
%Various techniques have been proposed to address this intractable likelihood function, including exactly computing the probability using an ODE flow~\cite{}, bounding the probability through an evidence lower bound (ELBO)~\cite{}, and approximating the probability using Tweedie’s formula~\cite{}. 
% Various techniques have been introduced to estimate this likelihood, such as computing the probability exactly using an ODE flow~\cite{feng2023score}, estimating the probability through evidence lower bound (ELBO)~\cite{feng2023efficient}, and approximating it using Tweedie’s formula~\cite{chung2022diffusion}.
%
% Among different techniques~\cite{feng2023score,feng2023efficient,chung2022diffusion}, 
Following DPS~\cite{chung2022diffusion}, we also assume $p_t( \observation | \signal_\noisetime )=\int_{\signal_0} \probDistribution{ \observation | \signal_0 }\probDistribution{ \signal_0 | \signal_t } \finiteNote{\signal_0} \approx  \probDistribution{ \observation | \hat{\signal}_0(\signal_t) } $, where $ \hat{\signal}_0(\signal_t) = \mathbb{E}\left[ \signal_0 | \signal_t \right]$, which is computational efficient and yields reasonable results. Moreover, we apply the same trick for the latent space sampling as well~\cite{rout2024solving}.

\begin{comment}
	Eq.~\eqref{eq:dps} operate in the pixel-space diffusion models. LatentDPS further extend it into latent sapce by:
	
	\begin{equation}
		\begin{aligned}
			\scoreFunc{\latent_\noisetime}{\latent_\noisetime | \observation} &= \scoreFunc{\latent_\noisetime}{\latent_\noisetime } + \scoreFunc{\latent_\noisetime}{  \observation | \latent_\noisetime} \\
		\end{aligned}
		\label{eq:latentdps}
	\end{equation}
	
	\wz{if we did the same as latentdps,  maybe move equation 7 here.}
\end{comment}

\textbf{Expectation-Maximum Algorithm.} \ \ \
The Expectation-Maximization (EM) algorithm~\cite{dempster1977maximum, gao2021deepgem} is an iterative optimization method used to estimate the parameters $\phi$ of the statistical models that involve underlying variables $\signal$ given the observations $\observation$. When the true values of the underlying variables are unknown, directly applying maximum likelihood estimation (MLE) to recover $\phi$ %on $p_{\phi}( \boldsymbol{y} )$ 
% to recover $\phi$ 
becomes infeasible. 
Instead, the EM algorithm minimizes the Kullback-Leibler (KL) divergence with respect to the model parameters $\phi$ and a variational distribution $q(\signal|\boldsymbol{y})$,
\begin{equation}
	\begin{aligned}
		D_{KL}(q(\signal|\boldsymbol{y})\parallel p_\phi(\signal|\boldsymbol{y}) )
		&= \int q(\signal|\boldsymbol{y}) \log \frac{q(\signal|\boldsymbol{y})p(\boldsymbol{y})}{p_{\phi}(\boldsymbol{y}|\signal)p(\signal)}d\signal \\
		&= \mathbb{E}_{q(\signal|\boldsymbol{y})} \left [ \log q(\signal|\boldsymbol{y}) + \log p(\boldsymbol{y}) - \log p_{\phi}(\boldsymbol{y}|\signal) - \log p(\signal) \right ], 
	\end{aligned}
	\label{eq:em}
\end{equation}
where $p_\phi(\signal | \observation)$ %,
% $\signal$ and $\boldsymbol{y}$ 
denotes the true posterior distribution. %, underlying variables and the observations, respectively. 
To solve this optimization problem, the EM algorithm iterates between optimizing $q(\signal|\boldsymbol{y})$ and $\phi$, known as the Expectation step (E-step) and the Maximization step (M-step), respectively.
In the E-step, $q(\signal|\boldsymbol{y})$ is optimized by sampling underlying variables from $p_\phi(\signal|\boldsymbol{y} )$ assuming known model parameters $\phi$. In the M-step, $\phi$ is optimized using the posterior samples obtained from the E-step. Through this iterative approach, the EM algorithm converges towards a local maximum of the observed data log-likelihood, making it a versatile tool for estimation problems with underlying variables.
% \wz{change variable names, check!}
% \vspace*{-0.1cm}

% theta不对，m-step不是在update prior
% \wz{what's $p_\phi$? explain $p_\theta(x|y), p(x|y)$, \etc.}
% such as Gaussian mixture clustering\cite{reynolds2009gaussian}, dynamical system identification\cite{sun2018identification}, and various other applications.

\vspace*{-0.1cm}
\section{Method}
We now describe how to solve \emph{blind} inverse problems using latent diffusion priors. Our method is formulated within a variational Expectation-Maximization (EM) framework, where the signals and forward operators are iteratively optimized through EM iterations. 
In the E-step (Sec.\ref{sec:E-step}), we leverage latent diffusion priors to draw signal posterior samples, where we introduce an annealing technique to stabilize the optimization process.
In the M-step (Sec.\ref{sec:M-step}), we estimate forward operators in a maximum-a-posteriori (MAP) manner, and adopt a skip-gradient method to improve the efficiency.
%To improve the stability and efficiency of blind inversion, we propose a scheduling method for arranging the execution of the E-steps and M-steps in Sec.\ref{sec:EM framework}. 
In Sec.\ref{sec:app}, we show how our framework can be applied to solve representative problems such as 2D blind deblurring and 3D pose-free sparse-view reconstruction.

\subsection{E-step: Posterior Sampling via Latent Diffusion}
\label{sec:E-step}
The goal of {\name}'s E-step is to solve for the posterior distribution $p_{\phi}(\signal | \observation)$ by leveraging the reverse time SDE in Eq.~\ref{equ:diffusion}. 
To utilize the latent priors, inspired by PSLD~\cite{rout2024solving}, we conduct posterior sampling in the  latent space by defining a conditional latent diffusion process: % where  the  score function $\scoreFunc{\signal_\noisetime}{\signal_\noisetime | \observation}$ is transformed as $\scoreFunc{\latent_\noisetime}{\latent_\noisetime | \observation}$:
	\begin{align}
		\scoreFunc{\latent_\noisetime}{\latent_\noisetime | \observation}
		&=
		\scoreFunc{\latent_\noisetime}{\latent_\noisetime}
		+ 	\scoreFunc{\latent_\noisetime}{ \observation | \latent_t} 
			\label{equ: blindinverse0}
		\\
		% \scoreFunc{\latent_\noisetime}{ \observation \mid \latent_t}
        &\approx \pretrainSdModel{(\latent_\noisetime, \noisetime)} + \nabla_{\latent_\noisetime} \log p_{\phi} ({ \observation | \pretraiedDecodeFunc{\mathbb{E}[{\latent}_{0}| \latent_{t}]} }) 
        	\label{equ: blindinverse00}
        	\\
        &= \pretrainSdModel{(\latent_\noisetime, \noisetime)} - \frac{1}{2\sigma^{2}} \nabla_{\latent_{t}}\left\|\observation-\mathcal{A}_\phi\left( \pretraiedDecodeFunc{\mathbb{E}[{\latent}_{0}| \latent_{t}]}\right)\right\|_{2}^{2}
        % \nabla_{\latent_\noisetime} \log p_{\phi} ({ \observation \mid \pretraiedDecodeFunc{\mathbb{E}[{\latent}_{0}| \latent_{t}]} }),
   %      &\approx
			% \nabla_{\latent_\noisetime} \log p ({ \observation \mid \pretraiedDecodeFunc{\mathbb{E}[{\latent}_{0}| \latent_{t}]} }).
	\label{equ: blindinverse1}
	\end{align}
where $\pretrainSdModel{(\latent_\noisetime, \noisetime)}$ is the pre-trained LDM that approximates the latent score function $\scoreFunc{\latent_\noisetime}{\latent_\noisetime}$, $\mathcal{A}_\phi$ is the parameterized forward model, $\sigma$ is the standard deviation of the additive observation noise, $\decodeNote^*$ is the pre-trained latent decoder, and $\mathbb{E}[{\latent}_{0}| \latent_{t}]$ can be estimated through a reverse time SDE from $\latent_{t}$~\cite{chung2022diffusion}.
%However, simply adopting Eq.~\ref{equ: blindinverse1} does not work, as the latent space to pixel space is not a one$\to$one mapping. 
% In fact, such LDMs for posterior sampling are notorious for the artifacts due to the vulnerability of latent space~\cite{rout2024solving, chung2023prompt}
However, Eq.~\eqref{equ: blindinverse1} works only when $\forwardFunc_\forwardParam$ is known, \ie, non-blind settings~\cite{rout2024solving}.  In the context of blind inversion, $\forwardFunc_\forwardParam$ is randomly initialized so that significant modeling errors perturb the optimization of $\boldsymbol{z}_t$ in the latent space. Consequently, there are significant artifacts when directly applying Eq.~\ref{equ: blindinverse1}. %, as shown in Fig.~\ref{fig:exp-aba-annealing}(b). 
We have to introduce an annealing technique to stabilize the training process:
% suppress the optimization artifacts.
%
\begin{technique}  [Annealing consistency]
	Suppose that the estimated forward operator $\mathcal{A}_\phi$ is optimized from coarse to fine in the iterations, we have that:
    \begin{equation}
		% \resizebox{1.0\hsize}{!}{$
				\scoreFunc{\latent_\noisetime}{\latent_\noisetime | \observation}
    \approx \pretrainSdModel{(\latent_\noisetime, \noisetime)} - \frac{1}{2\zeta_t \sigma^2} \nabla_{\latent_{t}}\left\|\observation-\mathcal{A}_\phi\left( \pretraiedDecodeFunc{\mathbb{E}[{\latent}_{0}| \latent_{t}]}\right)\right\|_{2}^{2} ,
    %$}
			\label{equ: scale-up}
	\end{equation}
 
	% \begin{equation}
	% 	\resizebox{1.0\hsize}{!}{$	\scoreFunc{\latent_\noisetime}{ \observation \mid \latent_t} \approx
	% 		\frac{1}{\zeta} \left[ 	\scoreFunc{\latent_\noisetime}{ \observation \mid \pretraiedDecodeFunc{\mathbb{E}[{\latent}_{0}| \latent_{t}]} } 
	% 		+ \gamma\nabla_{z_{t}}\left\|\mathbb{E}\left[\boldsymbol{z}_{0}|\boldsymbol{z}_{t}\right]-\mathcal{E}^*\left(\mathcal{A}_\phi^{T} \mathcal{A}_\phi \hat{\boldsymbol{x}}_{0}+\left(\boldsymbol{I}-\mathcal{A}_\phi^{T} \mathcal{A}_\phi\right) \mathcal{D}^*\left(\mathbb{E}\left[\boldsymbol{z}_{0}| \boldsymbol{z}_{t}\right]\right)\right)\right\|^{2} \right] ,$}
	% 		\label{equ: scale-up}
	% \end{equation}
	where $\zeta_t$ is a time-dependent factor that decreases over time, \eg, it anneals linearly from 10 at $t=1000$ to 1 at $t=600$ and then holds. 
	\label{technique:annealing}
\end{technique}
We refer to this scaling technique as \textit{Annealing consistency}. Intuitively, $\forwardFunc_\forwardParam$ is randomly initialized at the beginning, which cannot provide correct gradient directions. Therefore, we reduce its influence on the evolution of $\boldsymbol{z}_t$ with a large factor ($\zeta_t=10$). As sampling progresses, $\forwardFunc_\forwardParam$ gradually aligns with the underlying true forward operator. We then anneal the factor ($\zeta_t=1$) to enforce data consistency. We find that this annealing technique is critical for blind inversion with latent priors; without it, the optimized signal $\signal$ consistently exhibits severe artifacts, as shown in Figure~\ref{fig:exp-aba-annealing}. Further theoretical explanations can be found in Appendix~\ref{app:anneal}.

\vspace*{-0.1cm}
\subsection{M-step: Forward Operator Estimation}
\label{sec:M-step}
The goal of  {\name}’s M-step is to update the forward operator parameters $\forwardParam$ with the estimated samples $\hat{\signal}_0$ from the E-step. %, that is, estimate the forward operator $\mathcal{A}_{\phi}$.
This can be achieved by solving a maximum-a-posterior (MAP) estimation problem:
\vspace*{-0.1cm}
\begin{equation}
    \forwardParam^* = \arg \max _{\phi}\mathbb{E}_{\hat{\signal}_0}\left[ \log p_\forwardParam(\observation | \hat{\signal}_0) +\log \probDistribution{\forwardParam}\right] = \arg \min_{\phi} \mathbb{E}_{\hat{\signal}_0}\left[|| \observation-\forwardFunc_\forwardParam(\hat{\signal}_0) ||^2_2  + \mathcal{R}(\phi)\right], 
\label{Equ: map}
\end{equation}
where $\probDistribution{\forwardParam}$ is the prior distribution of $\phi$, $\hat{\boldsymbol{x}}_{0}=\mathcal{D}^*(\hat{\boldsymbol{z}}_{0})=\mathcal{D}^*(\mathbb{E}\left[\boldsymbol{z}_{0}| \boldsymbol{z}_{t}\right])$, and $\mathcal{R}(\forwardParam)$ is an equivalent regularizer. $\mathcal{R}(\forwardParam)$ can include sparsity, patch-based priors, plug-and-play denoisers~\cite{pan2016l_0, sun2013edge}, etc. This MAP estimation problem can be solved using either gradient-based optimization~\cite{laroche2024fast} or neural networks~\cite{chung2022parallel}. Compared to BlindDPS~\cite{chung2022parallel}, which jointly optimizes 
$\signal$ and $\forwardParam$ using two diffusion processes, our method leverages the properties of EM~\cite{gao2021deepgem}, resulting in faster convergence and better performance.

%Typically, Monte Carlo sampling can be applied to draw $N$ samples in the E-step and aggregrate their infomation to solve the best forward operator parameters $\forwardParam^* $.  In practice, we find one sample is enough, thanks to the powerful diffusion process. 
%
% For general blind inverse problems, 
% Our EM framework provides a general framework to update $\forwardParam$ iteratively, depending on the corresponding blind inverse tasks. It supports $\probDistribution{ \observation | \hat{\signal}_0, \forwardParam }$ to be either updated by an implicit neural network approximation~\cite{chung2022parallel} or traditional optimization schemes~\cite{laroche2024fast}. 
%We show $\forwardParam$. different blind inverse tasks and derive different algorithms to update 
% Compared to BlindDPS~\cite{chung2022parallel} which jointly optimize $\signal$ and $\forwardParam$ using two diffusion processes, our method inherits properties of EM~\cite{gao2021deepgem} which is faster and yields better convergence. 

Different from pixel-space diffusion models, latent diffusion models require encoding and decoding operations that map between latent space and pixel space (\ie, $\signal  = \mathcal{D}^*(\latent)$ and $\latent = \mathcal{E}^*(\signal)$), which takes primary time consumption. 
%
% While the encoding and decoding steps are mainly used to provide gradients from the sampled data $\hat{\signal}_0$ and forward parameters $\forwardParam$, again,  
%
% in the early stages of the diffusion posterior sampling process,  $\hat{\signal}_0$ and $\forwardParam$ are far from the true value, making frequent LDM encoding and decoding unnecessary, as they won't provide useful gradient signals. 
%
%
% Besides the annealing technique, 
Therefore, we further design an acceleration method that ``skips'' these operations to improve the efficiency of {\name}.
Specifically, in the whole EM iteration, the E-step comprises two sub-steps: prior-based diffusion ($	\scoreFunc{\latent_\noisetime}{\latent_\noisetime}$) 
and data likelihood-based diffusion ($	\scoreFunc{\latent_\noisetime}{\observation | \latent_\noisetime} $).
% and gluing regularization  %($\nabla_{z_{t}}\left\|\mathbb{E}\left[\boldsymbol{z}_{0}|\boldsymbol{z}_{t}\right]-\mathcal{E}^*\left(\mathcal{A}_\phi^{T} \mathcal{A}_\phi \hat{\boldsymbol{x}}_{0}+\left(\boldsymbol{I}-\mathcal{A}_\phi^{T} \mathcal{A}_\phi\right) \hat{\boldsymbol{x}}_{0}\right)\right\|^{2}$,
(See Eq.~\ref{equ: blindinverse0} for the two terms), where the former happens in latent space($	\scoreFunc{\latent_\noisetime}{\latent_\noisetime}$) while the latter happens in pixel space that requires encoder-decoder operations. Moreover, the M-step % relies on MAP estimation which 
also involves the encoder-decoder operations. We propose to skip these operations to accelerate the training process: % technique to do acceleration:

\begin{technique}  [\text{Skip gradient}]
	In early stages ($t>\skipEnd$), %, where $\skipEnd=800$),
	 performing  $\skipNum$ times E-step in latent space with $	\scoreFunc{\latent_\noisetime}{\latent_\noisetime}$ only, then perform the whole EM-step in both latent space and image space. 
\end{technique}

We refer to this new technique as \textit{Skip gradient}. We find it largely accelerates the training process without hurting the performance for two reasons. %We propose to skip these computations and rely on the latent space diffusion only to accelerate the process for two reasons.  
First, similar to the annealing case, in the early stages of the diffusion posterior sampling process,  the sampled data $\hat{\signal}_0$ and forward parameters $\forwardParam$ are far from the true value, making frequent LDM encoding and decoding unnecessary, as they won't provide useful gradient signals. 
Second, though in the skip-gradient steps we rely on latent diffusion  ($	\scoreFunc{\latent_\noisetime}{\latent_\noisetime}$) only which is unconditional, to some extent, the optimization will follow the previous conditional sampling trajectory, still leading meaningful convergence, as is also observed in ~\cite{song2023solving}. %We refer to this new technique as \textit{Skip gradient}.  

We typically set $\skipEnd=500$ and $\skipNum=8$, which means the total skipped number $M=(T-\skipEnd)(1-1/\skipNum)=(1000-500)(1-1/8)\approx437$ full gradient computation steps. 
We show it significantly reduces computation overhead while keeping PSNR values approximate to the non-skip version, as demonstrated in Table~\ref{tab:skipgrad}. Our full algorithm is described in Algorithm~\ref{algo:1}.

\subsection{Blinding Inversion Tasks}
\label{sec:app}
        \begin{algorithm}[tbp]
            % \scriptsize % Reduce font size to footnotesize
            \begin{algorithmic}
                \caption{LatentDEM for Blind Deblurring}
                \label{algo:1}
                \REQUIRE $T, \boldsymbol{y}, \{\zeta_i\}_{i=1}^T,\{\beta_i\}_{i=1}^T, \{\tilde{\sigma}_i\}_{i=1}^T, \sigma, \delta, \lambda, \mathcal{E}^*, \mathcal{D}^*, \boldsymbol{s}^*_\theta, K, S_T$
                \STATE $\boldsymbol{z}_T \sim \mathcal{N}(\boldsymbol{0}, \boldsymbol{I})$
                \FOR{$t = T$ \textbf{to} $0$}
                \STATE $\boldsymbol{s} \leftarrow \boldsymbol{s}^*_\theta(\boldsymbol{z}_t, t)$
                \STATE $\hat{\boldsymbol{z}}_0 \leftarrow \frac{1}{\sqrt{\bar{\alpha}_t}} (\boldsymbol{z}_t + (1 - \bar{\alpha}_t)\boldsymbol{s})$
                \STATE $\boldsymbol{\epsilon} \sim \mathcal{N}(\boldsymbol{0}, \boldsymbol{I})$
                \STATE $\boldsymbol{z}_{t-1} \leftarrow \frac{\sqrt{\alpha_t}(1-\bar{\alpha}_{t-1})}{1-\bar{\alpha}_t} \boldsymbol{z}_t + \frac{\sqrt{\bar{\alpha}_{t-1}}\beta_t}{1-\bar{\alpha}_t} \hat{\boldsymbol{z}}_0 + \tilde{\sigma}_t \boldsymbol{\epsilon}$
                \IF{$(t>S_T~\AND~t\mid K)~\OR~t<S_T$}
                \STATE $\hat{\boldsymbol{x}}_0 = \mathcal{D}^*(\hat{\boldsymbol{z}}_0)$ 
                \textcolor{blue}{\COMMENT{Skip gradient}}
                \STATE $\forwardFunc_{\phi^{(t-1)}} = \mathrm{M\mbox{-}step}(\boldsymbol{y},\hat{\boldsymbol{x}}_0, \forwardFunc_{\phi^{(t)}},\lambda, \delta)$ 
                \STATE $\boldsymbol{z}_{t-1} \leftarrow \boldsymbol{z}_{t-1} - \frac{1}{2\zeta_t\sigma^2} \nabla_{\boldsymbol{z}_t} \lVert \boldsymbol{y} - \forwardFunc_{\phi^{(t-1)}}(\hat{\boldsymbol{x}}_0) \rVert^2_2$ \textcolor{blue}{\COMMENT{Annealing consistency}}
                \ENDIF
                %\STATE $\boldsymbol{z}_{i-1} \leftarrow \boldsymbol{z}''_{i-1} - \gamma_i \nabla_{\boldsymbol{z}_i} \lVert \tilde{\boldsymbol{z}}_0 - \mathcal{E}^*(\forwardFunc_{i-1}^T \forwardFunc_{i-1} \tilde{\boldsymbol{x}}^*_0 + (\boldsymbol{I} - \forwardFunc_{i-1}^T \forwardFunc_{i-1})\mathcal{D}^*(\tilde{\boldsymbol{z}}_0)) \rVert^2$
                \ENDFOR
                \STATE \textbf{return} $\hat{\boldsymbol{x}}_0, \hat{\forwardFunc}_\phi$
            \end{algorithmic}
        \end{algorithm}
        \vspace*{-0.3cm} % Adds vertical space after the algorithm
    % \end{minipage}
% \end{figure}

Our framework incorporates powerful LDM priors within the EM framework, which enables solving both linear and non-linear inverse problems.  We showcase two representative tasks: the 2D blind debluring task, and the high-dimension, non-linear 3D blind inverse rendering problem. 

\textbf{2D Blind Deblurring.} \ \ \ 
In the blind deblurring task, we aim to jointly estimate the clean image $\image$ and the blurring kernel $\phi$ given a blurry observation $\observation = \mathcal{A}_{\phi}(\image)= \phi * \image$. The LatentDEM approach proceeds as follows:
\begin{itemize}[leftmargin=*]
\item E-step: Assuming a known blurring kernel $\phi$, sample the latent code $\boldsymbol{z}_{t}$ and the corresponding image $\hat{\image}_0^{(t)}=\mathcal{D}(\hat{\latent}_0(\latent_t))$ based on Eq.~\ref{equ: blindinverse1}. To enhance training stability, we adopt the "gluing" regularization \cite{rout2024solving} to address the non-injective nature of the latent-to-pixel space mapping. More discussions about this regularization are shown in Appendix~\ref{app:2d}.
\item M-step: Estimate blur kernels using Half-Quadratic Splitting (HQS) optimization~\cite{geman1995nonlinear, laroche2024fast}:
\begin{equation}
\resizebox{0.8\hsize}{!}{$
    \mathbf{Z}^{(t-1)}=\mathcal{F}^{-1}(\frac{\mathcal{F}(\observation)\mathcal{F}(\hat{\signal}_0^{(t)})+\delta \sigma^2\mathcal{F}(\forwardParam^{(t)})}{\mathcal{F}^{2}(\hat{\signal}_0^{(t)})+\delta \sigma^2}), \ \ \ \ \ 
	\forwardParam^{(t-1)} =\mathbf{D}_{\sqrt{\lambda/\delta }}(\mathbf{Z}^{(t-1)}),
 $}
	\label{eq:hqs2}
\end{equation}
where $\mathbf{Z}$ is a intermediate variable, $\mathbf{D}$ is a Plug-and-Play (PnP) neural denoiser~\cite{laroche2024fast, zhang2017beyond}, $\mathcal{F}$ and $\mathcal{F}^{-1}$ are forward and inverse Fourier transforms, $\sigma$ defines the noise level of measurements, and $\lambda$, $\delta$ are tunable hyperparameters~\cite{zhang2021plug}. The superscripts $^{(t-1)}$ and $^{(t)}$ index diffusion steps. More details on implementation are provided in Appendix~\ref{app:2d}.
\end{itemize}

\textbf{Pose-free Sparse-view 3D Reconstruction.} \ \ \ 
We also demonstrate for the first time that LDM-based blind inversion can be applied to sparse-view, unposed 3D reconstruction, a challenging task that jointly reconstructs the 3D object and camera parameters. Zero123, a conditional LDM, is utilized to approximate the 3D diffusion prior in our task. Given an input image $\observation$ and camera parameters $\phi = (R, T)$ at a target view, Zero123 generates a novel-view image $\hat{\image}_0^{(t)}=\mathcal{D}(\hat{\latent}_0), \hat{\latent}_0=\mathbb{E}[{\latent}_{0}| \latent_{t}]$ through a conditional latent diffusion process $\scoreFunc{\latent_\noisetime}{\latent_\noisetime | \observation, \phi}$. However, current Zero123 is limited to view synthesis and 3D generation from a single image.

By integrating Zero123 into LatentDEM, we can reconstruct a view-consistent 3D object from multiple unposed images. Without loss of generality, we illustrate this with two images $\observation_1$ and $\observation_2$ without knowing their relative pose. The LatentDEM approach becomes: 
\begin{itemize}[leftmargin=*]
\item E-step: Assuming known camera parameters $\phi_1$ and $\phi_2$, aggregate information through a joint latent diffusion process $\scoreFunc{\latent_\noisetime}{\latent_\noisetime | \observation_1, \phi_1, \observation_2, \phi_2}$ to create view-consistent latent codes $\latent_t$ and synthesized image $\hat{\image}_0^{(t)}$.
\item M-step: Estimate camera parameters based on $\hat{\image}_0^{(t)}$ by aligning unposed images to synthetic and reference views via gradient-based optimization:
\begin{equation}
    \phi_2^{(t-1)} = \phi_2^{(t)} - \lambda \nabla_{\phi_2^{(t)}} \|\boldsymbol{z}_{t}(\boldsymbol{y}_2, \phi_2^{(t)}) - \boldsymbol{z}_{t}(\hat{\image}_0^{(t)}, \mathbf{0})\|_2^2 - \delta \nabla_{\phi_2^{(t)}} \|\boldsymbol{z}_{t}(\boldsymbol{y}_2, \phi_2^{(t)}) - \boldsymbol{z}_{t}(\boldsymbol{y}_1, \phi_1)\|_2^2,
	\label{equ: pose estimation}
\end{equation}
where $\boldsymbol{z}_{t}(\cdot, \cdot)$ represents the time-dependent latent features of an image after specified camera transformation, $\mathbf{0}$ indicates no transformation, and $\lambda, \delta$ are tunable hyperparameters. Note only $\phi_2$ is optimized, as $\phi_1$ defines the target view.
\end{itemize}

Through the synthesis of multiple random novel views from input images and subsequent volumetric rendering, we finally generate a comprehensive 3D representation of the object. This approach extends to arbitrary n unposed images, where n-1 camera poses should be estimated. More views yield better 3D generation/reconstruction performance. It outperforms state-of-the-art pose-free sparse-view 3D baselines~\cite{jiang2022LEAP} and generates well-aligned images for detailed 3D modeling~\cite{liu2023one2345}. Further details, including the derivation of view-consistent diffusion process from traditional Zero123 and 3D reconstruction results from various numbers of images, are provided in Appendix~\ref{app:3d}.

\vspace*{-0.1cm}
\section{Experiments}
\vspace*{-0.1cm}

In this section, we first apply our method on the 2D blind deblurring task in Sec.~\ref{sec:re:deblur}. We then demonstrate our method on pose-free, sparse-view 3D reconstruction in Sec.~\ref{sec:re:3drecon}. Lastly, we perform extensive ablation studies to demonstrate the proposed techniques. % different design choices. 
%compare its performance with prior state-of-the-art methods, and perform extensive ablation studies of different design choices. 
% Besides, we demonstrate our method on sparse-view consistent 3D modeling without poses.

% TODO：挑好看一些的samples，能抓眼球
% 再加一列？
\begin{comment}
\begin{figure*}[tbp]
	\centering
	\includegraphics[height=6cm,width=14cm]{Styles/figure/BlindInverse-fig-deblur-result.pdf}
	\centering
	\caption{\textbf{Results of blind deblurring}. (row 1-2): ImageNet 256×256 motion deblurring, (row 3-4): FFHQ 256×256 motion deblurring. (a) Observation, (b) MPRNet~\cite{zamir2021multi}, (c) Self-Deblur~\cite{ren2020neural}, (d) FastEM~\cite{laroche2024fast}, (e) BlindDPS~\cite{chung2022parallel}, (f) LatentEM (Ours),
		(g) Ground truth. For (b), kernels are not shown as the method only estimates images. \wz{refine. ugly. add one more colume}}
	\label{fig:exp-deblurring}
	\vspace*{-0.6cm}
\end{figure*}
\end{comment}

\begin{figure*}
	% \vspace*{-0.5cm}
	\centering
	\setlength{\tabcolsep}{1pt}
	\setlength{\fboxrule}{1pt}
	%\vspace*{1.5cm}
	\begin{tabular}{c}
		\begin{tabular}{ccccccccc}
			& 
			\tiny{Observation}  & 
			\tiny{MPRNet~\cite{chung2022parallel}} & 
			\tiny{ DeblurGAN~\cite{kupyn2019deblurgan} } &
			\tiny{Self-Deblur~\cite{ren2020neural}}& 
			\tiny{FastEM~\cite{laroche2024fast}}  & 
			\tiny{BlindDPS~\cite{chung2022parallel}} & 
			\tiny{ Ours } &
			\tiny{GT}
			\\ 
			\begin{turn}{90} \!\!\! \!\!\! \!\!\! \!\!\! \!\!\! \!\!\!\small{ImageNet} \end{turn} & 
			\multicolumn{1}{c}{
				\begin{overpic}[width=0.11\linewidth]{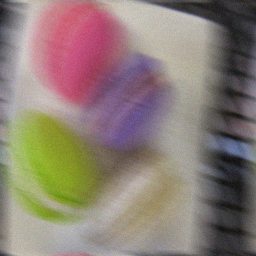}
				\end{overpic}
			}  &
			\multicolumn{1}{c}{
				\begin{overpic}[width=0.11\linewidth]{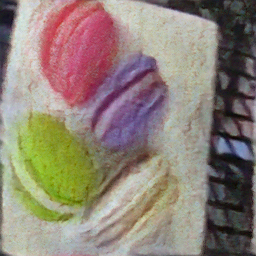}
				\end{overpic}
			}  &
			\multicolumn{1}{c}{
				\begin{overpic}[width=0.11\linewidth]{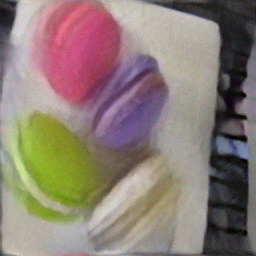}
				\end{overpic}
			}  &
			\multicolumn{1}{c}{
				\begin{overpic}[width=0.11\linewidth]{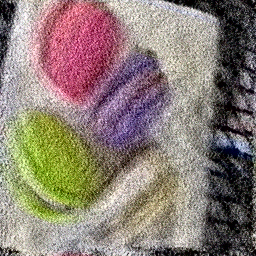}
					\put(65,0){{\includegraphics[width=0.0385\linewidth]{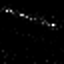}}}
				\end{overpic}
			}  &
			\multicolumn{1}{c}{
				\begin{overpic}[width=0.11\linewidth]{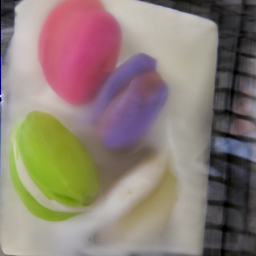}
					\put(65,0){{\includegraphics[width=0.0385\linewidth]{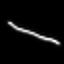}}}
				\end{overpic}
			}  &
			\multicolumn{1}{c}{
				\begin{overpic}[width=0.11\linewidth]{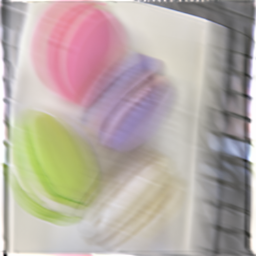}
					\put(65,0){{\includegraphics[width=0.0385\linewidth]{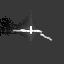}}}
				\end{overpic}
			}  &
			\multicolumn{1}{c}{
				\begin{overpic}[width=0.11\linewidth]{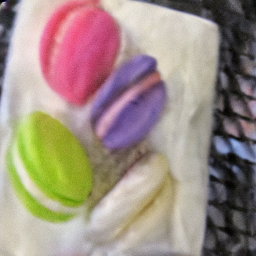}
					\put(65,0){{\includegraphics[width=0.0385\linewidth]{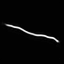}}}
				\end{overpic}
			}  &
			\multicolumn{1}{c}{
				\begin{overpic}[width=0.11\linewidth]{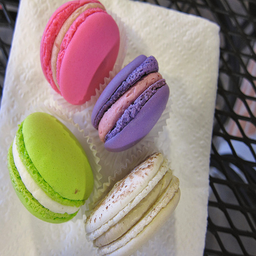}
					\put(65,0){{\includegraphics[width=0.0385\linewidth]{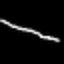}}}
				\end{overpic}
			} 
			\\
			  & 
			  \multicolumn{1}{c}{
			  	\begin{overpic}[width=0.11\linewidth]{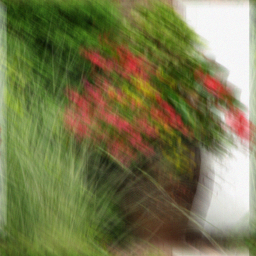}
			  	\end{overpic}
			  }  &
			  \multicolumn{1}{c}{
			  	\begin{overpic}[width=0.11\linewidth]{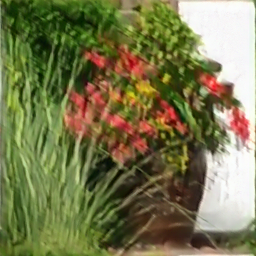}
			  	\end{overpic}
			  }  &
			  \multicolumn{1}{c}{
			  	\begin{overpic}[width=0.11\linewidth]{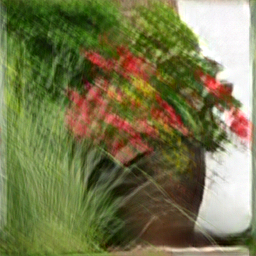}
			  	\end{overpic}
			  }  &
			  \multicolumn{1}{c}{
			  	\begin{overpic}[width=0.11\linewidth]{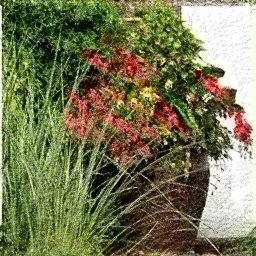}
			  		\put(65,0){{\includegraphics[width=0.0385\linewidth]{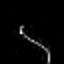}}}
			  	\end{overpic}
			  }  &
			  \multicolumn{1}{c}{
			  	\begin{overpic}[width=0.11\linewidth]{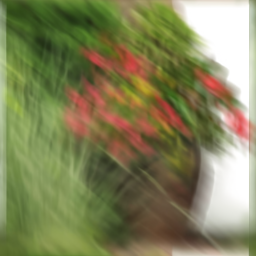}
			  		\put(65,0){{\includegraphics[width=0.0385\linewidth]{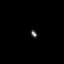}}}
			  	\end{overpic}
			  }  &
			  \multicolumn{1}{c}{
			  	\begin{overpic}[width=0.11\linewidth]{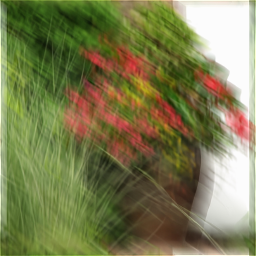}
			  		\put(65,0){{\includegraphics[width=0.0385\linewidth]{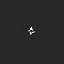}}}
			  	\end{overpic}
			  }  &
			  \multicolumn{1}{c}{
			  	\begin{overpic}[width=0.11\linewidth]{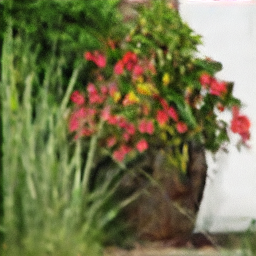}
			  		\put(65,0){{\includegraphics[width=0.0385\linewidth]{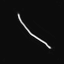}}}
			  	\end{overpic}
			  }  &
			  \multicolumn{1}{c}{
			  	\begin{overpic}[width=0.11\linewidth]{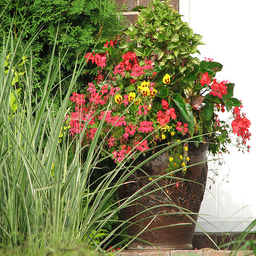}
			  		\put(65,0){{\includegraphics[width=0.0385\linewidth]{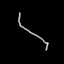}}}
			  	\end{overpic}
			  } 
			  \\
			  \begin{turn}{90} \!\!\! \!\!\! \!\!\! \!\!\! \!\!\! \small{FFHQ} \end{turn} & 
			  \multicolumn{1}{c}{
			  	\begin{overpic}[width=0.11\linewidth]{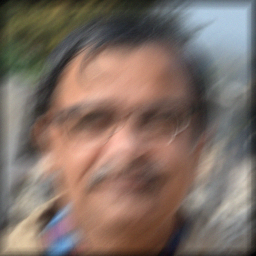}
			  	\end{overpic}
			  }  &
			  \multicolumn{1}{c}{
			  	\begin{overpic}[width=0.11\linewidth]{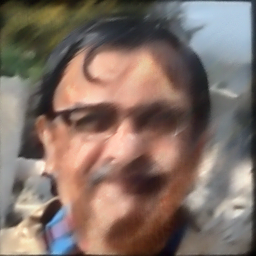}
			  	\end{overpic}
			  }  &
			  \multicolumn{1}{c}{
			  	\begin{overpic}[width=0.11\linewidth]{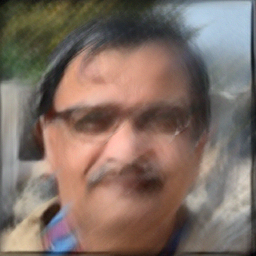}
			  	\end{overpic}
			  }  &
			  \multicolumn{1}{c}{
			  	\begin{overpic}[width=0.11\linewidth]{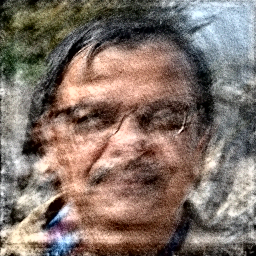}
			  		\put(65,0){{\includegraphics[width=0.0385\linewidth]{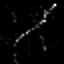}}}
			  	\end{overpic}
			  }  &
			  \multicolumn{1}{c}{
			  	\begin{overpic}[width=0.11\linewidth]{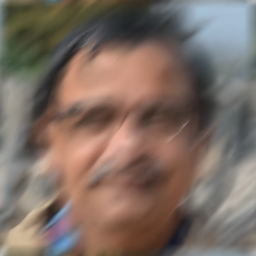}
			  		\put(65,0){{\includegraphics[width=0.0385\linewidth]{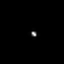}}}
			  	\end{overpic}
			  }  &
			  \multicolumn{1}{c}{
			  	\begin{overpic}[width=0.11\linewidth]{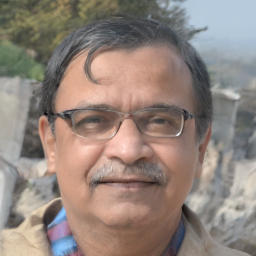}
			  		\put(65,0){{\includegraphics[width=0.0385\linewidth]{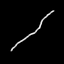}}}
			  	\end{overpic}
			  }  &
			  \multicolumn{1}{c}{
			  	\begin{overpic}[width=0.11\linewidth]{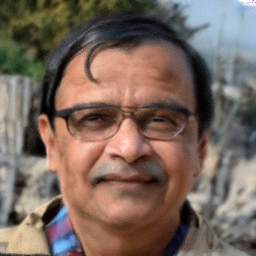}
			  		\put(65,0){{\includegraphics[width=0.0385\linewidth]{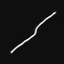}}}
			  	\end{overpic}
			  }  &
			  \multicolumn{1}{c}{
			  	\begin{overpic}[width=0.11\linewidth]{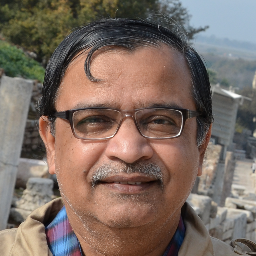}
			  		\put(65,0){{\includegraphics[width=0.0385\linewidth]{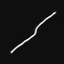}}}
			  	\end{overpic}
			  } 
			  \\
			  & 
			  \multicolumn{1}{c}{
			  	\begin{overpic}[width=0.11\linewidth]{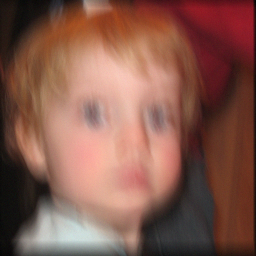}
			  	\end{overpic}
			  }  &
			  \multicolumn{1}{c}{
			  	\begin{overpic}[width=0.11\linewidth]{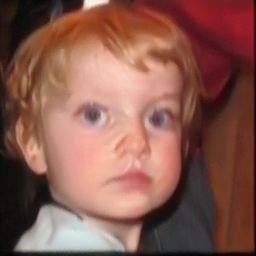}
			  	\end{overpic}
			  }  &
			  \multicolumn{1}{c}{
			  	\begin{overpic}[width=0.11\linewidth]{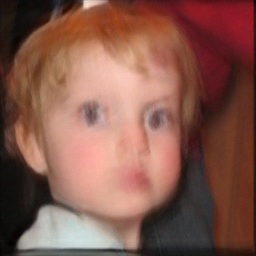}
			  	\end{overpic}
			  }  &
			  \multicolumn{1}{c}{
			  	\begin{overpic}[width=0.11\linewidth]{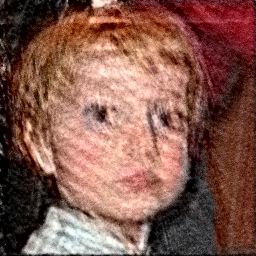}
			  		\put(65,0){{\includegraphics[width=0.0385\linewidth]{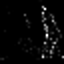}}}
			  	\end{overpic}
			  }  &
			  \multicolumn{1}{c}{
			  	\begin{overpic}[width=0.11\linewidth]{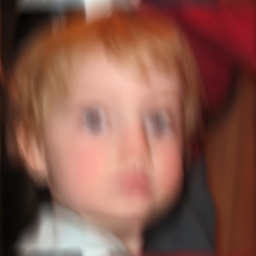}
			  		\put(65,0){{\includegraphics[width=0.0385\linewidth]{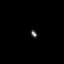}}}
			  	\end{overpic}
			  }  &
			  \multicolumn{1}{c}{
			  	\begin{overpic}[width=0.11\linewidth]{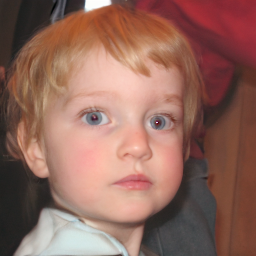}
			  		\put(65,0){{\includegraphics[width=0.0385\linewidth]{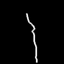}}}
			  	\end{overpic}
			  }  &
			  \multicolumn{1}{c}{
			  	\begin{overpic}[width=0.11\linewidth]{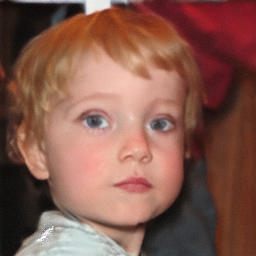}
			  		\put(65,0){{\includegraphics[width=0.0385\linewidth]{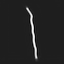}}}
			  	\end{overpic}
			  }  &
			  \multicolumn{1}{c}{
			  	\begin{overpic}[width=0.11\linewidth]{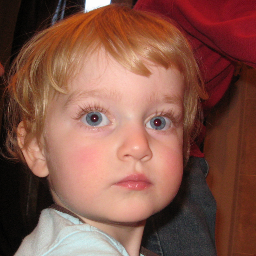}
			  		\put(65,0){{\includegraphics[width=0.0385\linewidth]{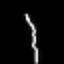}}}
			  	\end{overpic}
			  } 
		\end{tabular}
	\end{tabular}
	\vspace{-0.1cm}
\caption{\textbf{ Blind motion deblurring results}. Row (1-2): ImageNet. Row (3-4): FFHQ.  Our method successfully recovers clean images and accurate blur kernels, consistently outperforming all the baselines, even under challenging cases where the observations are severely degraded. }
\label{fig:exp-deblurring}
\vspace*{-0.2cm}
\end{figure*}

% TODO：挑好看一些的samples，能抓眼球
% 再加一列？
\begin{comment}
\begin{figure*}[tbp]
	\centering
	\includegraphics[height=6cm,width=14cm]{Styles/figure/BlindInverse-fig-deblur-result.pdf}
	\centering
	\caption{\textbf{Results of blind deblurring}. (row 1-2): ImageNet 256×256 motion deblurring, (row 3-4): FFHQ 256×256 motion deblurring. (a) Observation, (b) MPRNet~\cite{zamir2021multi}, (c) Self-Deblur~\cite{ren2020neural}, (d) FastEM~\cite{laroche2024fast}, (e) BlindDPS~\cite{chung2022parallel}, (f) LatentEM (Ours),
		(g) Ground truth. For (b), kernels are not shown as the method only estimates images. \wz{refine. ugly. add one more colume}}
	\label{fig:exp-deblurring}
	\vspace*{-0.6cm}
\end{figure*}
\end{comment}

\subsection{2D Blind Motion Deblurring}
\label{sec:re:deblur}

\begin{comment}

\paragraph{Dataset.} \ \ \ 
We evaluate our method on the images from ImageNet~\cite{bibid} and FFHQ~\cite{bibid}. Following xx~\cite{bibid}, we randomly choose 64 validation images of the widely used ImageNet $256\times256$ and FFHQ $256\times256$ to perform the blind deblurring task.
For pixel-space DMs, we leverage pre-trained priors as in the experiment setting of \cite{chung2022parallel}.
For latent-space LDMs, Stable Diffusion v-1.5 model is chosen as our cross-domain prior.
We also test our algorithm on LDM-VQ-4, which is an LDM before Stable Diffusion, thus it does
not match the performance of Stable Diffusion.
For quantitative comparison, we adopt the following three metrics: peak signal-to-noise-ratio (PSNR), structural similarity index (SSIM), and learned perceptual image patch similarity (LPIPS). 
Besides, we use mean-squared error (MSE), and maximum of normalized convolution (MNC)~\cite{hu2012good} for kernel estimation. 
which is computed by $\operatorname{MNC}:=\max \left(\frac{\tilde{\boldsymbol{k}} * \boldsymbol{k}^{*}}{\|\tilde{\boldsymbol{k}}\|_{2}\left\|\boldsymbol{k}^{*}\right\|_{2}}\right)$
where $\tilde{\boldsymbol{k}}$, $\boldsymbol{k}^{*}$ are the estimated, and the ground truth kernels, respectively.
All experiments are done on NVIDIA A100 80GB GPUs.
\end{comment}

\paragraph{Dataset.} \ 
We evaluate our method on the images from the widely used ImageNet~\cite{imagenet_cvpr09} and FFHQ~\cite{karras2019stylebased}. We randomly choose 64 validation images from each dataset, where the resolutions are both $256\times256$. % of the widely used ImageNet $256\times256$ and FFHQ $256\times256$ to perform the blind deblurring task.
% For our method, 
We chose to use the state-of-the-art Stable Diffusion v-1.5 model~\cite{rombach2022high} as our cross-domain prior.
% Moreover, our baselines contains BlindDPS~\cite{chung2022parallel} and FastEM~\cite{laroche2024fast} which utlize pixel-space diffusion models. We apply xx model as stated in BlindDPS paper~\cite{bibid}.  
% We compare our method with 
% For pixel-space DMs, we leverage pre-trained priors as in the experiment setting of \cite{chung2022parallel}.
% For latent-space LDMs, Stable Diffusion v-1.5 model is chosen as our cross-domain prior.
% We also test our algorithm on LDM-VQ-4, which is an LDM before Stable Diffusion, thus it does
% not match the performance of Stable Diffusion.
For quantitative comparison, we evaluate the image quality with three metrics: peak signal-to-noise-ratio (PSNR), structural similarity index (SSIM), and learned perceptual image patch similarity (LPIPS). 
Besides, we assess the estimated kernel via mean-squared error (MSE), and maximum of normalized convolution (MNC)~\cite{hu2012good},
All experiments are done on NVIDIA A100 80GB GPUs.

\begin{table}[htbp]
  \centering
  \caption{Quantitative evaluation (PSNR, SSIM, LPIPS) of blind deblurring task and (MSE, MNC) of kernel estimation on ImageNet and FFHQ. \textbf{Bold}: Best, \underline{under}: second best.}
  \setlength{\tabcolsep}{0.1mm}{
    \begin{tabular}{ccccccccccc}
    \toprule
    \multirow{3}[4]{*}{\textbf{Method}} & \multicolumn{5}{c}{\textbf{ImageNet ($\mathbf{256}\times \mathbf{256}$)}} & \multicolumn{5}{c}{\textbf{FFHQ ($\mathbf{256}\times \mathbf{256}$)}} \\
\cmidrule{2-11}          & \multicolumn{3}{c}{\textbf{Image}} & \multicolumn{2}{c}{\textbf{Kernel}} & \multicolumn{3}{c}{\textbf{Image}} & \multicolumn{2}{c}{\textbf{Kernel}} \\
          & PSNR $\uparrow$  & SSIM $\uparrow$  & LPIPS $\downarrow $ & MSE $\downarrow $  & MNC $\uparrow$  & PSNR $\uparrow$ & SSIM $\uparrow$ & LPIPS $\downarrow $ & MSE $\downarrow $  & MNC $\uparrow$\\
    \midrule
    MPRNet & \textbf{19.85} & 0.433 & 0.470 & - & - & {21.60} & 0.517 & 0.399 & - & - \\
    Self-Deblur & 16.74 & 0.232 & 0.493 & \underline{0.016} & 0.036 & 18.84 & 0.328 & 0.493 & \underline{0.017} & 0.045 \\
    BlindDPS & 17.31 & \underline{0.472} & \underline{0.309} & 0.036 & \underline{0.274} & \underline{22.58} & \underline{0.583} & 0.245 & 0.048 & 0.270 \\
    FastEM & {17.36} & 0.422 & 0.377 & 0.440 & 0.266 & 17.46 & 0.554 & \underline{0.169} & {0.035} & \underline{0.399} \\
    \textbf{Ours} & \underline{19.35} & \textbf{0.496} & \textbf{0.256} & \textbf{0.010} & \textbf{0.441} & \textbf{22.65} & \textbf{0.653} & \textbf{0.167} & \textbf{0.009} & \textbf{0.459} \\
    \bottomrule
    \end{tabular}}
  \label{tab:deblurring}
  \vspace*{-0.2cm}
\end{table}

\vspace*{-0.2cm}
\paragraph{Results.} We provide motion deblurring results in Figure~\ref{fig:exp-deblurring} and Table~\ref{tab:deblurring}.
Our method is compared with two state-of-the-art methods that directly apply pixel-space diffusion models for blind deblurring: BlindDPS~\cite{chung2022parallel} and FastEM~\cite{laroche2024fast}, and three widely applied methods: MPRNet~\cite{zamir2021multi}, DeblurGAN V2~\cite{kupyn2019deblurgan}, and Self-Deblur~\cite{ren2020neural}.
% We observe that our method and BlindDPS outperform other baselines significantly, as we impose a rather aggressive degradation with severe motion blur. 
%
Several interesting observations can be found here.  
First, {\name} outperforms all the baselines qualitatively.  As shown in Fig.~\ref{fig:exp-deblurring}, in challenging cases with severe motion blur and aggressive image degradation, previous methods are unable to accurately estimate the kernel, while the proposed method enables accurate kernel estimation and high-quality image restoration.
We attribute this to the fact that the powerful LDM priors provide better guidance than pixel-space DM priors in the posterior sampling, together with the deliberately designed EM optimization policies.
% Notably, unlike pixel-space DMs that are mostly limited to a single domain, foundation LDMs such as Stable Diffusion can serve as a cross-domain prior, solving inverse problems in a fully general domain. \wz{what does this mean?}
%
Moreover, as shown in Table~\ref{tab:deblurring}, {\name} also achieves the best scores in most metric evaluations, especially for kernel estimation, which demonstrates the efficacy of our EM framework. Interestingly,  MPRNet shows higher PSNR in ImageNet dataset but visually it produces smooth and blurry results, which indicates the quality of deblurring cannot be well-reflected by PSNR. Nevertheless, we still largely outperform it in SSIM and LPIPS metrics. % than all the baselines 
% Therefore, LatentDEM outperforms BlindDPS and FastDEM easily on ImageNet, as the distribution of ImageNet is much more complex than FFHQ, and pixel-space prior provided by~\cite{chung2022parallel} is poorer than our latent-space prior. 

\vspace*{-0.1cm}
\subsection{Experiments on Pose-free Sparse-view  3D Reconstruction}
\label{sec:re:3drecon}

\paragraph{Dataset.} \
We evaluate the pose-free sparse-view 3D reconstruction performance on Objaverse dataset~\cite{deitke2023objaverse}, which contains millions of high-quality 3D objects. We pick up 20 models and for each model, we randomly render two views without knowing their poses.
Our goal is to synthesize novel views and reconstruct the underlying 3D model from the unposed sparse-view inputs, which is a very challenging task~\cite{jiang2022LEAP} and cannot be easily addressed by NeRF~\cite{mildenhall2020nerf} or Gaussian Splatting~\cite{kerbl3Dgaussians} that require image poses. 

\begin{figure*}[t]
	% \vspace*{-0.5cm}
	\centering
	\setlength{\tabcolsep}{1pt}
	\setlength{\fboxrule}{1pt}
	%\vspace*{1.5cm}
	\resizebox{1.0\textwidth}{!}{
	\begin{tabular}{c}
		\begin{tabular}{cc|ccc|cccc} 
			\multicolumn{2}{c}{Input Views} &
			\multicolumn{3}{c}{ Novel Views of Our Method} &
			\multicolumn{3}{c}{Novel Views of Baselines} &
			\\ 
			\multicolumn{1}{c}{
				\begin{overpic}[width=0.11\linewidth]{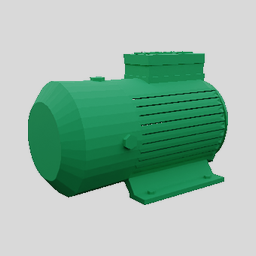}
				\end{overpic}
			}  &
			\multicolumn{1}{c|}{
				\begin{overpic}[width=0.11\linewidth]{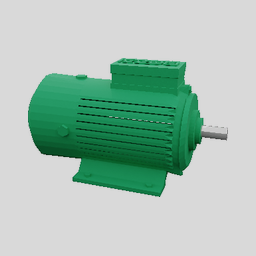}
				\end{overpic}
			}  &
			\multicolumn{1}{c}{
				\begin{overpic}[width=0.11\linewidth]{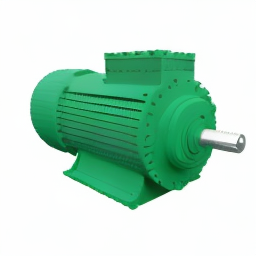}
				\end{overpic}
			}  &
			\multicolumn{1}{c}{
				\begin{overpic}[width=0.11\linewidth]{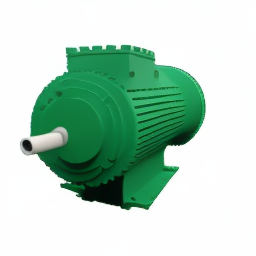}
				\end{overpic}
			}  &
			\multicolumn{1}{c|}{
				\begin{overpic}[width=0.11\linewidth]{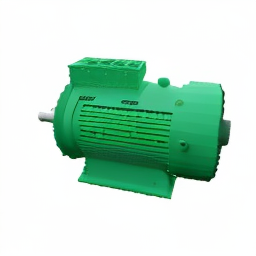}
				\end{overpic}
			}  &
			\multicolumn{1}{c}{
				\begin{overpic}[width=0.11\linewidth]{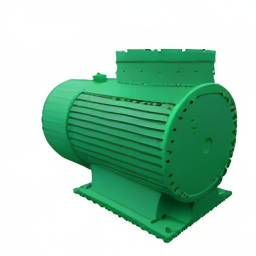}
				\end{overpic}
			}  &
			\multicolumn{1}{c}{
				\begin{overpic}[width=0.11\linewidth]{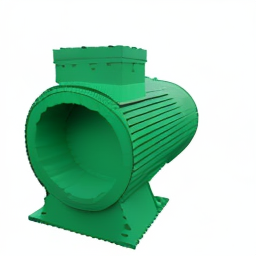}
				\end{overpic}
			}  &
			\multicolumn{1}{c}{
				\begin{overpic}[width=0.11\linewidth]{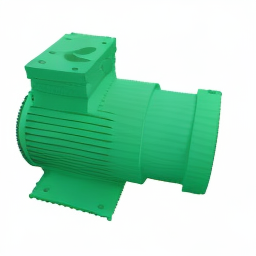}
				\end{overpic}
			}  & 
                \begin{turn}{90} \small{\,\,\,Zero123~\cite{liu2023zero1to3}} \end{turn}
			\\
			\multicolumn{1}{c}{
				\begin{overpic}[width=0.11\linewidth]{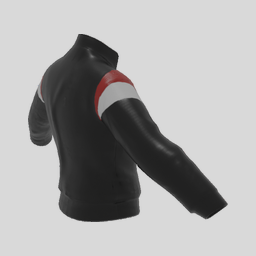}
				\end{overpic}
			}  &
			\multicolumn{1}{c|}{
				\begin{overpic}[width=0.11\linewidth]{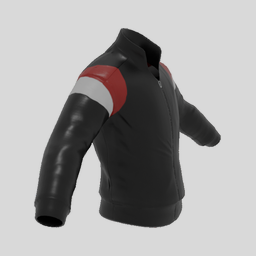}
				\end{overpic}
			}  &
			\multicolumn{1}{c}{
				\begin{overpic}[width=0.11\linewidth]{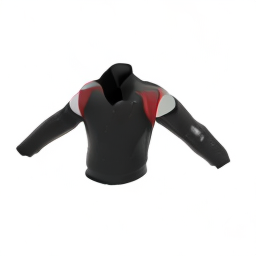}
				\end{overpic}
			}  &
			\multicolumn{1}{c}{
				\begin{overpic}[width=0.11\linewidth]{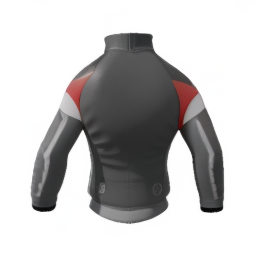}
				\end{overpic}
			}  &
			\multicolumn{1}{c|}{
				\begin{overpic}[width=0.11\linewidth]{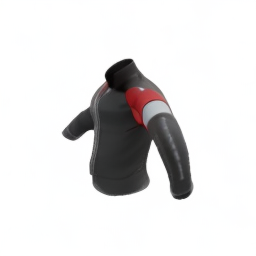}
				\end{overpic}
			}  &
			\multicolumn{1}{c}{
				\begin{overpic}[width=0.11\linewidth]{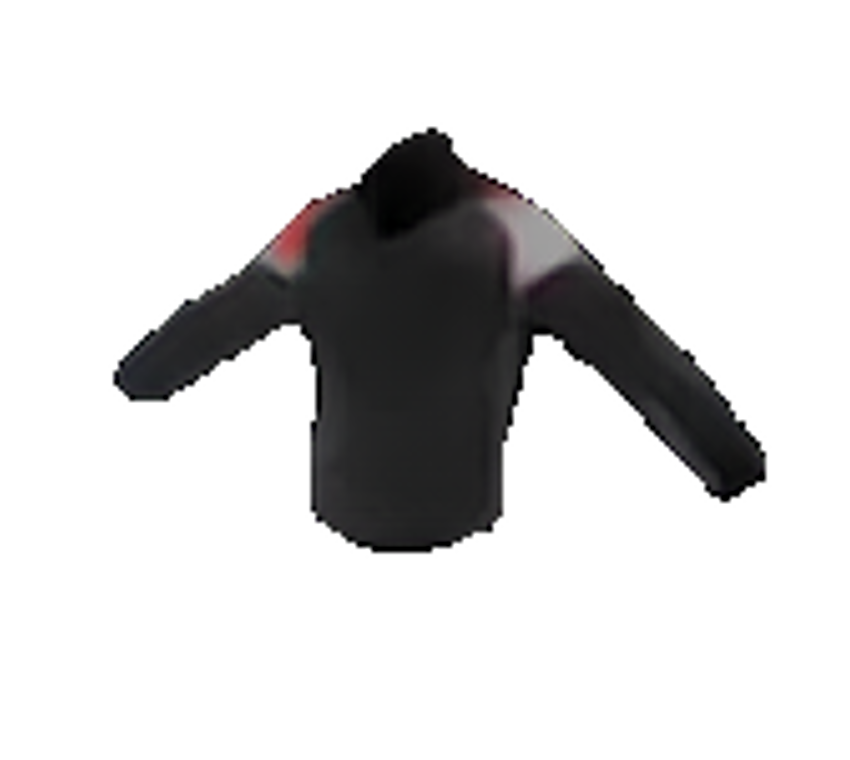}
				\end{overpic}
			}  &
			\multicolumn{1}{c}{
				\begin{overpic}[width=0.11\linewidth]{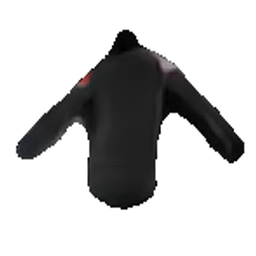}
				\end{overpic}
			}  &
			\multicolumn{1}{c}{
				\begin{overpic}[width=0.11\linewidth]{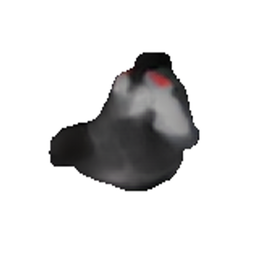}
				\end{overpic}
			}  & \begin{turn}{90} \small{\,\,LEAP~\cite{jiang2022LEAP}} \end{turn}
		\end{tabular}
	\end{tabular}
}
		\vspace*{-0.1cm}
	\caption{\textbf{Pose-free sparse-view 3D reconstruction results.} Our method successfully synthesizes consistent novel view images given two sparse input views. In contrast,  Zero123~\cite{liu2023zero} produces images missing the engine handle that are not consistent with the input views, while LEAP~\cite{jiang2022LEAP} fails to generate photo-realistic images. %\wz{replace new ims}
 }
	\label{fig:exp-3D}
	\vspace*{-0.1cm}
\end{figure*}

% 不要用abcdefg，不好看。在caption中说清楚
% c-f只需要present一个最好的就行，其他的放到appendix里
\begin{comment}
\begin{figure*}[tbp]
	\centering
	\includegraphics[width=14cm]{Styles/figure/BlindInverse-exp-aba-annealing.pdf}
	\centering
	\caption{\textbf{Effectiveness of our annealed consistency technique.} We evaluate different annealing schemes on blind deblurring. (a) Observation, (b) vanilla LatentDEM without annealing, (c-f) improved LatentDEM with different annealing schemes,
		(g) Ground truth.}
	\label{fig:exp-aba-annealing}
\end{figure*}

\end{comment}

\begin{figure*}
\begin{minipage}[t]{0.55\textwidth}
\begin{tabular}{c}
 \includegraphics[width=0.98\linewidth]{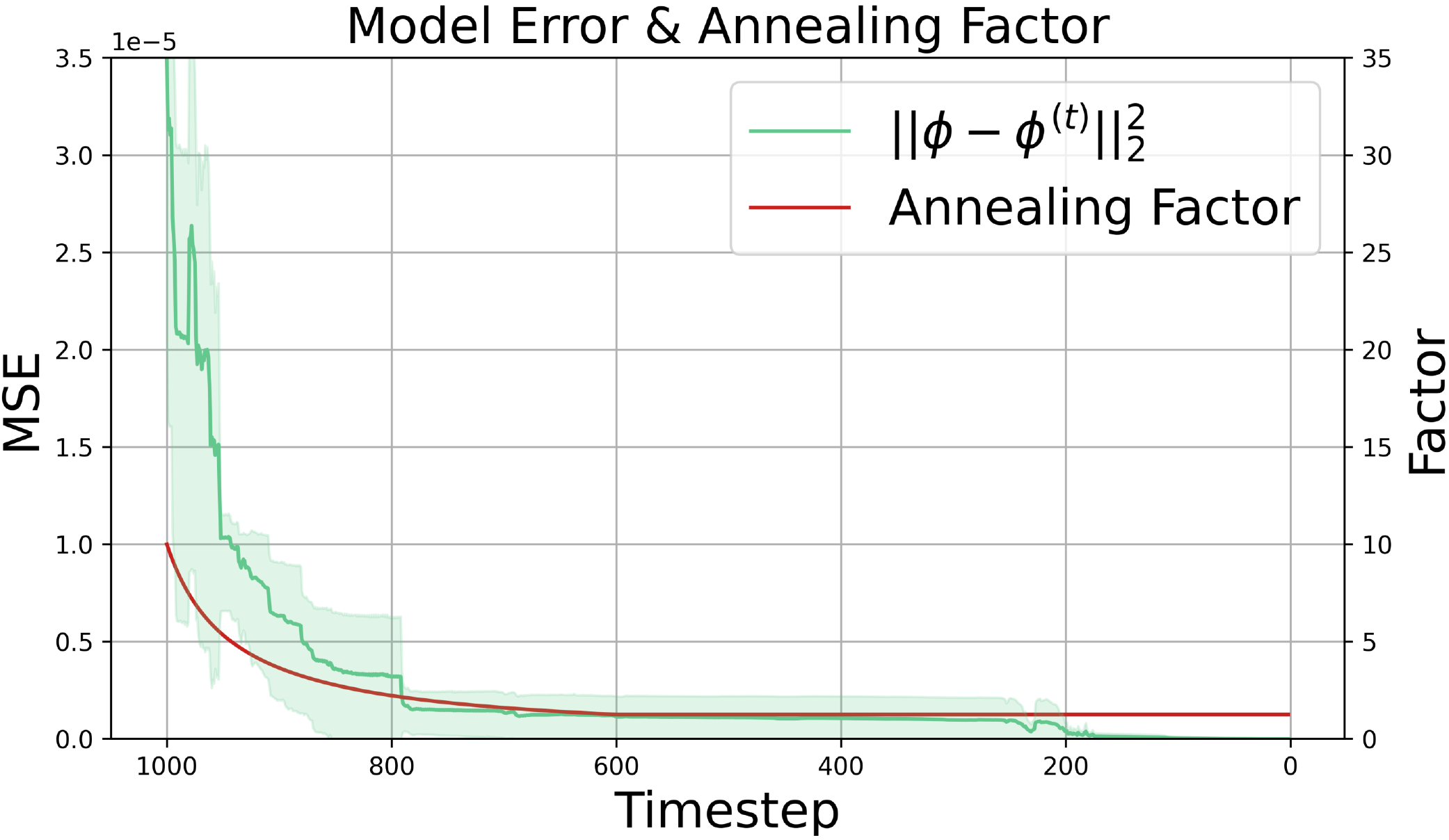}
\end{tabular}
         \end{minipage}
    \begin{minipage}[t]{0.44\textwidth}
    % \vspace*{-0.5cm}
	\centering
	\setlength{\tabcolsep}{1pt}
	\setlength{\fboxrule}{1pt}
	%\vspace*{1.5cm}
	\begin{tabular}{c}
		\begin{tabular}{cccc}
			 \tiny{Observation}  & \tiny{No Annealing} & \tiny{ W. Annealing } &\tiny{GT}
			\\ 
			\multicolumn{1}{c}{
				\begin{overpic}[width=0.24\linewidth]{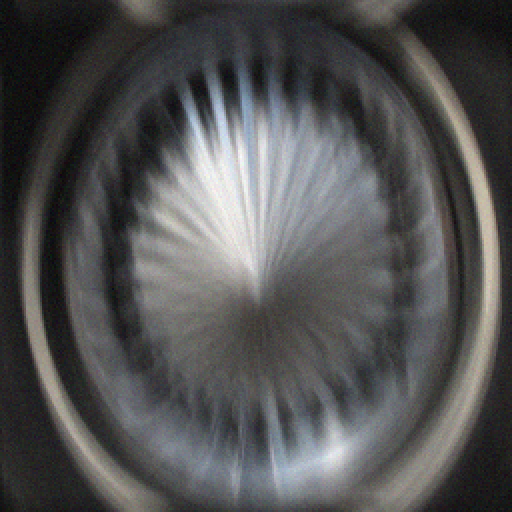}
				\end{overpic}
			}  &
			\multicolumn{1}{c}{
				\begin{overpic}[width=0.24\linewidth]{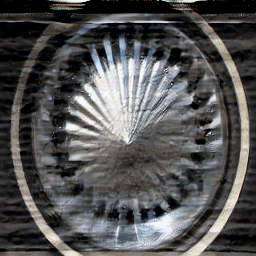}
					\put(65,0){{\includegraphics[width=0.077\linewidth]{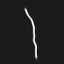}}}
				\end{overpic}
			}  &
			\multicolumn{1}{c}{
				\begin{overpic}[width=0.24\linewidth]{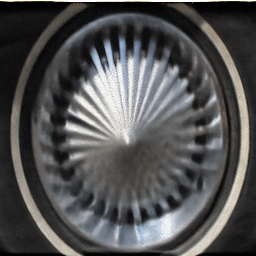}
					\put(65,0){{\includegraphics[width=0.077\linewidth]{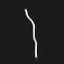}}}
				\end{overpic}
			}  &
			\multicolumn{1}{c}{
				\begin{overpic}[width=0.24\linewidth]{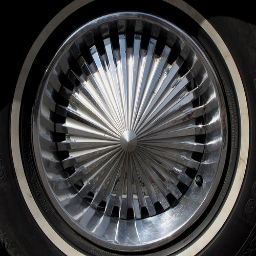}
					\put(65,0){{\includegraphics[width=0.077\linewidth]{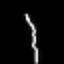}}}
				\end{overpic}
			}  \\
			\multicolumn{1}{c}{
				\begin{overpic}[width=0.24\linewidth]{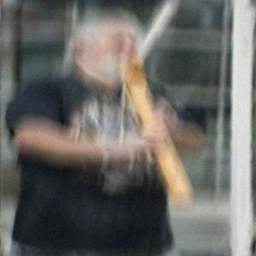}
				\end{overpic}
			}  &
			\multicolumn{1}{c}{
				\begin{overpic}[width=0.24\linewidth]{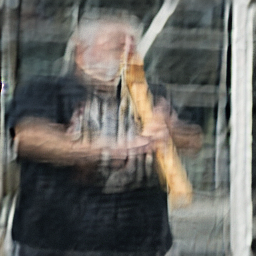}
					\put(65,0){{\includegraphics[width=0.077\linewidth]{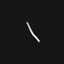}}}
				\end{overpic}
			}  &
			\multicolumn{1}{c}{
				\begin{overpic}[width=0.24\linewidth]{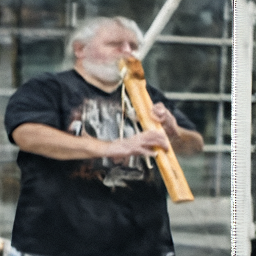}
					\put(65,0){{\includegraphics[width=0.077\linewidth]{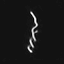}}}
				\end{overpic}
			}  &
			\multicolumn{1}{c}{
				\begin{overpic}[width=0.24\linewidth]{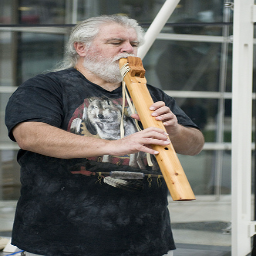}
					\put(65,0){{\includegraphics[width=0.077\linewidth]{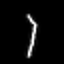}}}
				\end{overpic}
			}  
		\end{tabular}
	\end{tabular}
    \end{minipage}
    \vspace*{-0.1cm}
		\caption{\textbf{Effectiveness of our annealed consistency technique.} Left: blur kernel accuracy curve (green) on 10 examples (std are represented by shadow). It indicates that the kernel is very wrong at the beginning but becomes meaningful when $t<600$, which corresponds to the annealing factor curve (red). Right: we further show that simply applying LDM priors in blind inversion produces images with severe artifacts due to the fragile latent space, while the annealing technique stabilizes the optimization and generates much better results. }
	\label{fig:exp-aba-annealing}
	  \vspace*{-1.0cm}
\end{figure*}

\vspace*{-0.1cm}
\paragraph{Results.} \ 
We provide novel view synthesis results in Fig~\ref{fig:exp-3D}. %, where we show the synthesized novel view images.
Our model is built on top of Zero123~\cite{liu2023zero} priors. The performance of Zero123 has demonstrated expressive in synthesizing high-quality novel-view images, but it sometimes fails in creating view-consistent results across different views, as shown in the first row of Fig.~\ref{fig:exp-3D} .
% The common issues like texture degradation and geometric misalignment.
A major reason is that it synthesizes new views from a single input only and cannot capture the whole information of the 3D object, resulting in hallucinated synthesis.
Instead, our method could easily crack this nut by aggregating the information of all input views and embedding them together through hard data consistency~\cite{song2023solving}. We show we acquire much more consistent novel view images from only two input views.
Moreover, our method doesn't need to know the poses of images as they can be jointly estimated, which supports the more challenging pose-free, sparse view 3D reconstruction. We compare our method with the current state-of-the-art pose-free 3D reconstruction baseline, LEAP~\cite{jiang2022LEAP}. As shown in the second row of Fig.~\ref{fig:exp-3D}, while LEAP fails to generate photo-realistic new views, our method could leverage the powerful Zero123 prior to overcome texture degradation and geometric misalignment, maintaining fine details like the geometry and the texture of the jacket. We provide more results and analysis in Appendix~\ref{subsec:3Drecon},~\ref{subsec:moreinputviews},~\ref{app:re}.
% Besides, our method successfully estimates the target pose compared to input view 2 from a zero initialization.
% This can be shown by the similar poses of the novel view with results from Zero123 since it only provides the target pose compared to input view 1.

\subsection{Ablation Studies}
% We perform extensive ablation studies to verify the proposed techniques: 
%1) Annealed Data Consistency and 
% 2) Skip-gradient in Figure~\ref{fig:exp-aba-annealing} and Figure~\ref{fig:exp-aba-skip}. \wz{name be consistent}

% \paragraph{Stable diffusion v.s. LDM-VQ-4.}

\paragraph{Annealed Consistency.}
A major problem when using LDMs instead of pixel-based DMs is the vulnerable latent space optimization. % which easily brings strong image artifacts~\cite{chung2023prompt}.
In the context of blind inverse problems, the inaccurate forward operators at the beginning could make the problem even worse, where the optimal solutions significantly deviate from the true value and contain strong image artifacts, as demonstrated in Fig.~\ref{fig:exp-aba-annealing} right. 
To ensure stable optimization, we should set our empirical annealing coefficients($\zeta_t$ anneals linearly from 10 at $t=1000$ to 1 at $t=600$ and then holds) based on the forward modeling errors, as shown in Fig.~\ref{fig:exp-aba-annealing}.
This technique show stabilizes the optimization process and produces a more accurate estimation(Fig.~\ref{fig:exp-aba-annealing} right). We provide more annealing analysis in Appendix~\ref{app:anneal}.

\begin{comment}
    To reduce such artifacts, we propose to scale up the guidance of the data consistency term in Eq.~\ref{equ: scale-up}.
Specifically, We compare four annealed optimization schedules: 
$\zeta$ anneals linearly 
1) from 10 at $t=1000$ to 1.25 at $t=800$,
2) from 10 at $t=1000$ to 1.25 at $t=200$,
3) from 10 at $t=1000$ to 1 at $t=600$,
4) from 10 at $t=1000$ to 1.25 at $t=600$, the experimental results are in Figure~\ref{fig:exp-aba-annealing}(c-f), respectively. The noise scheduler is the vanilla DDPM scheduler with a total of 1000 timesteps. Empirical results verify the effectiveness of our annealed technique.
\end{comment}

\paragraph{Skip Gradient for Acceleration.}
We also investigate the influence of the skip-gradient technique, where we compute the full EM step every $\skipNum$ steps while in the middle steps we only run the latent space diffusion. 
% given the skip-gradient ending iteration $\skipEnd$ and the skipped number $\skipNum$, before $\skipEnd$'s iteration, we compute the full EM step every $\skipNum$ time while the rest we only run the latent space diffusion, resulting in the total accelerated steps $M$ as $(T-\skipEnd)(1-1/\skipNum)$, \ie, $\skipEnd=1000$ and $\skipNum=1$ means no acceleration.  
We validate 4 different groups of hyperparameters and compare their running times and imaging quality on 2D blind deblurring task (Table~\ref{tab:skipgrad}).
We find the running time for LatentDEM linearly decreases as the number of accelerated steps $M$ increases. In the accelerated steps, the encoder-decoder inference and gradients are ignored, therefore largely reducing the total optimization time. 
Moreover, though we have skipped a lot of computation burden, due to the fact that the gradient descent tends to follow the previous optimization trajectory, it still results in meaningful convergence. %the optimization direction, we didn;t observe severe signals that the performce still keeps. 
In an extreme setting (case f) where we skip 900 gradients, our method still outperforms baselines, as well as achieving the fastest optimization speed.

 \begin{table}
	\centering
 \caption{{\textbf{Effectiveness of our skip gradient technique.} We evaluate different skip gradient schemes on the blind deblurring task. Compared to the setting without skipping, skipping half steps or even more steps performs similarly better than baselines while reducing the running time.}}
	\resizebox{.9\textwidth}{!}{
		\begin{tabular}{ccccc}
			\toprule
			
		&	\multirow{2}[1]{*}{\textbf{Method}} & Running & Image & Kernel \\
		&	&   Time & PNSR & MSE \\
			\midrule
		a &	FastEM & 1min30sec  & 17.46 & 0.048\\
		b &	BlindDPS & 1min34sec  & 22.58 & 0.035 \\
		c & 	No skipping. ($\skipEnd=1000,\skipNum=1, M=0$) & 6min33sec & 23.45 & 0.011\\ 
		d &	Skip-grad. ($\skipEnd=500,\skipNum=8, M=437$) & 4min17sec & 23.44 & 0.011\\
		e &	Skip-grad. ($\skipEnd=0,\skipNum=8, M=875$) & 2min54sec & 23.00 & 0.010\\
	f &		Skip-grad. ($\skipEnd=0,\skipNum=16, M=937$) & 1min20sec & 22.56 & 0.010\\
			\bottomrule
	\end{tabular}}
	
		\label{tab:skipgrad}
	\vspace*{-0.1cm}
\end{table}

\section{Conclusion}
In this work, we proposed LatentDEM, a novel method that incorporates powerful latent diffusion priors for solving blind inverse problems.  
Our method jointly estimates underlying signals and forward operator parameters inside an EM framework with two deliberately designed optimization techniques, which showcases more accurate and efficient 2D blind debluring performance than prior arts, as well as demonstrates new capabilities in 3D sparse-view reconstruction.
%
% {\name} is a promising technique that opens a new door to a wide range of inverse problems, where the future directions could be its adaptation to more challenging medical and astronomy imaging tasks. 
%
Its limitation includes that in 3D tasks it still relies on LDMs fine-tuned with multi-view images.  It is interesting to think about how to combine {LatentDEM} with SDS loss to directly run 3D inference from purely 2D diffusion models.

%%%%%%%%%%%%%%%%%%%%%%%%%%%%%%%%%%%%%%%%%%%%%%%%%%%%%%%%%%%%

\bibliographystyle{plain} %ieeetr为一种IEEE期刊标准的参考引用格式，其余还有多种选择，可根据自己所投期刊进行修改

% \bibliographystyle{plain} %ieeetr为一种IEEE期刊标准的参考引用格式，其余还有多种选择，可根据自己所投期刊进行修改
% \bibliography{ref}%%reference就是你所命名的bib文件的文件名字
% \medskip

%%%%%%%%%%%%%%%%%%%%%%%%%%%%%%%%%%%%%%%%%%%%%%%%%%%%%%%%%%%%

\newpage
\appendix

We provide an appendix to describe the details of our derivations and algorithms, as well as show more results.  We first provide a theoretical explanation for the annealing consistency technique in Sec.~\ref{app:anneal}. We then provide implementation details of the 2D blind deblurring in Sec.~\ref{app:2d} and the 3D pose-free, sparse-view reconstruction in Sec.~\ref{app:3d}.  Lastly, we report more results in Sec.~\ref{app:re}.

\section{Theoretical Explanations for Annealing Consistency}
\label{app:anneal}
We consider the blind inversion as a time-dependent modeling process and the annealing consistency technique helps address time-dependent modeling errors. Specifically, we express the image formation model  as:

\begin{equation}
	\observation = \forwardFunc_{\forwardParam^{(t)}}(\hat{\signal}_0) + w_t + \boldsymbol{n}, \quad w_t \sim \mathcal{N}(\mathbf{0}, \nu_t^2\mathbf{I}), \quad \boldsymbol{n} \sim \mathcal{N}(\mathbf{0}, \sigma^2\mathbf{I}), 
\end{equation}

where $\forwardParam^{(t)}$ represents the estimated forward model parameters at step $t$,  $\hat{\signal}_0=\decodeNote^*(\mathbb{E}(\latent_0|\latent_t))$,
$w_t$ is the time-dependent modeling noise which is assumed to follow a Gaussian distribution with a time-dependent standard deviation of $\nu_t$, and $\boldsymbol{n}$ is the observation noise with a constant standard deviation of $\sigma$. 
Therefore, following ~\cite{chung2022diffusion} we derive below  data likelihood term to account for both modeling errors and observation noise during the diffusion posterior sampling: 

% $n$ is the constant observation noise while $w_t$ is time-dependent modeling error and we assume it is also a Gaussian distribution. 

% Since $\forwardFunc_\forwardParam$ is randomly initialized and inaccurate initially, the exact image formation model can be expressed as:

% Here, 
\begin{equation}
\scoreFunc{\latent_\noisetime}{\observation | \latent_\noisetime} \approx - \frac{1}{2(\nu_t^2 + \sigma^2)} \nabla_{\signal_{t}} \left\| \observation - \mathcal{A}_\phi \left( \pretraiedDecodeFunc{\mathbb{E}[{\latent}_{0} | \latent_{t}]}\right)\right\|_{2}^{2},
\label{equ:modelerror}
\end{equation}
where $\nu_t^2$ should gradually decrease from a large value to zero as the estimated model parameters converge to the ground truth.  
This is consistent with the proposed technique, where in Eq.~\ref{equ: scale-up}  $\zeta_t$ linearly anneals from a large number to a constant. 
As a result, annealing consistency aligns better with the blind inversion problem and brings more stable optimization, as well as superior performance.

\section{Implementation Details of 2D Blind Deblurring}
\label{app:2d}
\subsection{E-step}
\paragraph{Diffusion Posterior Sampling with Gluing Regularization.}
In the 2D blind deblurring task, the E-step performs latent diffusion posterior sampling (DPS) to reconstruct the underlying image, assuming a known blur kernel.
% In the blind 2d deblurring task, we first assume the knowledge of the right blur kernel in the E-step. Our goal is to recover the underlying image with the known blur kernel and the latent diffusion model. 
The basic latent DPS takes the form as follows:
\begin{equation}
\begin{split}
\scoreFunc{\latent_\noisetime}{\latent_\noisetime | \observation}
    &\approx \pretrainSdModel{(\latent_\noisetime, \noisetime)} + \nabla_{\boldsymbol{z}_{t}} p\left(\boldsymbol{y} | \mathcal{D}^*\left(\mathbb{E}\left[\boldsymbol{z}_{0} | \boldsymbol{z}_{t}\right]\right)\right),\\
    &= \pretrainSdModel{(\latent_\noisetime, \noisetime)} - \frac{1}{2\zeta_t \sigma^2} \nabla_{\latent_{t}}\left\|\observation-\mathcal{A}_\phi\left( \pretraiedDecodeFunc{\mathbb{E}[{\latent}_{0}| \latent_{t}]}\right)\right\|_{2}^{2} ,
     % \boldsymbol{z}_{t-1} \leftarrow \boldsymbol{z}_{t-1} - \frac{1}{2\zeta_t\sigma^2} \nabla_{\boldsymbol{z}_t} \lVert \boldsymbol{y} - \forwardFunc_{\phi^{(t-1)}}(\mathcal{D}^*(\mathbb{E}\left[\boldsymbol{z}_{0} | \boldsymbol{z}_{t}\right]) \rVert^2.
\end{split}
     \label{equ:pixel DPS}
\end{equation}
which simply transform the equation from pixel-based DPS~\cite{chung2022diffusion} to the latent space. However, this basic form always produces severe artifacts or results in reconstructions inconsistent with the measurements~\cite{song2023solving}. 
A fundamental reason is that the decoder is an one-to-many mapping, where numerous latent codes $ \boldsymbol{z}_0$ that represent underlying images can match the measurements. 
Computing the gradient of the density specified by Eq.~\ref{equ:pixel DPS} could potentially drive $\boldsymbol{z}_t$ towards multiple different directions. 
To address this, we introduce an additional constraint called ``gluing''~\cite{rout2024solving} to properly guide the optimization in the latent space:
\begin{equation}
\begin{aligned}
\nabla_{\boldsymbol{z}_{t}} \log p\left(\boldsymbol{y} | \boldsymbol{z}_{t}\right) & =\underbrace{\nabla_{\boldsymbol{z}_{t}} p\left(\boldsymbol{y} | \mathcal{D}^*\left(\mathbb{E}\left[\boldsymbol{z}_{0} | \boldsymbol{z}_{t}\right]\right)\right)}_{\text {DPS vanilla extension }} \\
& +\gamma \underbrace{\nabla_{\boldsymbol{z}_{t}}|| \mathbb{E}\left[\boldsymbol{z}_{0} | \boldsymbol{z}_{t}\right]-\mathcal{E}^*\left(\mathcal{A}_{\phi^{(t-1)}}^{T} \boldsymbol{y} +\left(\boldsymbol{I}-\mathcal{A}_{\phi^{(t-1)}}^{T} \mathcal{A}_{\phi^{(t-1)}}\right) \mathcal{D}^*\left(\mathbb{E}\left[\boldsymbol{z}_{0} | \boldsymbol{z}_{t}\right]\right)\right) \|^{2}_2}_{\text {``gluing'' regularization}},
\end{aligned}
\end{equation}
where $\gamma$ is a tunable hyperparameter. 
The gluing objective~\cite{rout2024solving} is critical for LatentDEM as it constrains the latent code update following each M-step, ensuring that the denoising update, data fidelity update, and the gluing update point to the same optima. {Note that gluing is also involved in the skip-gradient technique, \ie, we will also ignore it during the skipped steps.}  
% \wz{I add one sentence, Check.}

\subsection{M-step}
The M-step solves the MAP estimation of the blur kernel using the posterior samples $\hat{\signal}_0$ from the E-step. This process is expressed by:
% \begin{equation}
% \forwardFunc_\phi^{(t-1)} = \arg \max_\forwardFunc\left[ Q(\forwardFunc, \forwardFunc_\phi^{(t)}) + \log(p(\forwardFunc)) \right]
% \label{equ:base Mstep}
% \end{equation}
% under the 2D motion deblur setting Eq.~\ref{equ: inverse} , the equation can be changed into:
\vspace*{-0.1cm}
\begin{equation}
    \forwardParam^*  = \arg \min_{\phi} \mathbb{E}_{\hat{\signal}_0}\left[\frac{1}{2\sigma^2}|| \observation-\forwardFunc_\forwardParam(\hat{\signal}_0) ||^2_2  + \mathcal{R}(\phi)\right], 
\label{equ:advance Mstep}
\end{equation}
% \begin{equation}
% \forwardFunc_{\phi^{(t-1)}} = \arg \min_\forwardFunc \left[ \frac{1}{2n\sigma^2}  \|y - \forwardFunc(\mathcal{D}^*(\hat{\boldsymbol{z}}_0)\|_2^2 + \lambda \Phi(\forwardFunc) \right]
% \label{equ:advance Mstep}
% \end{equation}
where $\sigma^2$ denotes the noise level of the measurements and $\mathcal{R}$ is the regularizer.
Common choices of the regulation term can be $l_2$ or $l_1$ regularizations on top of
the physical constraints on the blur kernel (non-negative values that add up to one). 
Despite being quite efficient when the blurry image does not have noise, they generally fail to provide high-quality results when the noise level increases~\cite{laroche2024fast}.
Therefore, we decide to leverage a Plug-and-Play (PnP) denoiser, $\mathbf{D}_{\sigma_d}$, as the regularizer. 
We find that training the denoiser on a dataset of blur kernels with various noise levels ($\sigma_d$) can lead to efficient and robust kernel estimation.
Now with this PnP denoiser as the regularizer, we can solve Eq.~\ref{equ:advance Mstep} with the Half-Quadratic Splitting (HQS) optimization scheme:

\begin{equation}
\mathbf{Z}_{i+1} = \arg \min_\mathbf{Z} \left[ \frac{1}{2\sigma^2} \|\forwardFunc_\mathbf{Z}(\hat{\signal}_0) - \observation\|_2^2 + \frac{\delta}{2} \|\mathbf{Z} - \phi_i\|_2^2 \right],
\label{equ:fourier}
\end{equation}

\begin{equation}
\phi_{i+1} = \mathbf{D}_{\sqrt{\lambda/\delta}}(\mathbf{Z}_{i+1})=\arg \min_\phi \left[\lambda \mathcal{R}(\phi) + \frac{\delta}{2} \|\phi - \mathbf{Z}_{i+1}\|_2^2 \right],
\label{equ:pnp process}
\end{equation}
where $\mathbf{Z}$ is a intermediate variable, $\mathbf{D}_{\sigma_d}$ is a PnP neural denoiser~\cite{laroche2024fast, zhang2017beyond}, $\sigma$ defines the noise level of measurements, and $\lambda$, $\beta$ are tunable hyperparameters~\cite{zhang2021plug}. The subscripts $_i$ and $_{i+1}$ index iterations of Eq.~\ref{equ:fourier} and Eq.~\ref{equ:pnp process} in one M-step.
For the deconvolution problem, Eq~\ref{equ:fourier} can easily be solved in the Fourier domain and Eq~\ref{equ:pnp process} corresponds to the regularization step. It corresponds to the MAP estimator of a Gaussian denoising problem on the variable
$\mathbf{Z}_{i+1}$. 
The main idea behind the PnP regularization is to replace this regularization step with a pre-trained denoiser. 
This substitution can be done becaue of the close relationship between the MAP and the MMSE estimator of a Gaussian denoising problem. 
In the end, the M-step can be expressed by Eq~\ref{eq:hqs2}.

\paragraph{Plug-and-Play Denoiser.}
We train a Plug-and-Play (PnP) denoiser to serve as the kernel regularizer in the M-step. 
For the architecture of the denoiser, we use a simple DnCNN~\cite{zhang2017beyond} with 5 blocks and 32 channels. 
In addition to the noisy kernel, we also take the noise level map as an extra channel and feed it to the network to control the denoising intensity. 
The settings are similar to one of the baseline methods, FastEM~\cite{laroche2024fast}. 
In the data preparation process, we generate 60k motion deblur kernels with random intensity and add random Gaussian noise to them. 
The noisy level map is a 2D matrix filled with the variance and is concatenated to the kernel as an additional channel as input to the network.
We train the network for 5,000 epochs by denoising the corrupted kernel and %,for the close relationship between the PnP regularization and a pretrained Gaussian denoise~\cite{laroche2024fast}. For the loss function, we 
use the MSE loss. All the training is performed on a NVIDIA A100 which lasts for seven hours. We also try different network architectures like FFDNet but find the DnCNN is sufficient to tackle our task and it's very easy to train.
% 
% During testing, the motion kernel is synthesized with fixed intensity.
% add a citation 
% 怎么产生kernel，怎么加噪音，多少个kernel，什么loss，训练多少。
% 我们还尝试了xxx网络结构，但是发现对结果的影响不大。

\paragraph{Hyperparameters.}

For the motion deblur task, we leverage the codebase of PSLD~\cite{rout2024solving}, which is based on Stable Diffusion-V1.5.
Besides the hyperparameters of the annealing and skip-gradient technique, we find it critical to choose the proper parameters for the gluing and M-step. 
Improper parameters result in strong artifacts.
In our experiments, we find the default hyperparameters in ~\cite{rout2024solving} won't work, potentially due to the more fragile latent space. The hyperparameters in our M-step are $\lambda = 1$ and $\delta = 5e6$, and We iterate Eq.~\ref{eq:hqs2} 20 times to balance solution convergence and computational efficiency.
% We define the hyperparameter, where $\lambda = 1$ and $\beta = 5e6$. 
% For each M step, we usually iterate  Eq.~\ref{eq:hqs2} several times till the kernel convergences and we empirically find that 20 iterations always be a good choice to balance the convergence and the iteration times.

% latent-dps，咱们怎么选的weighting，有什么心得
% 参考latent-dps和quqing的paper怎么说的

\section{Implementation Details of Pose-free Sparse-view 3D Reconstruction}
\label{app:3d}
% method部分详细些的推导
% 尽可能凑到10页，可以多放些实验
\subsection{Basics}
\paragraph{Problem Formulation.}
The pose-free, sparse-view 3D reconstruction problem aims to reconstruct a 3D object from multiple unposed images. This task can be formulated as a blind inversion problem with the following forward model:
\begin{equation}
\observation = \forwardFunc_\forwardParam \left( \signal \right), \observation = \{\observation_1, \cdots, \observation_n\}, \quad \forwardParam = \{\forwardParam_1, \cdots, \forwardParam_n\},
\end{equation}
where $\signal$ represents the underlying 3D object, $\observation$ is the set of observations containing images from multiple views, and $\phi$ denotes the camera parameters corresponding to different views. The task requires us to jointly solve for both the 3D model, $\signal$, and the image poses, $\phi$, to reconstruct a view-consistent 3D object. 3D models can be represented in various forms, including meshes, point clouds, and other formats. In our paper, we implicitly represent the 3D model as a collection of random view observations of the 3D object. These views can then be converted into 3D geometry using volumetric reconstruction methods such as Neural Radiance Fields (NeRF) or 3D Gaussian Splatting (3DGS).

% Sparse View unposed Image 3d Reconstruction is the task in which you input multiple images like 2 ~ 3 images without knowing their relative position to reconstruct a 3d object, the main obstacles are 1. You have to estimate the relative position between the input images to effectively use the additive information to reconstruct a 3d object compared with the 3d reconstruction task with only one image.2.you have to reconstruct a 3d object that is consistent with multiple input images. So this task can be easily turned into a blind inverse problems framework:
% \begin{equation}
% 	\observation = \forwardFunc_\forwardParam \left( \signal_1,\signal_2,... \right) 
% 	\label{equ: 3d inverse}
% \end{equation}
% the $\signal_1,\signal_2$ can represents the input images and the $\forwardParam$ can represent their relative location , and $\observation$ is the output 3d object which can satisfy all these constaints.
\paragraph{Camera Model Representation.}
\begin{figure*}
\centering
\includegraphics[width=5cm]{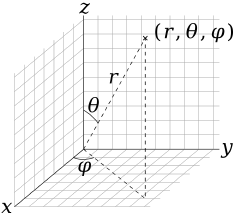}
\centering
\caption{\textbf{Spherical coordinate system~\cite{website:coord}.}}
\label{fig:exp-aba-cam}
\end{figure*}
We employ a spherical coordinate system to represent camera poses and their relative transformations. As shown in Fig.~\ref{fig:exp-aba-cam}, we place the origin of the coordinate system at the center of the object. In this system, $\theta$, $\phi$, and $r$ represent the polar angle, azimuth angle, and radius (distance from the center to the camera position), respectively.
The relative camera pose between two views is derived by directly subtracting their respective camera poses. For instance, if two images have camera poses $(\theta_1, \phi_1, r_1)$ and $(\theta_2, \phi_2, r_2)$, their relative pose is calculated as $(\theta_2 - \theta_1, \phi_2 - \phi_1, r_2 - r_1)$.

% To generate a novel-view image, it is necessary to specify the relative camera transformation from the input image to the new image. For instance, when the camera poses of the input and novel-view images are denoted as $(\theta_1, \phi_1, r_1)$ and $(\theta_2, \phi_2, r_2)$, the relative camera transformation is $(\theta_2 - \theta_1, \phi_2 - \phi_1, r_2 - r_1)$. 

% To produce a new image with the camera location represented by $(\theta_2, \phi_2, r_2)$, we can express the relative camera transformation as $(\theta_2 - \theta_1, \phi_2 - \phi_1, r_2 - r_1)$ and apply Zero123 model to generate the corresponding new view.

% \paragraph{Pose-free Sparse-view 3D Reconstruction Task.}
% Zero123 only takes single image as input.  Though it is simple, as shown in Fig.~\ref{fig:exp-3D} in the main paper, it is challenging to generate new views that are consistent with the input views. 
% %
% In contrast, we consider the pose-free, sparse-view 3D reconstruction which we reformulate it as a blind inversion problem since we do not require any pose information.  Specifically, given the first, input view,  we require the user to specify a target view. % with relative camera transformation between the first input view and the target view. 
% The relative camera transformations between the rest unposed images and the target view can be estimated using our algorithm.

\paragraph{Zero123.}
Zero123 is a conditional latent diffusion model, $\scoreFunc{\latent_\noisetime}{\latent_\noisetime | \observation, \phi}$, fine-tuned from Stable Diffusion for novel view synthesis. It generates a novel-view image at a target viewpoint, $\signal_0 = \mathcal{D}(\latent_0)$, given an input image, $\observation$, and the relative camera transformation, $\phi$, where $\mathcal{D}$ maps the latent code to pixel space. This model enables 3D object generation from a single image: by applying various camera transformations, Zero123 can synthesize multiple novel views, which can then be used to reconstruct a 3D model. As a result, Zero123 defines a powerful 3D diffusion prior.
\begin{wrapfigure}{r}{0.3\textwidth} % {r} 表示图片在文本右侧，{0.5\textwidth} 表示图片宽度为半栏  
  \centering  
  \includegraphics[width=0.3\textwidth]{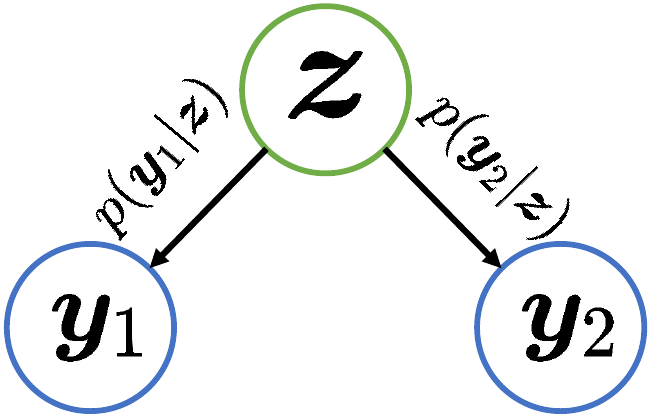} % 图片文件名为your-image.jpg，并设置图片宽度为0.45\textwidth  
  \caption{\textbf{Graphic model of the E-step.}}  
  \label{fig:your-label}  
  \vspace{-0.5in}
\end{wrapfigure}
% todo: y|z

% zero123~\cite{liu2023zero1to3}is a specially trained latent diffusion model fine-tuned on stable diffusion to generate new view images based on 1 input image and the relative position of the new viewpoint. and specifically, zero123's workflow is mainly a conditional diffusion model,the reverse diffusion process also starts with completely chaos guassians and with embedding condition of the input image and relative new position , it can automatically generate the new  image of given new view point.our program is based on the zero123 and with designed EM algorithm, we don't have to perform any advance training of the latent diffusion model to settle the sparse-view pose-free 3d reconstruction problems.
% \subsection{overall algorithm}
% the overall algorithm of 2 input view's 3d reconstruction is shown in algorithm 2.The main idea of this algorithm is we perform the zero123 generation process for each input image, and using E-step to generate reasonable results and M-step to estimate the unknown pose information.And because the starting point of the unknown pose is completely random, we choose to empirically separate the whole generation process into 2 stage, the first stage is to optimize the pose in order to get a nearly right pose to perform the E-step(which can be interpret a stage whose learning rate of E-step is 0);the second stage perform the completely EM optimization framework.
% \begin{figure}[tbp]
%     \centering
    % \begin{minipage}[t]{1.0\textwidth}
        % \vspace{0pt} % Resets any previous vertical space
        \begin{algorithm}[tbp]
            % \scriptsize % Reduce font size to footnotesize
            \begin{algorithmic}
                \caption{LatentDEM for Pose-free Sparse-view 3D Reconstruction}
                \label{algo:2}
                \REQUIRE $T, \boldsymbol{y}_1,\boldsymbol{y}_2,\phi_1,\phi_2^{(T)},  \{\beta_i\}_{i=1}^T, \{\tilde{\sigma}_i\}_{i=1}^T,\{{\gamma}_i\}_{i=0}^{T-1},\delta, \lambda,  \mathcal{E}^*, \mathcal{D}^*, \boldsymbol{s}^*_\theta$
                \STATE $\boldsymbol{z}_T(\boldsymbol{y}_1,\phi_1) \sim \mathcal{N}(\boldsymbol{0}, \boldsymbol{I})$
                \STATE $\boldsymbol{z}_T(\boldsymbol{y}_2,\phi_2^{(T)}) \sim \mathcal{N}(\boldsymbol{0}, \boldsymbol{I})$
                \FOR{$t = T$ \textbf{to} $0$}
                \STATE $\boldsymbol{s}_1 \leftarrow \boldsymbol{s}^*_\theta(\boldsymbol{z}_t(\boldsymbol{y}_1,\phi_1), t,\boldsymbol{y}_1,\phi_1)$
                \STATE $\boldsymbol{s}_2 \leftarrow \boldsymbol{s}^*_\theta(\boldsymbol{z}_t(\boldsymbol{y}_2,\phi_2^{(t)}), t,\boldsymbol{y}_2,\phi_2^{(t)})$
                \STATE $\hat{\boldsymbol{z}}_0(\boldsymbol{y}_1,\phi_1) \leftarrow \frac{1}{\sqrt{\bar{\alpha}_i}} (\boldsymbol{z}_t(\boldsymbol{y}_1,\phi_1) + (1 - \bar{\alpha}_t)\boldsymbol{s}_1)$
                \STATE $\hat{\boldsymbol{z}}_0(\boldsymbol{y}_2,\phi_2^{(t)}) \leftarrow \frac{1}{\sqrt{\bar{\alpha}_i}} (\boldsymbol{z}_t(\boldsymbol{y}_2,\phi_2^{(t)}) + (1 - \bar{\alpha}_t)\boldsymbol{s}_2)$
                \STATE $\boldsymbol{\epsilon} \sim \mathcal{N}(\boldsymbol{0}, \boldsymbol{I})$
                \STATE $\boldsymbol{z}_{t-1}(\boldsymbol{y}_1,\phi_1) \leftarrow \frac{\sqrt{\alpha_t}(1-\bar{\alpha}_{t-1})}{1-\bar{\alpha}_t} \boldsymbol{z}_t(\boldsymbol{y}_1,\phi_1) + \frac{\sqrt{\bar{\alpha}_{t-1}}\beta_t}{1-\bar{\alpha}_t} \hat{\boldsymbol{z}}_0(\boldsymbol{y}_1,\phi_1) + \tilde{\sigma}_t \boldsymbol{\epsilon}$
                \STATE $\boldsymbol{z}_{t-1}(\boldsymbol{y}_2,\phi_2^{(t-1)}) \leftarrow \frac{\sqrt{\alpha_t}(1-\bar{\alpha}_{t-1})}{1-\bar{\alpha}_t} \boldsymbol{z}_t(\boldsymbol{y}_2,\phi_2^{(t)}) + \frac{\sqrt{\bar{\alpha}_{t-1}}\beta_t}{1-\bar{\alpha}_t} \hat{\boldsymbol{z}}_0(\boldsymbol{y}_2,\phi_2^{(t)}) + \tilde{\sigma}_t \boldsymbol{\epsilon}$
                
                \STATE $\boldsymbol{z}_{t-1}(\boldsymbol{y}_1,\phi_1,\boldsymbol{y}_2,\phi_2^{(t-1)})= \mathrm{E\mbox{-}step}(\boldsymbol{z}_{t-1}(\boldsymbol{y}_1,\phi_1) ,\boldsymbol{z}_{t-1}(\boldsymbol{y}_2,\phi_2^{(t-1)}) ,\gamma_{t-1})$

                \STATE $\phi_2^{(t-1)} = \mathrm{M\mbox{-}step}(\boldsymbol{z}_{t-1}(\boldsymbol{y}_1,\phi_1) ,\boldsymbol{z}_{t-1}(\boldsymbol{y}_2,\phi_2^{(t)}), \boldsymbol{z}_{t-1}(\boldsymbol{y}_1,\phi_1,\boldsymbol{y}_2,\phi_2^{(t-1)}),\delta, \lambda)$ 
               \STATE $\boldsymbol{z}_{t-1}(\boldsymbol{y}_1,\phi_1) \leftarrow \boldsymbol{z}_{t-1}(\boldsymbol{y}_1,\phi_1,\boldsymbol{y}_2,\phi_2^{(t-1)}) $
              
                %\STATE $\boldsymbol{z}_{i-1} \leftarrow \boldsymbol{z}''_{i-1} - \gamma_i \nabla_{\boldsymbol{z}_i} \lVert \tilde{\boldsymbol{z}}_0 - \mathcal{E}^*(\forwardFunc_{i-1}^T \forwardFunc_{i-1} \tilde{\boldsymbol{x}}^*_0 + (\boldsymbol{I} - \forwardFunc_{i-1}^T \forwardFunc_{i-1})\mathcal{D}^*(\tilde{\boldsymbol{z}}_0)) \rVert^2$
                \ENDFOR
                \STATE \textbf{return} $\mathcal{D}^*(\hat{\boldsymbol{z}}_0(\boldsymbol{y}_1,\phi_1) ), \phi_2^{(0)}$
            \end{algorithmic}
        \end{algorithm}
        \vspace*{-0.3cm} % Adds vertical space after the algorithm
        
\subsection{E-Step} \label{subsec:3D_Estep}
In the E-step, we perform a view-consistent diffusion process to generate target novel views from multiple input images, where we assume the camera pose of each image is known.

% todo: graphic model
\paragraph{Derivation of View-consistent Diffusion Model from Zero123.}
The view-consistent generation is governed by a latent diffusion process conditioned on multiple images, defined as $\scoreFunc{\latent_\noisetime}{\latent_\noisetime | \observation_1, \phi_1, \observation_2, \phi_2}$. In the graphical model of the 3D reconstruction problem, the input images $\observation_1$ and $\observation_2$ represent different views of the same object. 
These views should be independent of each other given the geometry of the 3D object, which is described by the latent code $\latent_\noisetime$. Consequently, the view-consistent diffusion can be derived from Zero123 as follows:
\begin{equation}
\begin{split}
    p(\latent_{\noisetime-1} | \latent_{\noisetime}, \observation_1, \phi_1, \observation_2, \phi_2) &\propto p(\observation_1, \observation_2 | \latent_{\noisetime-1}, \latent_{\noisetime}, \phi_1, \phi_2)\\
    &= p(\observation_1| \latent_{\noisetime-1}, \latent_{\noisetime}, \phi_1) p(\observation_2| \latent_{\noisetime-1}, \latent_{\noisetime}, \phi_2)
    \\
    &\propto p(\latent_{\noisetime-1}|\latent_{\noisetime}, \observation_1, \phi_1) p(\latent_{\noisetime-1}| \latent_{\noisetime}, \observation_2, \phi_2),
\end{split}
\label{eq:view-consistent}
\end{equation}
where the conditional diffusions from single images, $p(\latent_{\noisetime-1}| \latent_{\noisetime}, \observation_1, \phi_1)$ and $p(\latent_{\noisetime-1}|\latent_{\noisetime}, \observation_2, \phi_2)$, are defined by Zero123, and they both follow Gaussian distributions according to the Langevin dynamics defined by the reverse-time SDE (Eq.~\ref{equ:diffusion}),
% $p_t(\latent_\noisetime| \observation_1, \phi_1)=\mathcal{N}(\sqrt{\bar{\alpha}_t}\hat{z_{0}}(\observation_1, \phi_1),(1-\bar{\alpha}_t)\textit{I})$ and $p_t(\latent_\noisetime| \observation_2, \phi_2)=\mathcal{N}(\sqrt{\bar{\alpha}_t}\hat{z_{0}}(\observation_2, \phi_2),\sigma_t^{2})$ 

\begin{equation}
\begin{split}
    p_t(\latent_{\noisetime-1}| \latent_{\noisetime}, \observation_1, \phi_1)=\mathcal{N}\left(\frac{1}{\sqrt{1-\beta_t}}\left[\latent_{\noisetime}+\beta_t \scoreFunc{\latent_\noisetime}{\latent_\noisetime | \observation_1, \phi_1}\right], \beta_t I \right)\\
    p_t(\latent_{\noisetime-1}| \latent_{\noisetime}, \observation_2, \phi_2)=\mathcal{N}\left(\frac{1}{\sqrt{1-\beta_t}}\left[\latent_{\noisetime}+\beta_t \scoreFunc{\latent_\noisetime}{\latent_\noisetime | \observation_2, \phi_2}\right], \beta_t I \right)
\end{split}
\end{equation}
In our pose-free 3D reconstruction task, we account for potential inaccuracies in $\phi_2$ during early diffusion stages, so the diffusion process is modified as:
\begin{equation}
    p_t(\latent_{\noisetime-1}| \latent_{\noisetime}, \observation_2, \phi_2)=\mathcal{N}\left(\frac{1}{\sqrt{1-\beta_t}}\left[\latent_{\noisetime}+\beta_t \scoreFunc{\latent_\noisetime}{\latent_\noisetime | \observation_2, \phi_2}\right], (\beta_t+\nu_t^2) I \right),
\end{equation}
where $\nu_t$ represents the standard deviation of time-dependent model errors. As a result, $p(\latent_{\noisetime-1} | \latent_{\noisetime}, \observation_1, \phi_1, \observation_2, \phi_2)$ becomes
\begin{equation}
    \mathcal{N}\left( \frac{1}{\sqrt{1-\beta_t}}\left[\latent_{\noisetime}+\beta_t \frac{(\beta_t+\nu_t^2)\scoreFunc{\latent_\noisetime}{\latent_\noisetime | \observation_1, \phi_1}+\beta_t\scoreFunc{\latent_\noisetime}{\latent_\noisetime | \observation_2, \phi_2}}{2\beta_t+\nu_t^2}\right], \frac{\beta_t(\beta_t+\nu_t^2)}{2\beta_t+\nu_t^2}\right),
\end{equation}
which defines a view-consistent diffusion model, whose score function is a weighted average of two Zero123 models:
\begin{equation}
\begin{split}
    \scoreFunc{\latent_\noisetime}{\latent_\noisetime | \observation_1, \phi_1, \observation_2, \phi_2} &= (1-\gamma_t)\scoreFunc{\latent_\noisetime}{\latent_\noisetime | \observation_1, \phi_1}+\gamma_t\scoreFunc{\latent_\noisetime}{\latent_\noisetime | \observation_2, \phi_2}, \\
    \gamma_t &= \frac{\beta_t}{2\beta_t+\nu_t^2}.
\end{split}
\label{eq:2viewscore}
\end{equation}
This derivation can be readily extended to scenarios with multiple unposed images.

\paragraph{Annealing of View-consistent Diffusion.}
An annealing strategy is essential for the view-consistent diffusion process in pose-free 3D reconstruction due to initial inaccuracies in estimated camera poses. In a two-image-based 3D reconstruction problem, for instance, model errors are quantified by $\nu_t$ in Eq.~\ref{eq:2viewscore}. This error term starts large and gradually decreases to zero as the camera pose $\phi_2$ becomes increasingly accurate. Consequently, $\gamma_t$ progressively increases from nearly 0 to 0.5. During early stages, the diffusion process primarily relies on a single reference view $\observation_1$, and the intermediate generative images are utilized to calibrate camera poses of other images. As camera poses gain accuracy, the other images exert growing influence on the diffusion process, ultimately achieving view-consistent results.

\subsection{M-step} \label{subsec:3D_Mstep}
The M-step estimates unknown camera poses by aligning unposed images to synthetic and reference views:
\begin{equation}
    \phi_2 = \arg \min_{\phi} \mathbb{E}_{\hat{\signal}_0} \left[\lambda \|\boldsymbol{z}_{t}(\boldsymbol{y}_2, \phi_2) - \boldsymbol{z}_{t}(\hat{\image}_0, \mathbf{0})\|_2^2 + \delta \|\boldsymbol{z}_{t}(\boldsymbol{y}_2, \phi_2) - \boldsymbol{z}_{t}(\boldsymbol{y}_1, \phi_1)\|_2^2\right],
	\label{eq:pose_update}
\end{equation}
where $\hat{\signal}_0$ are the posterior samples from the E-step, and $\lambda$ and $\delta$ balance the calibration loss on the synthetic and reference images, respectively. $\boldsymbol{z}_{t}(\cdot,\cdot)$ is the time-dependent latent variable representing semantic information of the transformed input image. As suggested by Eq.~\ref{equ: pose estimation}, this optimization problem can be efficiently solved using gradient-based optimization. We dynamically adjust the ratio of the two balancing factors, $\lambda/\delta$, throughout the diffusion process. In early stages, we set $\lambda/\delta$ to a small value, primarily relying on the reference image for pose calibration. As the synthetic image becomes more realistic during the diffusion process, we gradually increase $\lambda/\delta$ until it converges to 1, balancing the influence of both synthetic and reference images in the final stages.

% \begin{figure*}[tbp]
% \centering
% \includegraphics[width=8cm]{Styles/figure/blindinverse-appendix-multiviews.pdf}
% \centering
% \caption{\textbf{3D Reconstruction of an Apple Using Different Numbers of Unposed Images.} We evaluate our method's performance with varying numbers of input sparse views. \textbf{Left}: One view (equivalent to Zero123) generates an unrealistic model that is hollow inside. \textbf{Middle}: Two views improve results but still exhibit hallucinations in the 3D geometry of the bitten apple. \textbf{Right}: Three views successfully recover all details.}
% \label{fig:exp-aba-viewconsistent}
% \end{figure*}

\begin{figure*}
	% \vspace*{-0.5cm}
	\centering
	\setlength{\tabcolsep}{1pt}
	\setlength{\fboxrule}{1pt}
	%\vspace*{1.5cm}
	\begin{tabular}{c}
		\begin{tabular}{ccc}
			\multicolumn{1}{c}{
				\begin{overpic}[width=0.25\linewidth]{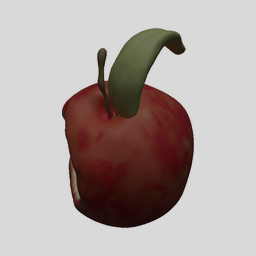}
				\end{overpic}
			}  &
                \multicolumn{1}{c}{
				\begin{overpic}[width=0.25\linewidth]{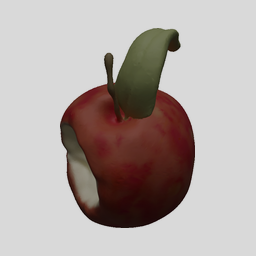}
				\end{overpic}
			}  &
			\multicolumn{1}{c}{
				\begin{overpic}[width=0.25\linewidth]{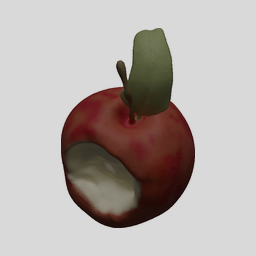}
				\end{overpic}
			}  
                \\
                \multicolumn{1}{c}{Input view 1} &
                \multicolumn{1}{c}{Input view 2} &
                \multicolumn{1}{c}{Input view 3}
			\\[-0.5ex]
			  \multicolumn{1}{c}{
			  	\begin{overpic}[width=0.25\linewidth]{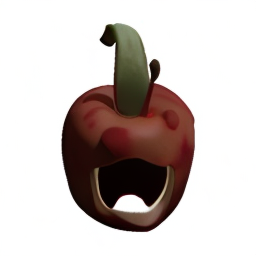}
			  	\end{overpic}
			  }  &
                \multicolumn{1}{c}{
				\begin{overpic}[width=0.25\linewidth]{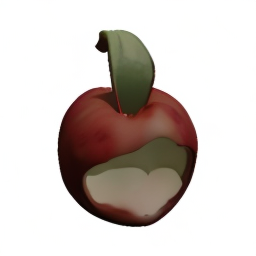}
				\end{overpic}
			}  &
			  \multicolumn{1}{c}{
			  	\begin{overpic}[width=0.25\linewidth]{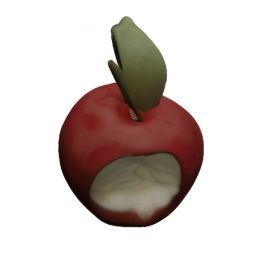}
			  	\end{overpic}
			  }  
                \\
                \multicolumn{1}{c}{Output 1} &
                \multicolumn{1}{c}{Output 2} &
                \multicolumn{1}{c}{Output 3}
		\end{tabular}
	\end{tabular}
	\caption{\textbf{3D reconstruction of an apple using different numbers of unposed images.} We evaluate our method's performance with varying numbers of input sparse views. \textbf{Left}: One view (Output 1 is from Input view 1 which is 1equivalent to Zero123) generates an unrealistic model that is hollow inside. \textbf{Middle}: Output 2 is from Input view 1 \& 2. Two views improve results but still exhibit hallucinations in the 3D geometry of the bitten apple, \eg the apple handle is missing. \textbf{Right}: Output 3 is from all three input views. Three views successfully recover all details.}
\label{fig:exp-aba-viewconsistent}
\end{figure*}

\begin{figure*}
	% \vspace*{-0.5cm}
	\centering
	\setlength{\tabcolsep}{1pt}
	\setlength{\fboxrule}{1pt}
	%\vspace*{1.5cm}
	\begin{tabular}{c}
		\begin{tabular}{cc|ccc|ccc}
			\begin{turn}{90} \!\!\! \!\!\! \!\!\! \!\!\! \!\!\! \!\!\!\small{Shark} \end{turn} & 
                \multicolumn{1}{c|}{
				\begin{overpic}[width=0.135\linewidth]{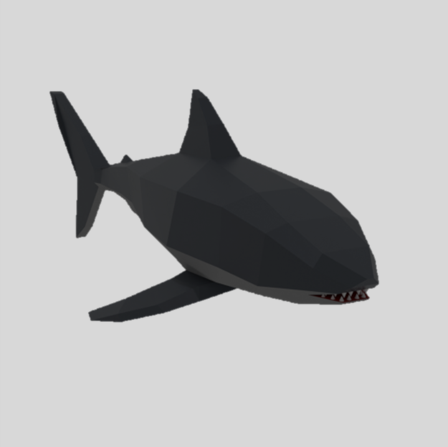}
				\end{overpic}
			}  &
                \multicolumn{1}{c}{
				\begin{overpic}[width=0.135\linewidth]{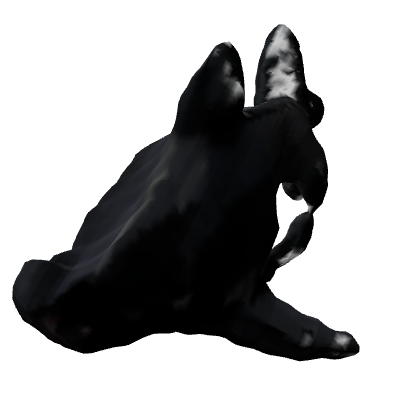}
				\end{overpic}
			}  &
                \multicolumn{1}{c}{
				\begin{overpic}[width=0.135\linewidth]{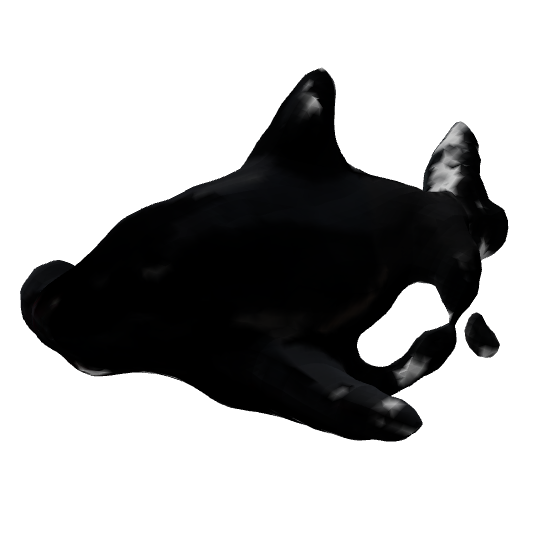}
				\end{overpic}
			}  &
			\multicolumn{1}{c|}{
				\begin{overpic}[width=0.135\linewidth]{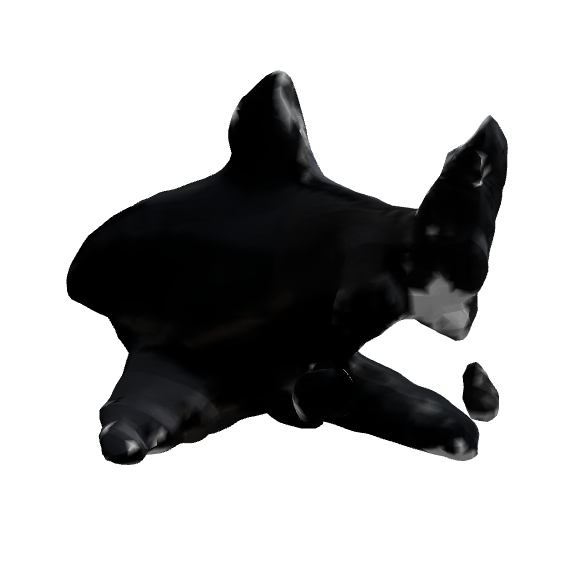}
				\end{overpic}
			}  &
                \multicolumn{1}{c}{
				\begin{overpic}[width=0.135\linewidth]{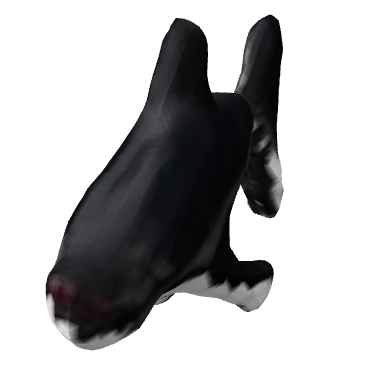}
				\end{overpic}
			}  &
                \multicolumn{1}{c}{
				\begin{overpic}[width=0.135\linewidth]{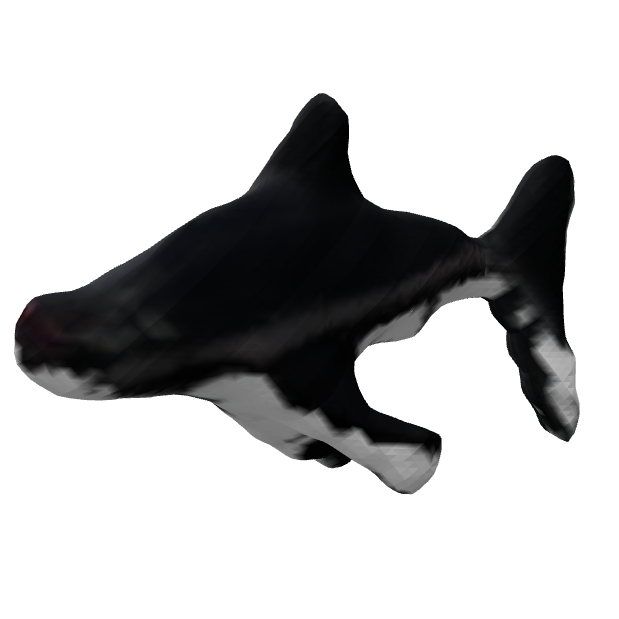}
				\end{overpic}
			}  &
			\multicolumn{1}{c}{
				\begin{overpic}[width=0.135\linewidth]{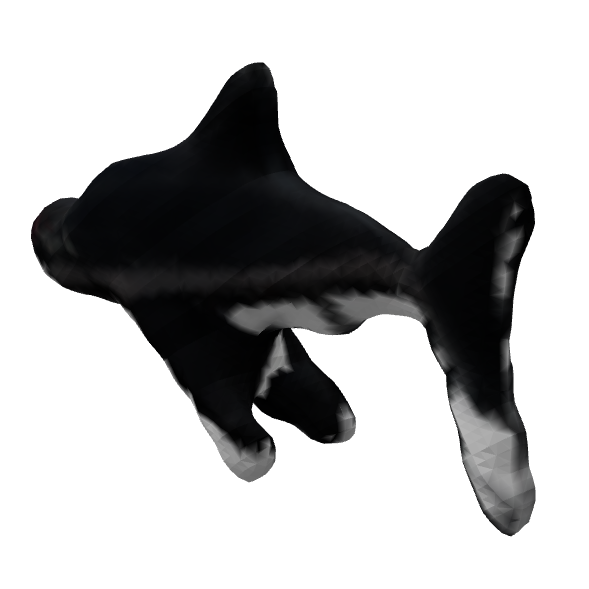}
				\end{overpic}
			}  
                \\[-0.8ex]
                &
			  \multicolumn{1}{c|}{
			  	\begin{overpic}[width=0.135\linewidth]{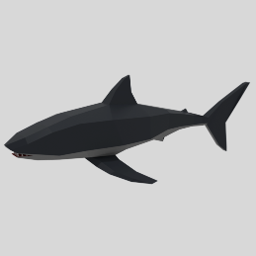}
			  	\end{overpic}
			  }  &
                \multicolumn{1}{c}{
				\begin{overpic}[width=0.135\linewidth]{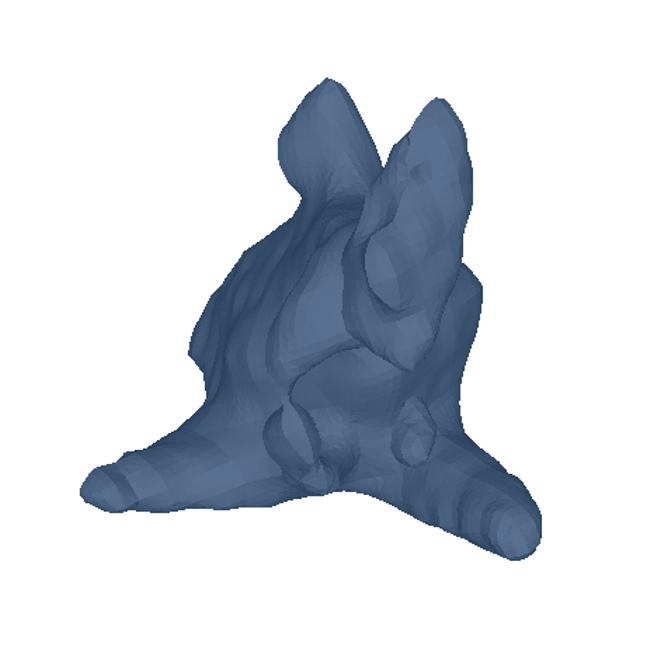}
				\end{overpic}
			}  &
                \multicolumn{1}{c}{
				\begin{overpic}[width=0.135\linewidth]{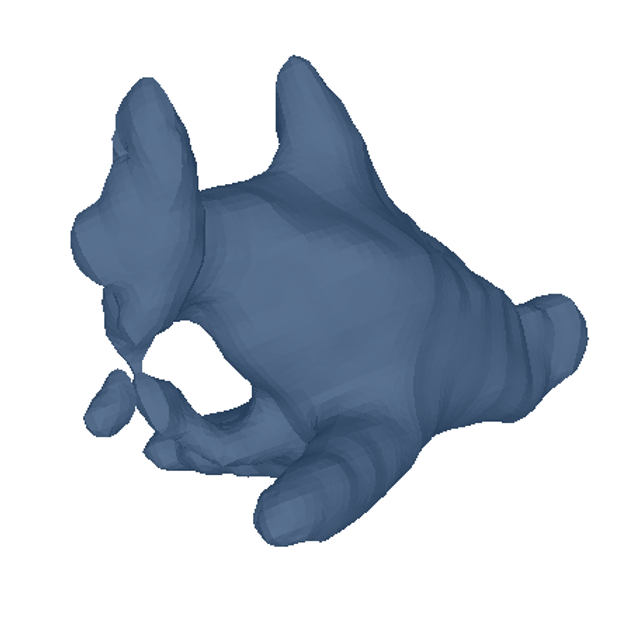}
				\end{overpic}
			}  &
			  \multicolumn{1}{c|}{
			  	\begin{overpic}[width=0.135\linewidth]{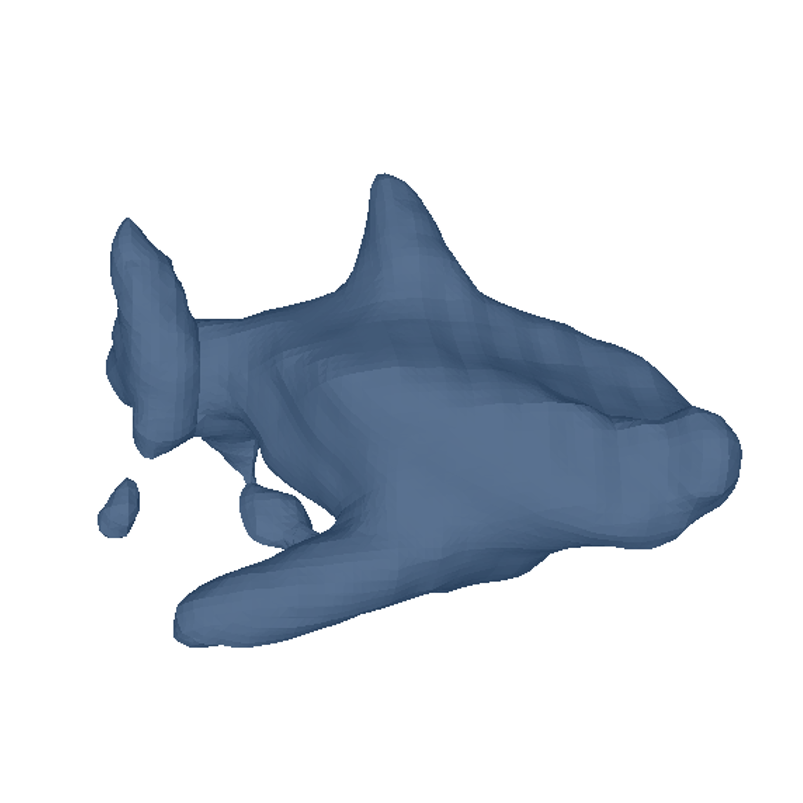}
			  	\end{overpic}
			  }  &
                \multicolumn{1}{c}{
				\begin{overpic}[width=0.135\linewidth]{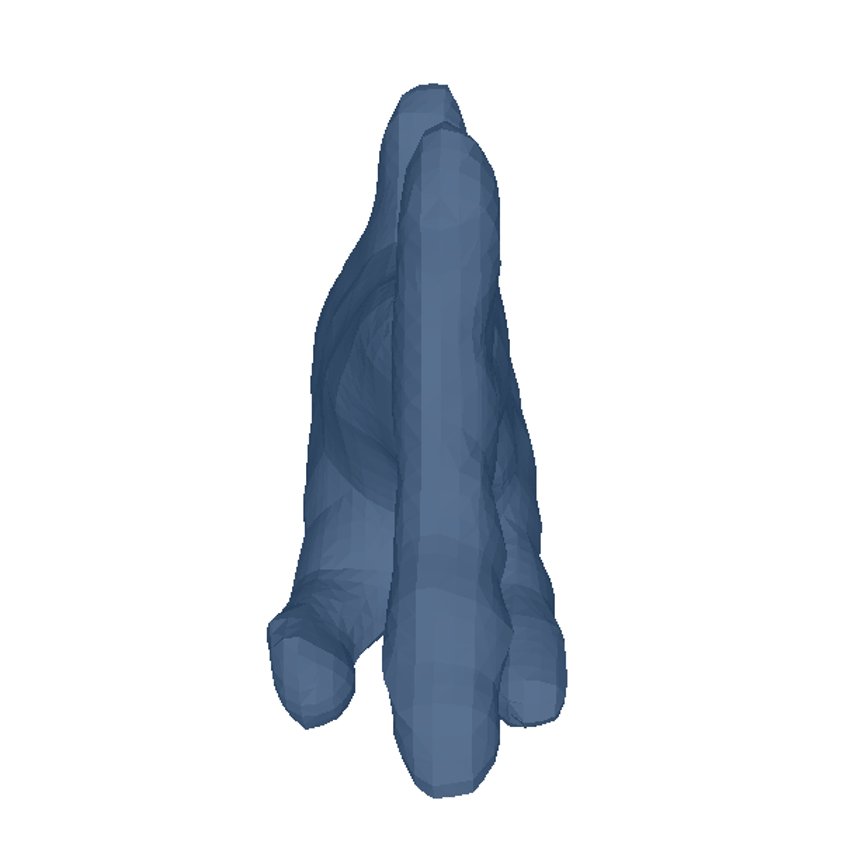}
				\end{overpic}
			}  &
                \multicolumn{1}{c}{
				\begin{overpic}[width=0.135\linewidth]{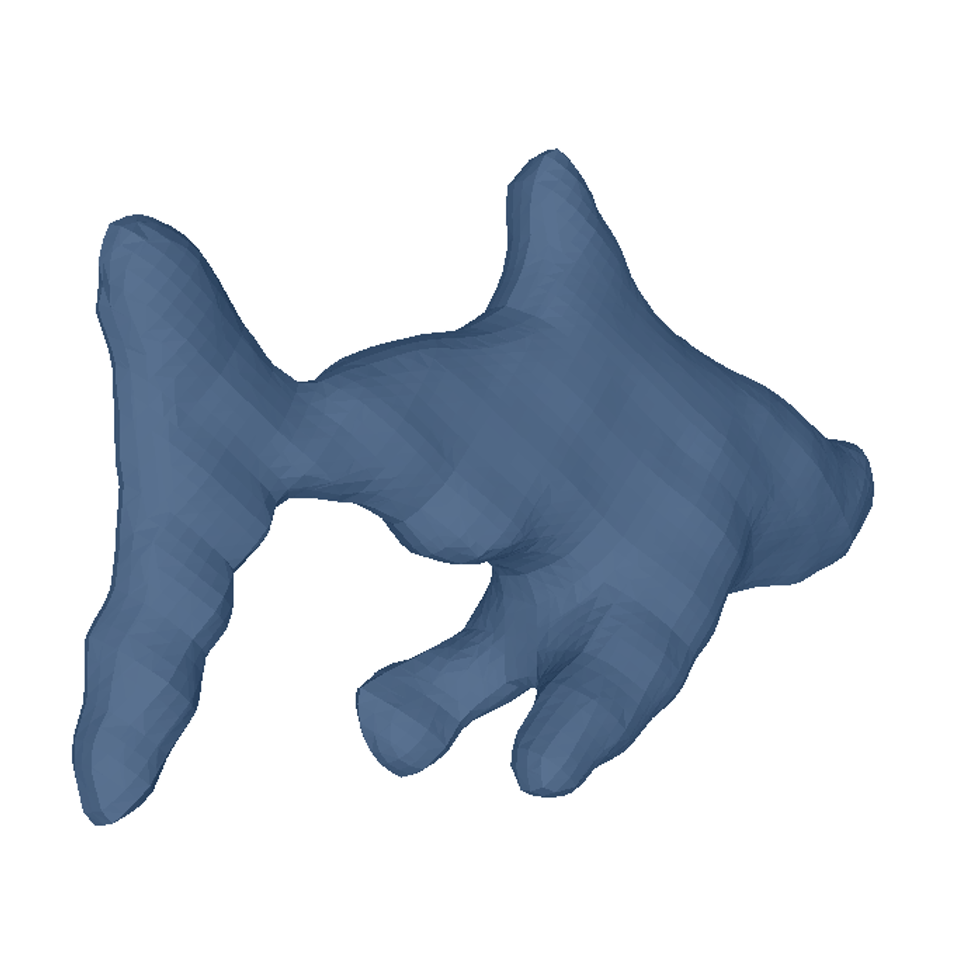}
				\end{overpic}
			}  &
			  \multicolumn{1}{c}{
			  	\begin{overpic}[width=0.135\linewidth]{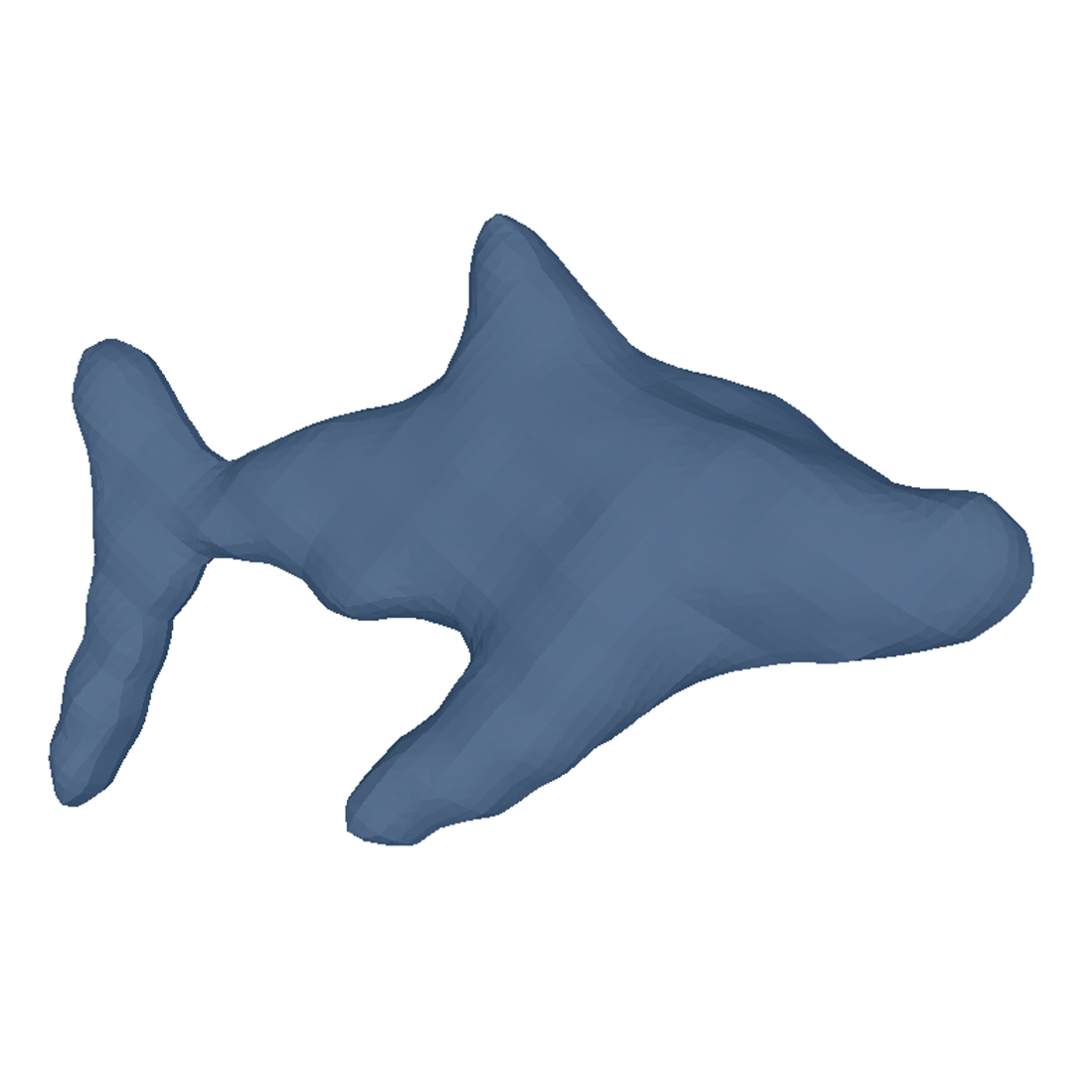}
			  	\end{overpic}
			  }  
                \\
                \begin{turn}{90} \!\!\! \!\!\! \!\!\! \!\!\! \!\!\! \!\!\!\small{Drum} \end{turn} & 
                \multicolumn{1}{c|}{
				\begin{overpic}[width=0.135\linewidth]{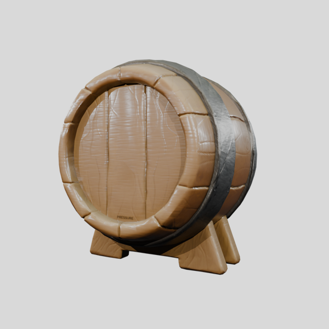}
				\end{overpic}
			}  &
                \multicolumn{1}{c}{
				\begin{overpic}[width=0.135\linewidth]{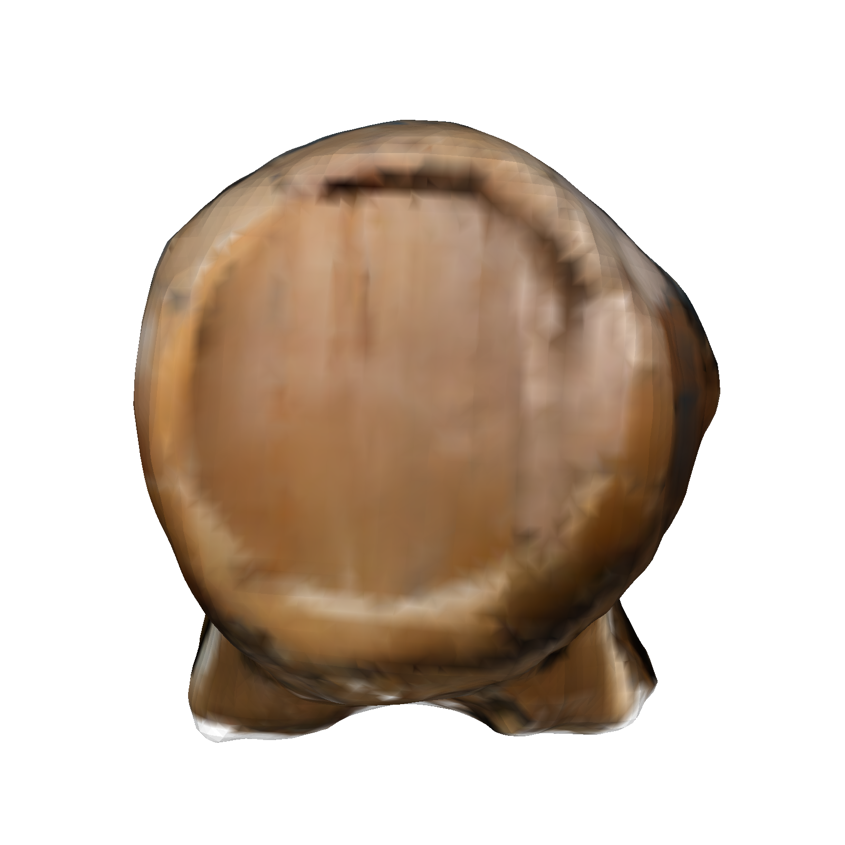}
				\end{overpic}
			}  &
                \multicolumn{1}{c}{
				\begin{overpic}[width=0.135\linewidth]{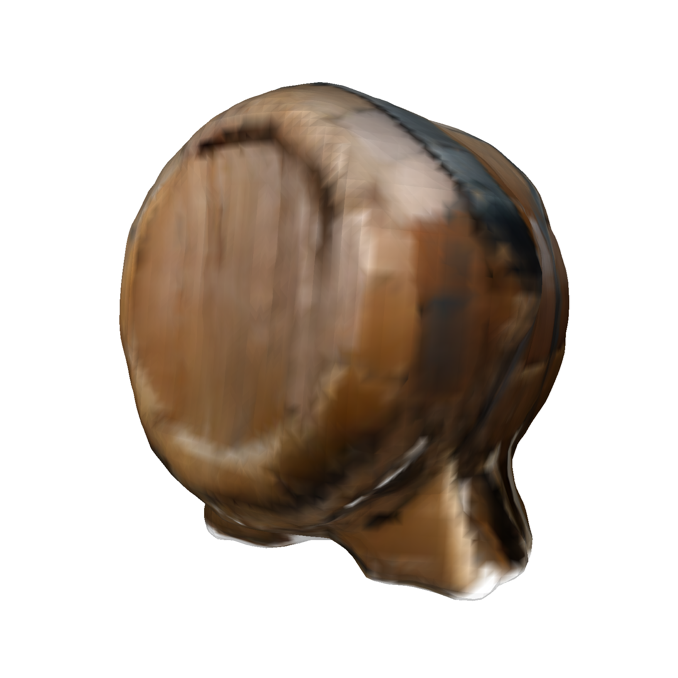}
				\end{overpic}
			}  &
			\multicolumn{1}{c|}{
				\begin{overpic}[width=0.135\linewidth]{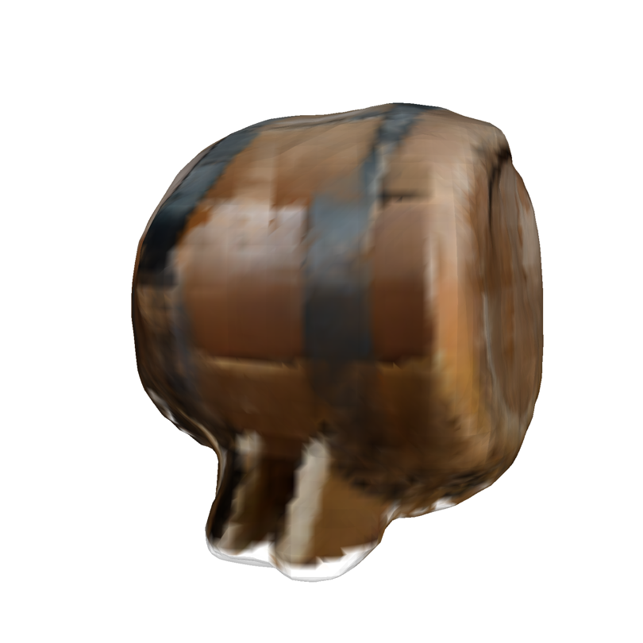}
				\end{overpic}
			}  &
                \multicolumn{1}{c}{
				\begin{overpic}[width=0.135\linewidth]{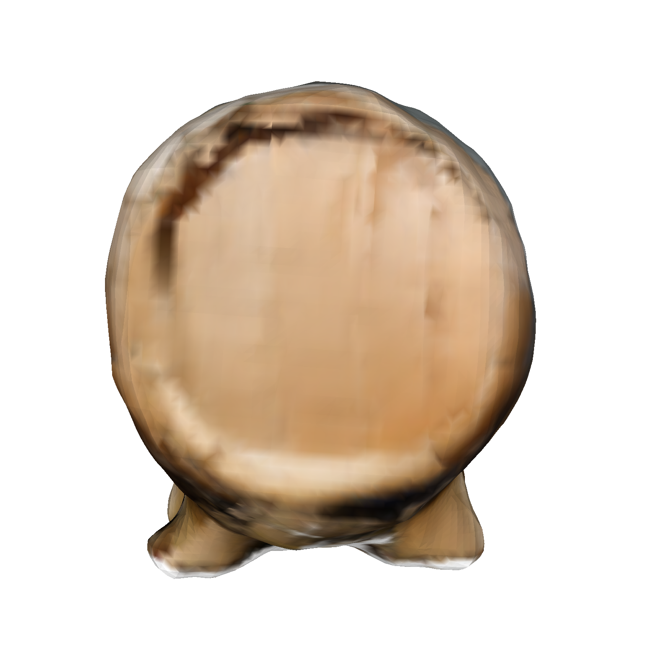}
				\end{overpic}
			}  &
                \multicolumn{1}{c}{
				\begin{overpic}[width=0.135\linewidth]{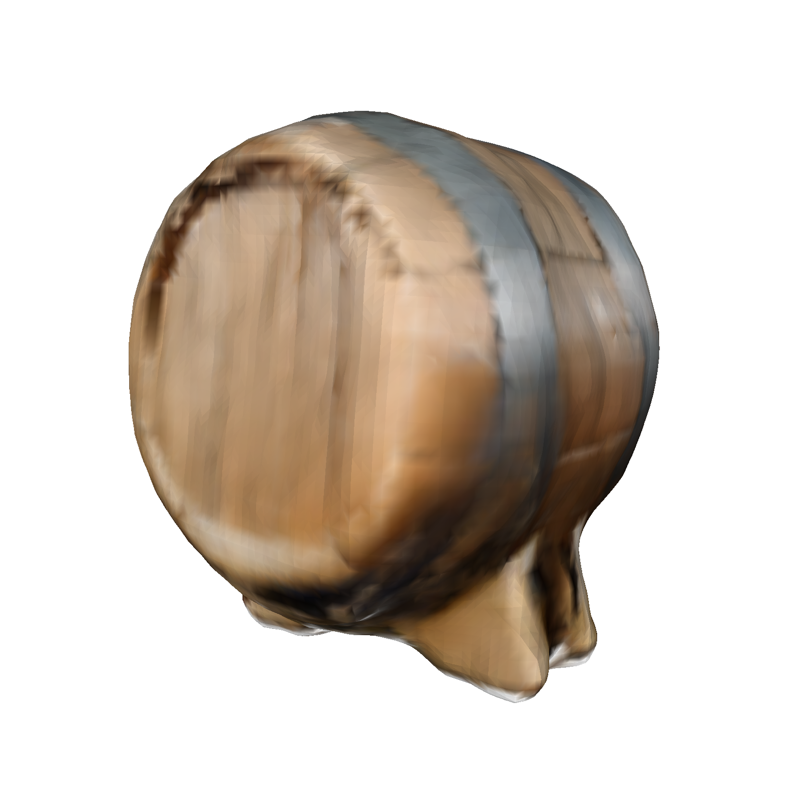}
				\end{overpic}
			}  &
			\multicolumn{1}{c}{
				\begin{overpic}[width=0.135\linewidth]{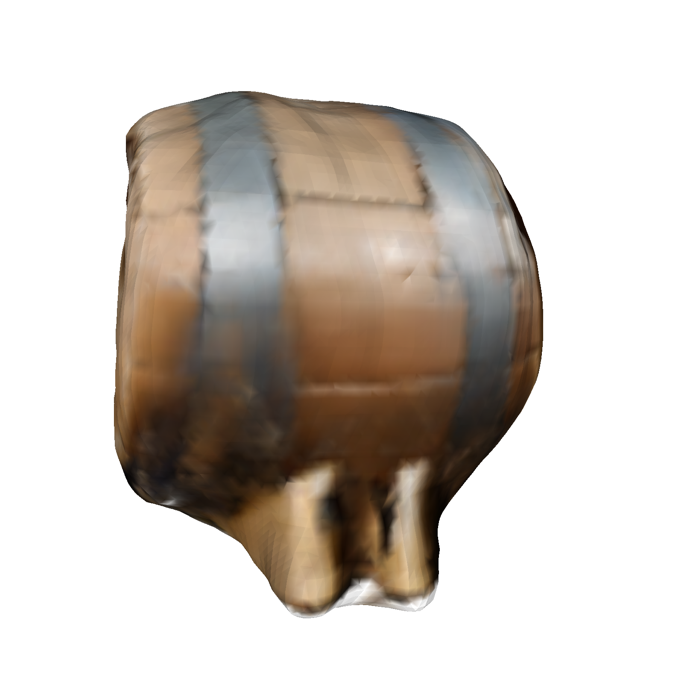}
				\end{overpic}
			}  
                \\[-0.8ex]
                &
			  \multicolumn{1}{c|}{
			  	\begin{overpic}[width=0.135\linewidth]{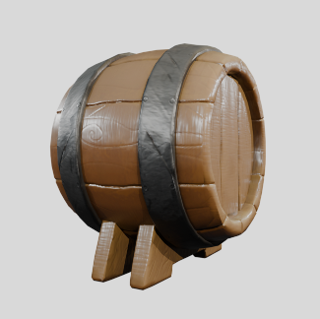}
			  	\end{overpic}
			  }  &
                \multicolumn{1}{c}{
				\begin{overpic}[width=0.135\linewidth]{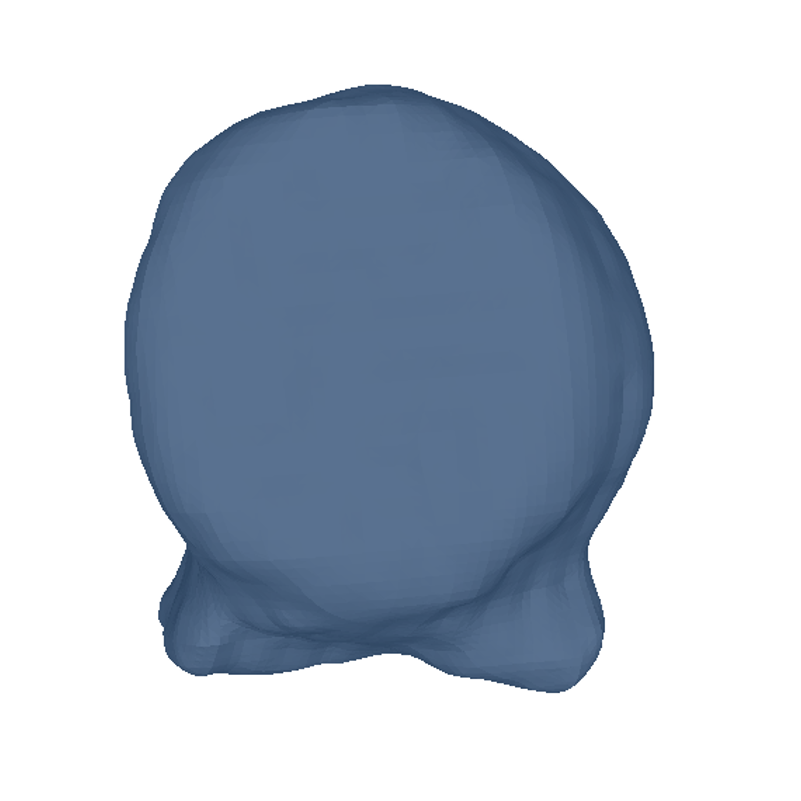}
				\end{overpic}
			}  &
                \multicolumn{1}{c}{
				\begin{overpic}[width=0.135\linewidth]{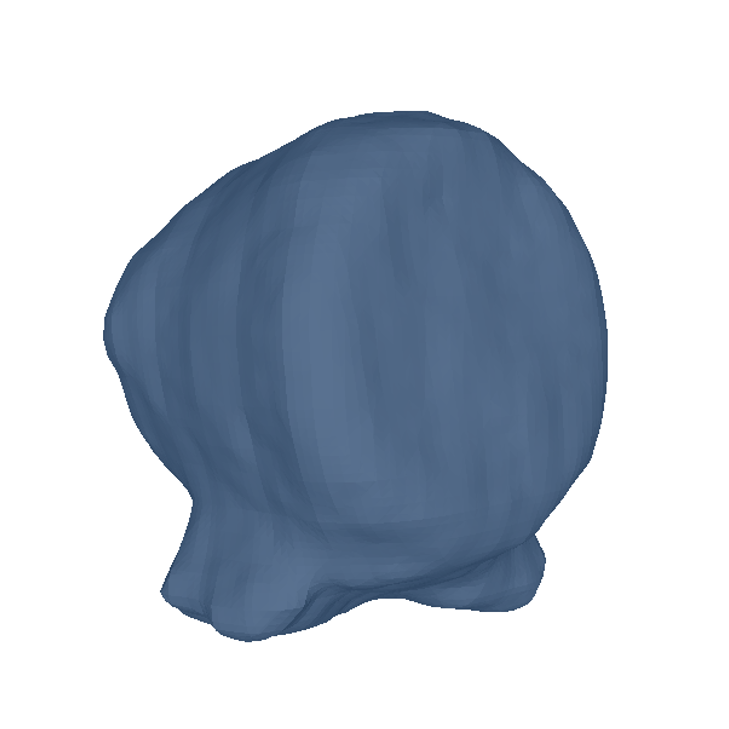}
				\end{overpic}
			}  &
			  \multicolumn{1}{c|}{
			  	\begin{overpic}[width=0.135\linewidth]{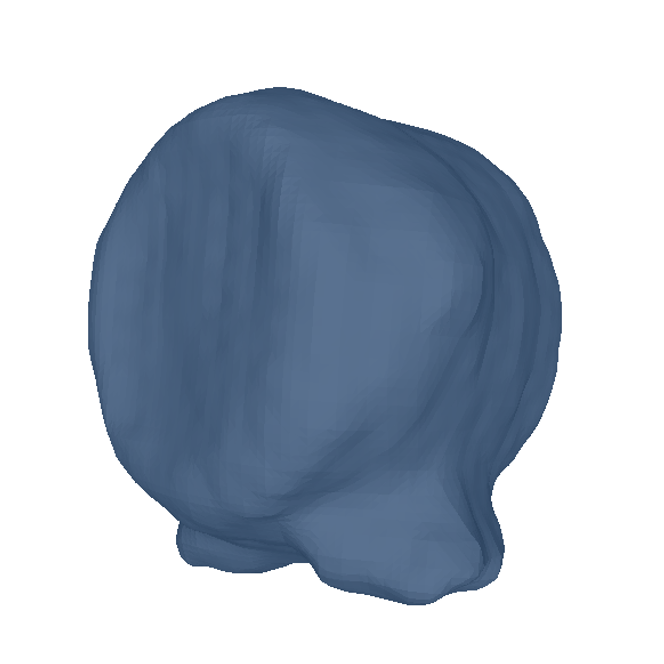}
			  	\end{overpic}
			  }  &
                \multicolumn{1}{c}{
				\begin{overpic}[width=0.135\linewidth]{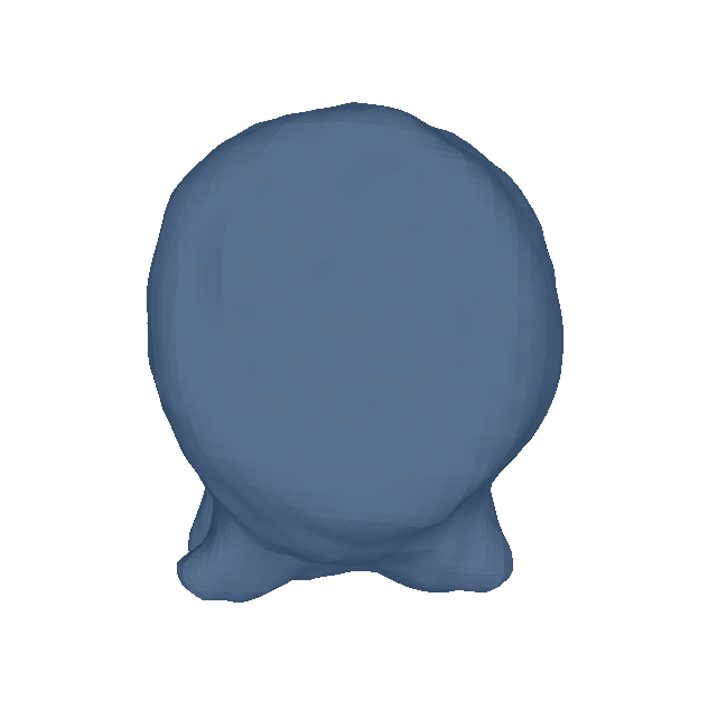}
				\end{overpic}
			}  &
                \multicolumn{1}{c}{
				\begin{overpic}[width=0.135\linewidth]{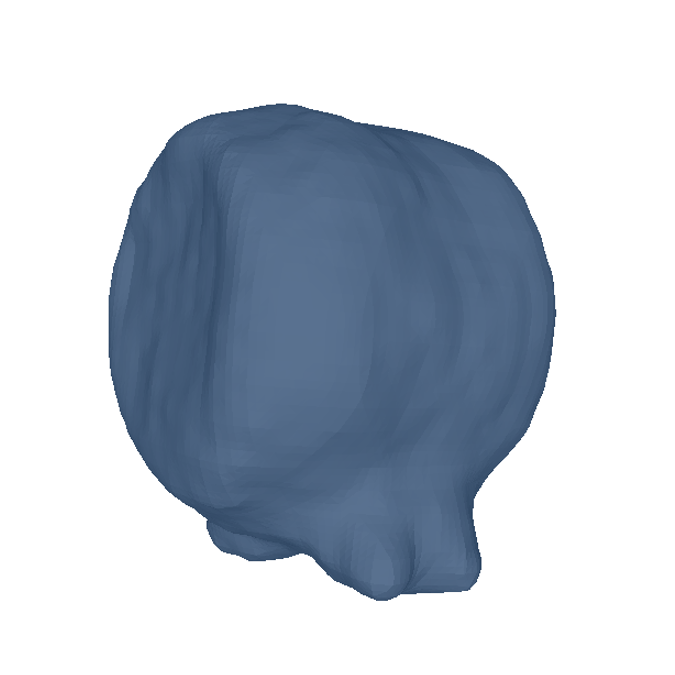}
				\end{overpic}
			}  &
			  \multicolumn{1}{c}{
			  	\begin{overpic}[width=0.135\linewidth]{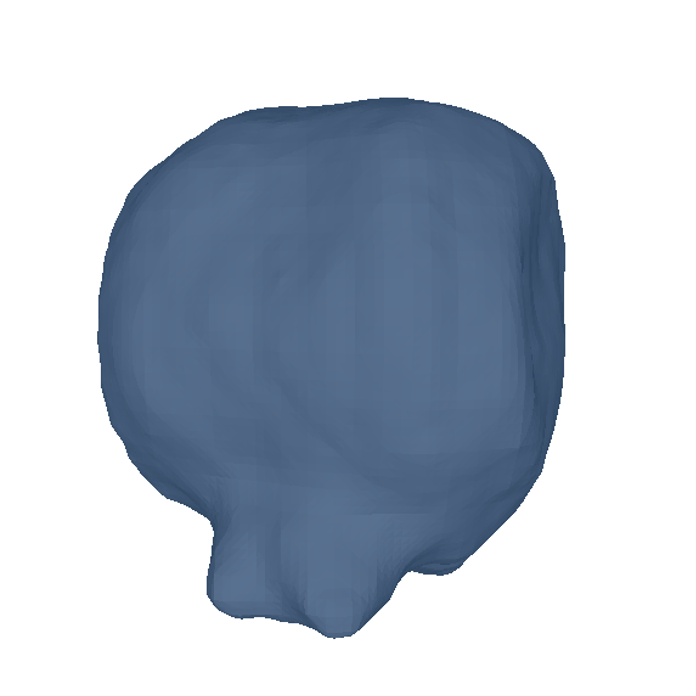}
			  	\end{overpic}
			  }  
                \\
                &
                \multicolumn{1}{c|}{Input views} &
                \multicolumn{3}{c|}{One-2-3-45} &
                \multicolumn{3}{c}{Ours}
		\end{tabular}
	\end{tabular}
	\caption{\textbf{3D mesh reconstruction with One-2-3-45~\cite{liu2023one2345}.} We compare our method's performance on 3D mesh reconstruction with One-2-3-45. Both texture and textureless meshes are shown. The baseline sometimes fails to recover fine details as they can only take one input view, while our method shows better mesh reconstruction with two input views.}
\label{fig:exp-aba-3dreconstruction}
\end{figure*}
\subsection{3D Reconstruction from 2D Novel Views} 
\label{subsec:3Drecon}
Previous sections primarily focused on synthesizing novel views from unposed input images, however, this technique can be extended to render 3D objects by generating multiple random novel views. Following the 3D reconstruction process described in One-2-3-45\cite{liu2023one2345}, we input these synthetic images and their corresponding poses into an SDF-based neural surface reconstruction module to achieve $360^{\circ}$ mesh reconstruction. Figure~\ref{fig:exp-aba-3dreconstruction} demonstrates the results of this process, showing 3D reconstructions of both textured and textureless meshes derived from two unposed input images.

\subsection{More Input Views Yield Better Reconstruction}
\label{subsec:moreinputviews}
In Sec.~\ref{subsec:3D_Estep}, Sec.~\ref{subsec:3D_Mstep} and Sec.~\ref{subsec:3Drecon}, we primarily explore 3D reconstruction based on images from two unposed views. This approach is readily adaptable to scenarios involving an arbitrary number of views. We evaluate the quality of 3D reconstructions using one, two, and three unposed images, as shown in Fig.~\ref{fig:exp-aba-viewconsistent}. The results indicate that adding more views significantly improves the fidelity of the 3D reconstruction. The LatentDEM framework facilitates consistent 3D reconstruction across different images. Specifically, 3D reconstruction from a single view (equivalent to Zero123) results in a hollow reconstruction, whereas incorporating additional views progressively yields more realistic 3D models.

\section{More Ablations and Experimental Results}
\label{app:re}
% Table2和figure7搞成一行？

% \paragraph{Annealing consistency.}
% As shown in Fig.~\ref{fig:appendix-anneal}, we provide an analysis of the effects of different annealing schemes.

\begin{figure*}
	% \vspace*{-0.5cm}
	\centering
	\setlength{\tabcolsep}{1pt}
	\setlength{\fboxrule}{1pt}
	%\vspace*{1.5cm}
	\begin{tabular}{c}
		\begin{tabular}{cccc}
			\multicolumn{1}{c}{
				\begin{overpic}[width=0.20\linewidth]{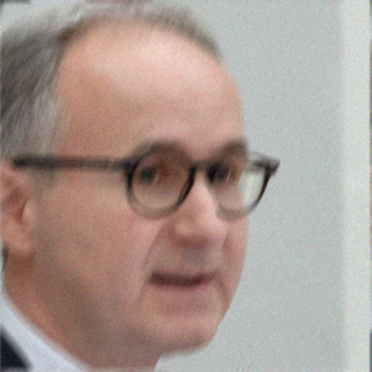}
				\end{overpic}
			}  &
                \multicolumn{1}{c}{
				\begin{overpic}[width=0.20\linewidth]{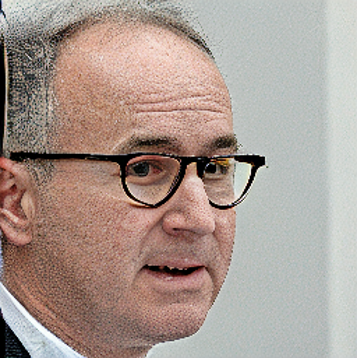}
				\end{overpic}
			}  &
                \multicolumn{1}{c}{
				\begin{overpic}[width=0.20\linewidth]{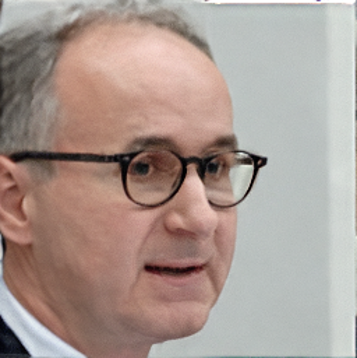}
				\end{overpic}
			}  &
			\multicolumn{1}{c}{
				\begin{overpic}[width=0.20\linewidth]{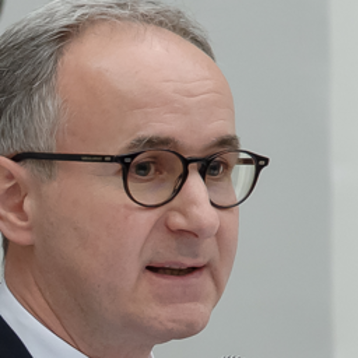}
				\end{overpic}
			}  
			\\[-0.5ex]
			  \multicolumn{1}{c}{
			  	\begin{overpic}[width=0.20\linewidth]{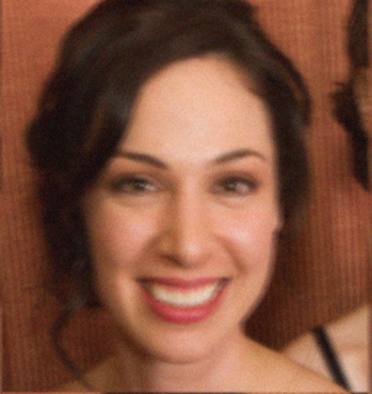}
			  	\end{overpic}
			  }  &
                \multicolumn{1}{c}{
				\begin{overpic}[width=0.20\linewidth]{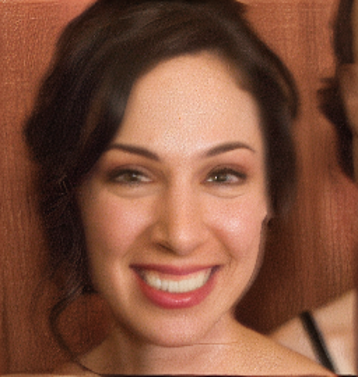}
				\end{overpic}
			}  &
                \multicolumn{1}{c}{
				\begin{overpic}[width=0.20\linewidth]{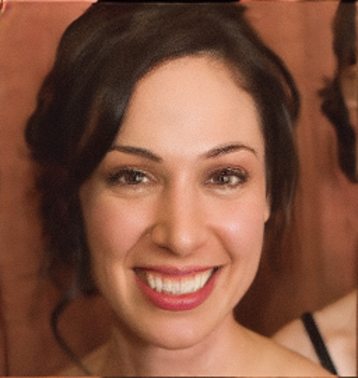}
				\end{overpic}
			}  &
			  \multicolumn{1}{c}{
			  	\begin{overpic}[width=0.20\linewidth]{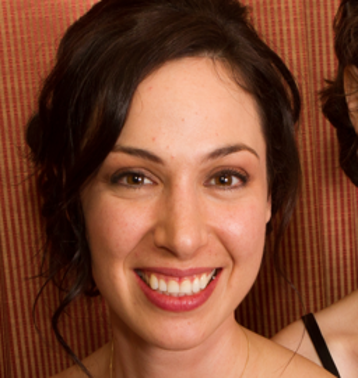}
			  	\end{overpic}
			  }  
                \\
                \multicolumn{1}{c}{Observation} &
                \multicolumn{1}{c}{LDMVQ-4} &
                \multicolumn{1}{c}{Stable Diffusion} &
                \multicolumn{1}{c}{Ground Truth}
		\end{tabular}
	\end{tabular}
	\caption{\textbf{LDMVQ-4 v.s. Stable Diffusion-V1.5.} Stable Diffusion V1.5 generates results with more detailed textures due to its more powerful priors.}
\label{fig:exp-aba-vq4sd}
\end{figure*}

\paragraph{LDMVQ-4 v.s. Stable Diffusion-V1.5.}
We evaluate LatentDEM's performance on 2D blind deblurring using two widely adopted foundational latent diffusion models: LDMVQ-4 and Stable Diffusion V1.5 \cite{rombach2022high}. Fig.~\ref{fig:exp-aba-vq4sd} presents the results of this comparison. Both models achieve satisfactory reconstruction outcomes, demonstrating LatentDEM's ability to generalize across various LDM priors when solving blind inverse problems. Notably, LatentDEM exhibits superior performance with Stable Diffusion compared to LDMVQ-4. This difference can be attributed to Stable Diffusion's more recent release and its reputation as a more advanced diffusion model. % is a latent-based model released before Stable Diffusion. 

\paragraph{Vanilla EM v.s. Diffusion EM.}
% We leverage EM algorithm in both our motion deblur task and 3D reconstruction task. However, in both settings, we make slight changes compared to the vanilla EM algorithm.  Note that the vanilla EM requires getting the best samples under current forward parameters and vice versa. while the diffusion EM may just perform one optimization iteration in the E-step or M-step. 
Traditional EM algorithms perform both E-step and M-step until convergence at each iteration before alternating. This approach guarantees a local optimal solution, as demonstrated by numerous EM studies. However, in the context of diffusion posterior sampling, which involves multiple reverse diffusion steps, this paradigm proves inefficient and computationally expensive.

Our implementation of the EMDiffusion algorithm for blind inverse tasks deviates from this conventional approach. Instead of waiting for the diffusion process to converge (typically 1,000 reverse steps with a DDIM scheduler), we perform model parameter estimation (M-step) after each single diffusion reverse step, except during the initial stage where we employ the skip gradient technique, as detailed in Sec.~\ref{sec:M-step}. Table~\ref{tab:vanillaem} compares the performance of vanilla EM and our method on the blind deblurring task. Our approach, LatentDEM, requires only 1.5 minutes compared to vanilla EM's 120 minutes, while achieving superior reconstruction quality and kernel estimation.

These results demonstrate that performing the M-step after each reverse step is both effective and efficient for blind inverse tasks. Moreover, this strategy offers improved escape from local minima and convergence to better solutions compared to vanilla EM, which completes the entire diffusion reverse process in each EM iteration.

\begin{table}[htbp]
  \centering
  \caption{Vanilla EM v.s. Diffusion EM in blind deblurring.}
    \begin{tabular}{ccccccc}
    \toprule
    \multirow{2}[4]{*}{\textbf{Method}} & 
    \multicolumn{3}{c}{\textbf{Image}} & 
    \multicolumn{2}{c}{\textbf{Kernel}} & 
    \textbf{Time} \\
\cmidrule{2-7}          & PSNR  & SSIM  & LPIPS & MSE   & MNC   & Minute \\
    \midrule
    Vanilla EM & 20.43 & 0.561 & 0.419 & 0.024 & 0.124 & 120 \\
    LatentDEM & 22.23 & 0.695 & 0.183 & 0.023 & 0.502 & 1.5 \\
    \bottomrule
    \end{tabular}%
  \label{tab:vanillaem}%
\end{table}%

\begin{figure*}
	% \vspace*{-0.5cm}
	\centering
	\setlength{\tabcolsep}{1pt}
	\setlength{\fboxrule}{1pt}
	%\vspace*{1.5cm}
	\begin{tabular}{c}
		\begin{tabular}{cc|cccc}
			\multicolumn{1}{c}{
				\begin{overpic}[width=0.16\linewidth]{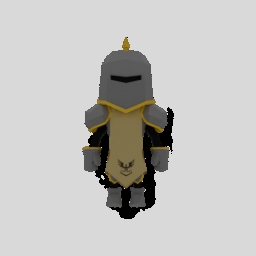}
				\end{overpic}
			}  &
                \multicolumn{1}{c|}{
				\begin{overpic}[width=0.16\linewidth]{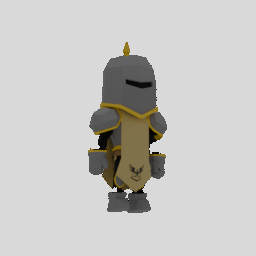}
				\end{overpic}
			}  &
                \multicolumn{1}{c}{
				\begin{overpic}[width=0.16\linewidth]{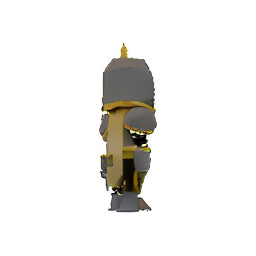}
				\end{overpic}
			}  &
                \multicolumn{1}{c}{
				\begin{overpic}[width=0.16\linewidth]{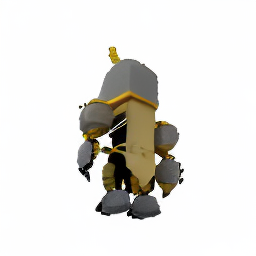}
				\end{overpic}
			}  &
                \multicolumn{1}{c}{
				\begin{overpic}[width=0.16\linewidth]{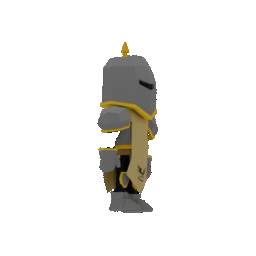}
				\end{overpic}
			}  &
			\multicolumn{1}{c}{
				\begin{overpic}[width=0.16\linewidth]{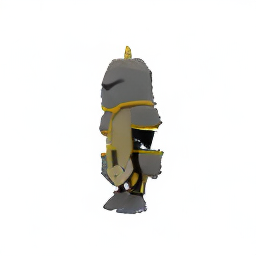}
				\end{overpic}
			}  
			\\
			  \multicolumn{1}{c}{
				\begin{overpic}[width=0.16\linewidth]{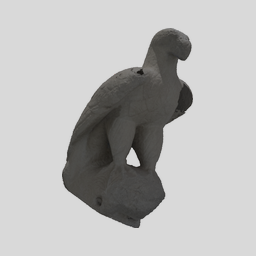}
				\end{overpic}
			}  &
                \multicolumn{1}{c|}{
				\begin{overpic}[width=0.16\linewidth]{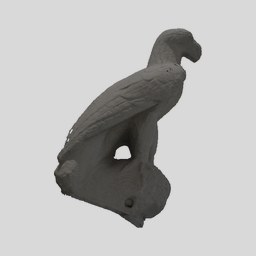}
				\end{overpic}
			}  &
                \multicolumn{1}{c}{
				\begin{overpic}[width=0.16\linewidth]{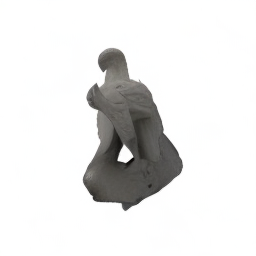}
				\end{overpic}
			}  &
                \multicolumn{1}{c}{
				\begin{overpic}[width=0.16\linewidth]{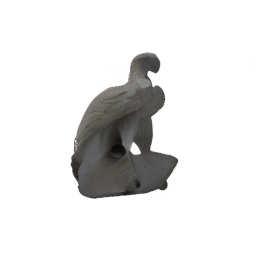}
				\end{overpic}
			}  &
                \multicolumn{1}{c}{
				\begin{overpic}[width=0.16\linewidth]{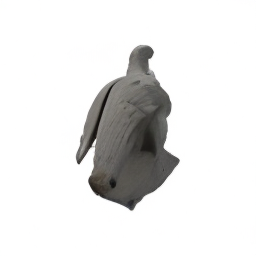}
				\end{overpic}
			}  &
			\multicolumn{1}{c}{
				\begin{overpic}[width=0.16\linewidth]{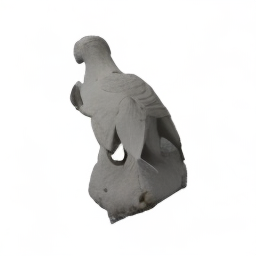}
				\end{overpic}
			}  
                \\
                \multicolumn{1}{c}{
				\begin{overpic}[width=0.16\linewidth]{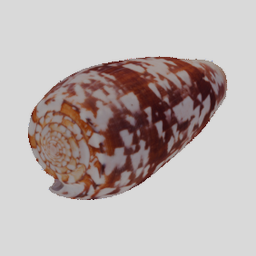}
				\end{overpic}
			}  &
                \multicolumn{1}{c|}{
				\begin{overpic}[width=0.16\linewidth]{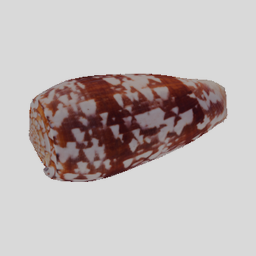}
				\end{overpic}
			}  &
                \multicolumn{1}{c}{
				\begin{overpic}[width=0.16\linewidth]{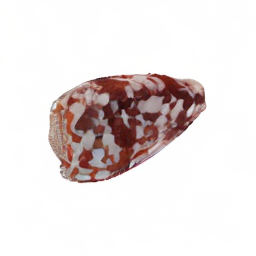}
				\end{overpic}
			}  &
                \multicolumn{1}{c}{
				\begin{overpic}[width=0.16\linewidth]{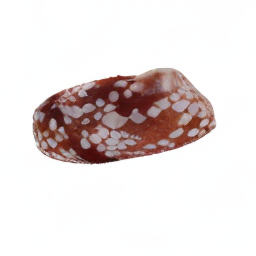}
				\end{overpic}
			}  &
                \multicolumn{1}{c}{
				\begin{overpic}[width=0.16\linewidth]{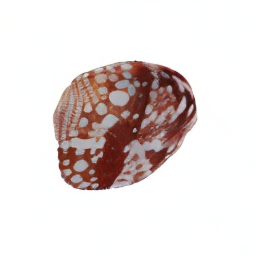}
				\end{overpic}
			}  &
			\multicolumn{1}{c}{
				\begin{overpic}[width=0.16\linewidth]{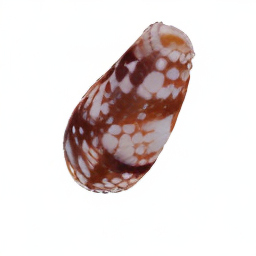}
				\end{overpic}
			}  
                \\
                \multicolumn{1}{c}{
				\begin{overpic}[width=0.16\linewidth]{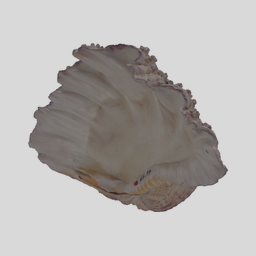}
				\end{overpic}
			}  &
                \multicolumn{1}{c|}{
				\begin{overpic}[width=0.16\linewidth]{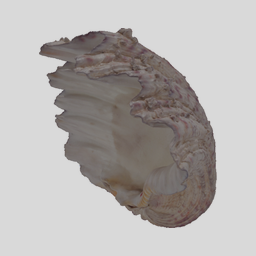}
				\end{overpic}
			}  &
                \multicolumn{1}{c}{
				\begin{overpic}[width=0.16\linewidth]{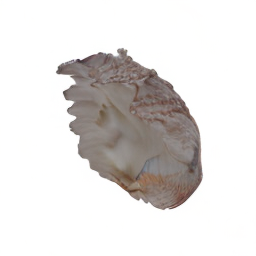}
				\end{overpic}
			}  &
                \multicolumn{1}{c}{
				\begin{overpic}[width=0.16\linewidth]{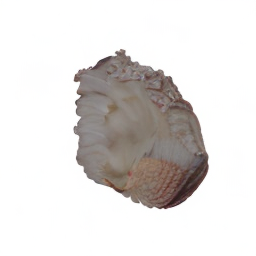}
				\end{overpic}
			}  &
                \multicolumn{1}{c}{
				\begin{overpic}[width=0.16\linewidth]{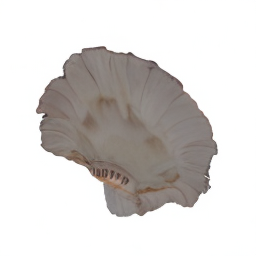}
				\end{overpic}
			}  &
			\multicolumn{1}{c}{
				\begin{overpic}[width=0.16\linewidth]{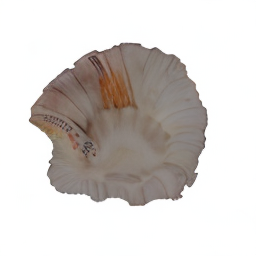}
				\end{overpic}
			}  
                \\
                \multicolumn{1}{c}{
				\begin{overpic}[width=0.16\linewidth]{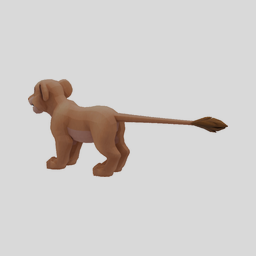}
				\end{overpic}
			}  &
                \multicolumn{1}{c|}{
				\begin{overpic}[width=0.16\linewidth]{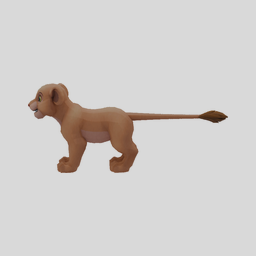}
				\end{overpic}
			}  &
                \multicolumn{1}{c}{
				\begin{overpic}[width=0.16\linewidth]{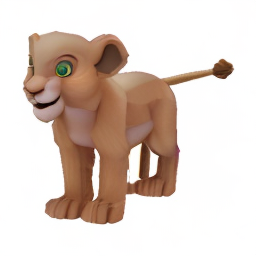}
				\end{overpic}
			}  &
                \multicolumn{1}{c}{
				\begin{overpic}[width=0.16\linewidth]{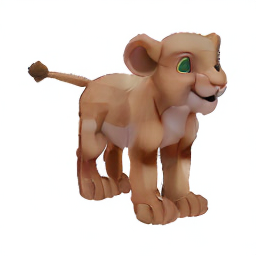}
				\end{overpic}
			}  &
                \multicolumn{1}{c}{
				\begin{overpic}[width=0.16\linewidth]{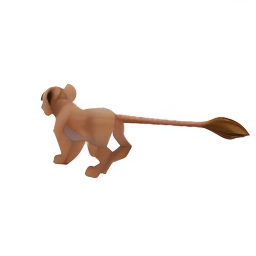}
				\end{overpic}
			}  &
			\multicolumn{1}{c}{
				\begin{overpic}[width=0.16\linewidth]{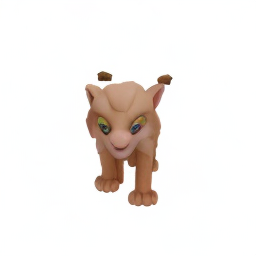}
				\end{overpic}
			}  
                \\
                \multicolumn{1}{c}{
				\begin{overpic}[width=0.16\linewidth]{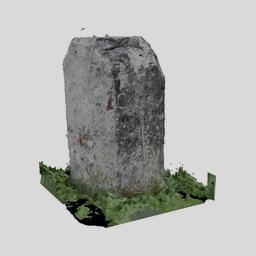}
				\end{overpic}
			}  &
                \multicolumn{1}{c|}{
				\begin{overpic}[width=0.16\linewidth]{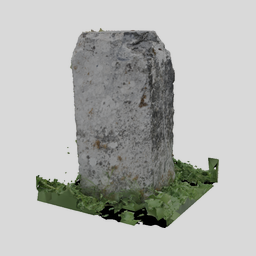}
				\end{overpic}
			}  &
                \multicolumn{1}{c}{
				\begin{overpic}[width=0.16\linewidth]{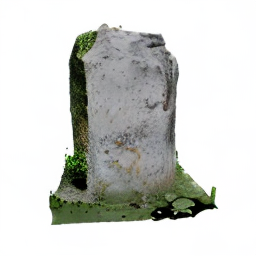}
				\end{overpic}
			}  &
                \multicolumn{1}{c}{
				\begin{overpic}[width=0.16\linewidth]{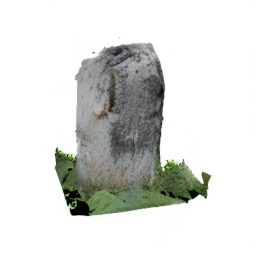}
				\end{overpic}
			}  &
                \multicolumn{1}{c}{
				\begin{overpic}[width=0.16\linewidth]{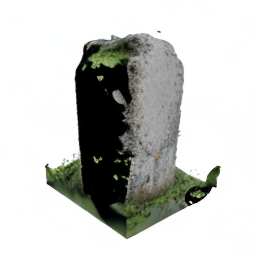}
				\end{overpic}
			}  &
			\multicolumn{1}{c}{
				\begin{overpic}[width=0.16\linewidth]{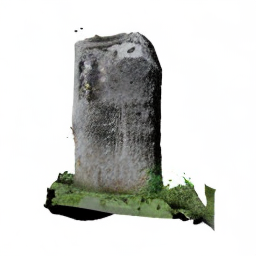}
				\end{overpic}
			}  
                \\
                \multicolumn{1}{c}{
				\begin{overpic}[width=0.16\linewidth]{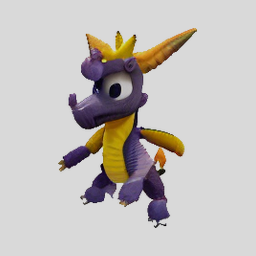}
				\end{overpic}
			}  &
                \multicolumn{1}{c|}{
				\begin{overpic}[width=0.16\linewidth]{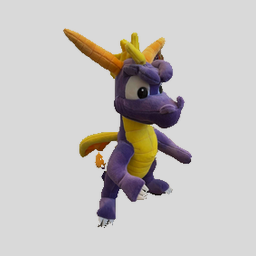}
				\end{overpic}
			}  &
                \multicolumn{1}{c}{
				\begin{overpic}[width=0.16\linewidth]{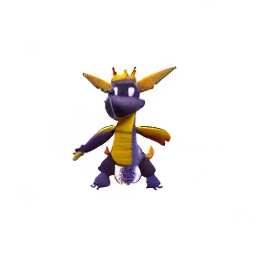}
				\end{overpic}
			}  &
                \multicolumn{1}{c}{
				\begin{overpic}[width=0.16\linewidth]{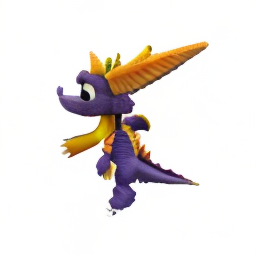}
				\end{overpic}
			}  &
                \multicolumn{1}{c}{
				\begin{overpic}[width=0.16\linewidth]{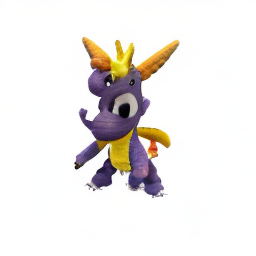}
				\end{overpic}
			}  &
			\multicolumn{1}{c}{
				\begin{overpic}[width=0.16\linewidth]{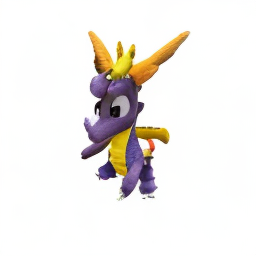}
				\end{overpic}
			}  
                \\
                \multicolumn{1}{c}{
				\begin{overpic}[width=0.16\linewidth]{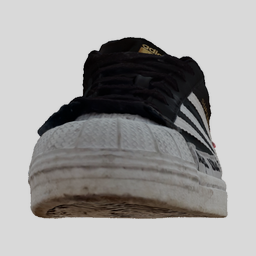}
				\end{overpic}
			}  &
                \multicolumn{1}{c|}{
				\begin{overpic}[width=0.16\linewidth]{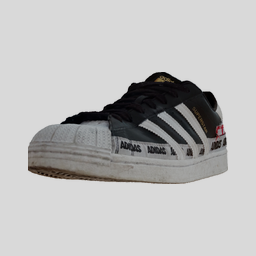}
				\end{overpic}
			}  &
                \multicolumn{1}{c}{
				\begin{overpic}[width=0.16\linewidth]{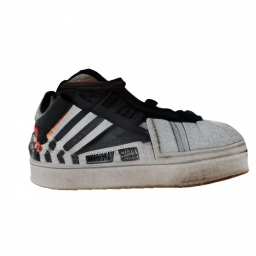}
				\end{overpic}
			}  &
                \multicolumn{1}{c}{
				\begin{overpic}[width=0.16\linewidth]{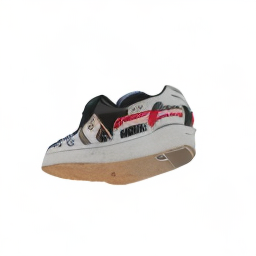}
				\end{overpic}
			}  &
                \multicolumn{1}{c}{
				\begin{overpic}[width=0.16\linewidth]{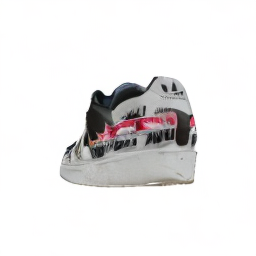}
				\end{overpic}
			}  &
			\multicolumn{1}{c}{
				\begin{overpic}[width=0.16\linewidth]{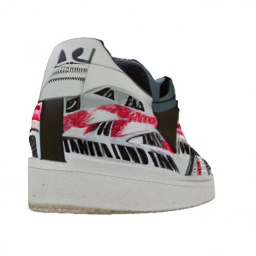}
				\end{overpic}
			}  
                \\
                \multicolumn{2}{c}{Input views} &
                \multicolumn{4}{c}{Output views}
		\end{tabular}
	\end{tabular}
	\caption{\textbf{Pose-free sparse-view novel-view synthesis.}}
 % \wz{can we add a qualitative comparison with leap like Fig.9? just fill the blank page.}}}
\label{fig:appendix-more3d}
\end{figure*}

\paragraph{Additional Results of Pose-free Sparse-view Novel-view Synthesis.}
Figure~\ref{fig:appendix-more3d} presents additional results of novel-view synthesis using LatentDEM.
% Most output views are consistent with the given input views, but few output views can be further improved with better LDM priors. 

% \begin{figure*}[tbp]
% \centering
% \includegraphics[width=12.5cm]{Styles/figure/blindinverse-appendix-more3d.pdf}
% \centering
% \caption{{Pose-free Sparse-view 3D Reconstruction.}}
% \label{fig:appendix-more3d}
% \end{figure*}

% todo lion

%%%%%%%%%%%%%%%%%%%%%%%%%%%%%%%%%%%%%%%%%%%%%%%%%%%%%%%%%%%%

% \input{checklist}

\end{document}